\documentclass{article} 
\usepackage{iclr2027_conference_arxiv,times}

\usepackage{amsmath,amsfonts,bm}

\def\eqref#1{equation~\ref{#1}}

\def\1{\bm{1}}

\DeclareMathAlphabet{\mathsfit}{\encodingdefault}{\sfdefault}{m}{sl}
\SetMathAlphabet{\mathsfit}{bold}{\encodingdefault}{\sfdefault}{bx}{n}

\usepackage{hyperref}
\usepackage{url}
\usepackage{graphicx}
\usepackage{comment}
\usepackage{booktabs}
\usepackage{amsmath,amssymb}\usepackage{algorithm}
\usepackage{algpseudocode}
\usepackage{titletoc}
\usepackage{xcolor}
\usepackage{rotating}
\usepackage{colortbl}
\usepackage{placeins}

\title{SLIM: Simplex-Lattice Interpolation Merging}

\author{Seongcheol Jeong, Masahiro Suzuki \& Yutaka Matsuo\\
The University of Tokyo\\
\texttt{jeong@weblab.t.u-tokyo.ac.jp} \\
}

\iclrfinalcopy 
\begin{document}

\maketitle

\begin{abstract}
Optimizing merging coefficients for large language models can require many costly benchmark evaluations.
We propose \textbf{Simplex-Lattice Interpolation Merging (SLIM)}, which constructs a quadratic surrogate of aggregate performance on the coefficient simplex using a classical mixture design.
Evaluations of individual experts and equal-weight pairs determine the surrogate with the minimum number of measurements needed to identify a general quadratic on this domain.
SLIM then optimizes the surrogate without further target-metric evaluations.
Experiments on two model architectures demonstrate accurate prediction of unseen multi-expert mixtures and competitive merge performance under limited evaluation budgets.
Matched-budget comparisons show that structured evaluation points improve prediction fidelity over random designs, including those using regularized fitting.
\end{abstract}

\section{Introduction}
\label{sec:intro}

Model merging can combine capabilities acquired by multiple trained models without further parameter training~\citep{yang2026model,song2026model}.
For large language models (LLMs), reusing publicly available fine-tuned checkpoints can reduce the additional cost of building models with multiple capabilities.
We study the widely used setting of merging \emph{experts}: models with the same architecture fine-tuned from a common base.
Task Arithmetic~\citep{ilharco2023editing} adds a weighted sum of task vectors---the parameter differences between experts and the base---to the base model.
Even in this linear setting, the coefficients controlling each expert's contribution affect merge performance.

We call the downstream measure to be improved the \emph{target metric}; here, we use AVG, the unweighted mean of domain scores on a common scale.
On generation-based LLM benchmarks, evaluating a candidate requires constructing merged weights, generating responses, and scoring outputs, making searches such as evolutionary merging costly~\citep{akiba2025evolutionary,su2025fine}.
A surrogate that predicts performance from coefficients can support coefficient selection using inexpensive predictions~\citep{li2024map}.
For example, MAP fits task-wise surrogates to measured scores and uses them to approximate Pareto fronts.
However, constructing the surrogate itself incurs an upfront evaluation cost because fitting it requires target-metric measurements from merged models.
For LLMs, these measurements remain expensive even when subsequent surrogate optimization is cheap.
We therefore ask: 

\begin{center}
    \emph{How few target-metric evaluations are needed to construct a surrogate\\ that supports competitive merge performance?}
\end{center}

Quadratic surrogates capture curvature and pairwise interactions with relatively few parameters, a choice also used by MAP~\citep{li2024map}.
To reduce the required evaluations, we restrict coefficients to the simplex, where they are nonnegative and sum to one.
Weight averaging and MetaGPT provide precedents for effective merging within this domain~\citep{wortsman2022model,zhou2024metagpt}.
For $T$ coefficients, the restriction reduces the number of independent quadratic parameters from $(T+1)(T+2)/2$ to $T(T+1)/2$.
However, the identification minimum does not guarantee stable fitting or accurate prediction, which also depend on evaluation placement and the adequacy of the quadratic approximation.
We test whether measurements of individual experts and pairs suffice to predict AVG at unseen mixtures, and whether coefficient selection within the simplex yields competitive merge performance.

We propose Simplex-Lattice Interpolation Merging (SLIM), which constructs a quadratic surrogate from exactly $T(T+1)/2$ target-metric evaluations.
SLIM adopts Scheff\'{e}'s $\{T,2\}$ simplex-lattice design from classical mixture experiments~\citep{scheffe1958experiments}.
Its vertices and edge midpoints correspond to individual experts and equal-weight expert pairs.
SLIM interpolates their measured AVG values and numerically optimizes the resulting quadratic over the simplex using its analytic gradient, without further target-metric evaluations.

\begin{figure}[t]
\centering
\includegraphics[width=\linewidth]{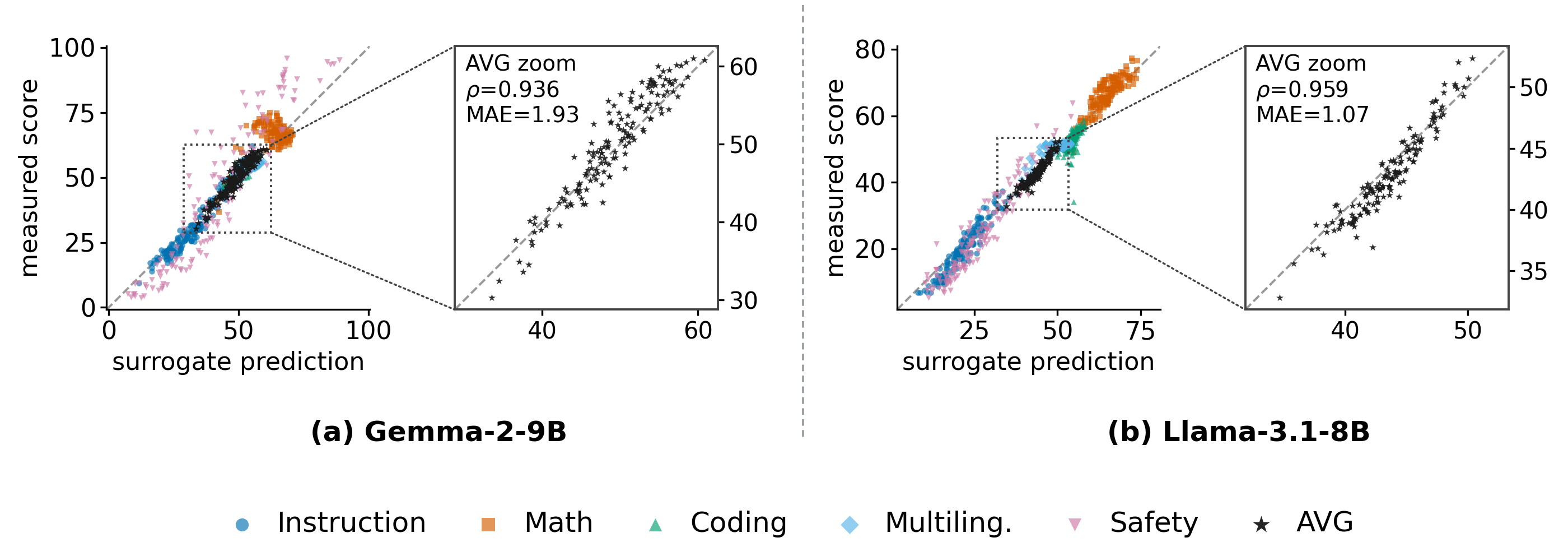}
\caption{A quadratic surrogate constructed from individual experts and equal-weight pairs accurately predicts AVG at unseen mixtures of all eight experts.
For each architecture, 36 construction evaluations are followed by testing on 150 held-out interior vectors sampled from $\operatorname{Dirichlet}(1,\ldots,1)$.
Axes show predicted (horizontal) and measured (vertical) scores; colored markers show domain scores and black stars show AVG.
AVG zoom panels report Spearman's $\rho$ and MAE; dashed lines indicate $y=x$.}
\label{fig:fidelity}
\end{figure}

We evaluate SLIM on Gemma-2-9B~\citep{team2024gemma} and Llama-3.1-8B~\citep{grattafiori2024llama}, each with eight experts spanning five domains.
The surrogates accurately predict AVG at unseen mixtures involving all eight experts (Figure~\ref{fig:fidelity}).
At the same evaluation budget, the lattice design improves prediction fidelity over random designs, including ridge-regularized fits.
We assess selected models on in-distribution (ID) benchmarks used for selection and held-out out-of-distribution (OOD) benchmarks.
With 36 evaluations, SLIM exceeds MAP at 45 evaluations in all four architecture-by-distribution settings and remains competitive with evolutionary merging using 150--200 evaluations in three settings (Figure~\ref{fig:cost_perf}).
On Llama ID, methods searching the broader coefficient domain achieve higher performance.

\paragraph{Contributions} Our contributions are twofold:
\begin{itemize}
    \item \textbf{Predictability:} We show that evaluations of individual experts and equal-weight pairs support accurate quadratic prediction of AVG at unseen mixtures involving all eight experts on both architectures studied.
    \item \textbf{Minimal-cost selection:} We propose SLIM, which selects coefficients using a surrogate constructed from the $T(T+1)/2$ evaluations needed to identify a general quadratic on the simplex.
\end{itemize}

\section{Related Work}
\label{sec:related}

\paragraph{Model Merging.}
A \emph{merging operator} specifies how expert parameters or task vectors are transformed and combined.
Both the operator and the coefficients controlling each expert's contribution affect merge performance.
Task Arithmetic~\citep{ilharco2023editing} combines task vectors linearly, while TIES-Merging~\citep{yadav2023ties} and DARE~\citep{yu2024language} transform parameter updates to address interference or redundancy.
Existing methods select coefficients using data-free rules~\citep{zhou2024metagpt}, prediction entropy minimization~\citep{AdaMerging_ICLR_2024}, or search guided by measured downstream performance~\citep{akiba2025evolutionary}.
Evaluating candidate LLM merges can be computationally expensive, especially on generation-based benchmarks that require generating and scoring model outputs~\citep{su2025fine}.
To reduce evaluation costs, MERGE$^3$~\citep{mencattini2025merge3} combines evolutionary merging with dataset reduction and performance estimation based on item response theory.
We focus on the widely studied setting of linear task-vector merging of LLM experts fine-tuned from a common base~\citep{yang2026model,song2026model}.
Within this setting, we investigate how few direct target-metric evaluations can support coefficient selection that yields competitive merge performance.

\paragraph{Surrogate-Based Coefficient Optimization.}
Surrogate-based methods fit predictive models to measured merge performance and use their predictions to guide coefficient or configuration search.
MM-MO~\citep{li2025morphing} sequentially selects DARE--TIES coefficients and sparsity levels using Gaussian process surrogates and multi-objective Bayesian optimization.
Multi-fidelity merging~\citep{su2025fine} uses a random forest surrogate to select configurations and varies the size of the validation set to allocate evaluation resources.
MAP~\citep{li2024map} is closely related to our work: it uses surrogates based on quadratic functions to approximate task performance and estimate Pareto fronts.
In its basic form, MAP fits these surrogates to scores measured at randomly sampled coefficient vectors, without requiring the coefficients to sum to one.
Constructing these surrogates incurs an upfront evaluation cost because the target metrics must be measured on merged models.
Bayesian MAP and Nested MAP reduce evaluation costs through adaptive sampling and hierarchical merging, respectively, but both rely on measured scores for surrogate construction.
For $T$ experts, SLIM uses a fixed experimental design to interpolate AVG on the coefficient simplex from $T(T+1)/2$ evaluations, the minimum needed to identify a general quadratic on this domain.
We test whether this surrogate can predict AVG at unseen multi-expert mixtures and support coefficient selection that yields competitive merge performance.

\paragraph{Mixture Experiments.}
Mixture experiments are a classical statistical framework for studying how a response depends on the relative proportions of components.
Applications include studying how ingredient proportions affect food flavor and how solvent proportions affect solubility in mixed-solvent formulations~\citep{cornell2002experiments}.
Since component proportions are nonnegative and sum to one, the experimental domain is a simplex.
We view linear task-vector merging under this simplex constraint as a mixture experiment.
The components are LLM experts, their merging coefficients specify the proportions, and the response is the merged model's target metric, AVG.
For $T$ components, a general quadratic on the simplex has $T(T+1)/2$ independent coefficients.
Scheff\'{e}'s $\{T,2\}$ simplex-lattice design uniquely determines this quadratic from evaluations at the $T$ vertices and $\binom{T}{2}$ edge midpoints~\citep{scheffe1958experiments}.
Under a quadratic regression model with uncorrelated, equal-variance errors, a design that allocates observations equally across these points is D-optimal~\citep{kiefer1961optimum}.
When a design is D-optimal, it maximizes the joint precision of the regression coefficient estimates for a fixed number of observations.
Recent work applies sparse Scheff\'{e} models and optimal design to LLM pretraining data mixtures~\citep{mao2026datamixing}.
In that setting, mixture proportions are the fractions of training tokens drawn from each data domain, and the response is validation loss.
SLIM adopts the classical simplex-lattice design and its quadratic interpolant.
We empirically test whether evaluations of individual experts and equal-weight pairs suffice to predict AVG at unseen mixtures, and whether restricting coefficients to the simplex permits competitive merge performance.

\section{Method}
\label{sec:method}

\subsection{Problem Setting}
\label{sec:problem-setting}

Consider $T$ expert models $\theta_1,\ldots,\theta_T$ fine-tuned
from a common base model $\theta_{\mathrm{pre}}$.
With task vectors $\tau_k=\theta_k-\theta_{\mathrm{pre}}$,
linear merging gives
\begin{equation}
\theta(\lambda)
=\theta_{\mathrm{pre}}+\sum_{k=1}^{T}\lambda_k\tau_k.
\label{eq:merge}
\end{equation}

Let $M(\lambda)$ denote the target metric of $\theta(\lambda)$,
with higher values indicating better performance.
For a feasible coefficient set $\Lambda\subseteq\mathbb{R}^{T}$,
the coefficient-selection problem is
\begin{equation}
\lambda^\star
\in\operatorname*{arg\,max}_{\lambda\in\Lambda}M(\lambda).
\label{eq:true-objective}
\end{equation}
In our experiments, $M$ is AVG, the unweighted mean of domain
scores expressed on a common scale.

\subsection{Quadratic Surrogate and Evaluation Design}
\label{sec:simplex-design}
To select coefficients without repeatedly measuring the expensive target metric during optimization, SLIM approximates $M$ with a quadratic surrogate $q$ (Algorithm~\ref{alg:slim}).
Quadratic models capture curvature and pairwise interactions, a choice also used by MAP~\citep{li2024map}.
Our goal is to construct this surrogate from few measurements and then optimize its predictions.

To reduce the number of surrogate parameters to identify, we restrict the feasible set to the simplex:
\begin{equation}
\Lambda=\Delta^{T-1}
=\left\{
\lambda\in\mathbb{R}^{T}
\,\middle|\,
\lambda_k\geq0\ \text{for all }k,\quad
\sum_{k=1}^{T}\lambda_k=1
\right\}.
\label{eq:simplex}
\end{equation}

Any quadratic on this domain has the canonical mixture form~\citep{scheffe1958experiments}
\begin{equation}
q(\lambda)
=\sum_{k=1}^{T}\beta_k\lambda_k
+\sum_{i<j}\beta_{ij}\lambda_i\lambda_j.
\label{eq:quadratic-mixture}
\end{equation}
It has $T(T+1)/2$ independent coefficients.
Identifying them with the minimum number of measurements requires an evaluation design that supplies independent constraints.

Let $e_k$ be the $k$-th standard basis vector in $\mathbb{R}^{T}$.
SLIM adopts Scheff\'{e}'s quadratic $\{T,2\}$ simplex-lattice design~\citep{scheffe1958experiments}, evaluating $M$ at the vertices and edge midpoints:
\begin{equation}
\mathcal{D}
=\{e_k\}_{k=1}^{T}
\cup
\left\{
\frac{e_i+e_j}{2}
\,\middle|\,
1\leq i<j\leq T
\right\}.
\label{eq:design}
\end{equation}
These points correspond to individual experts,
$\theta(e_k)=\theta_k$, and equal-weight pairs,
$\theta((e_i+e_j)/2)=(\theta_i+\theta_j)/2$.
The vertex measurements determine the linear coefficients, and the midpoint measurements determine the pairwise coefficients.
The total evaluation count is
\begin{equation}
|\mathcal{D}|
=T+\binom{T}{2}
=\frac{T(T+1)}{2}.
\label{eq:design-size}
\end{equation}

\begin{figure}[t]
\centering
\includegraphics[width=0.7\linewidth]{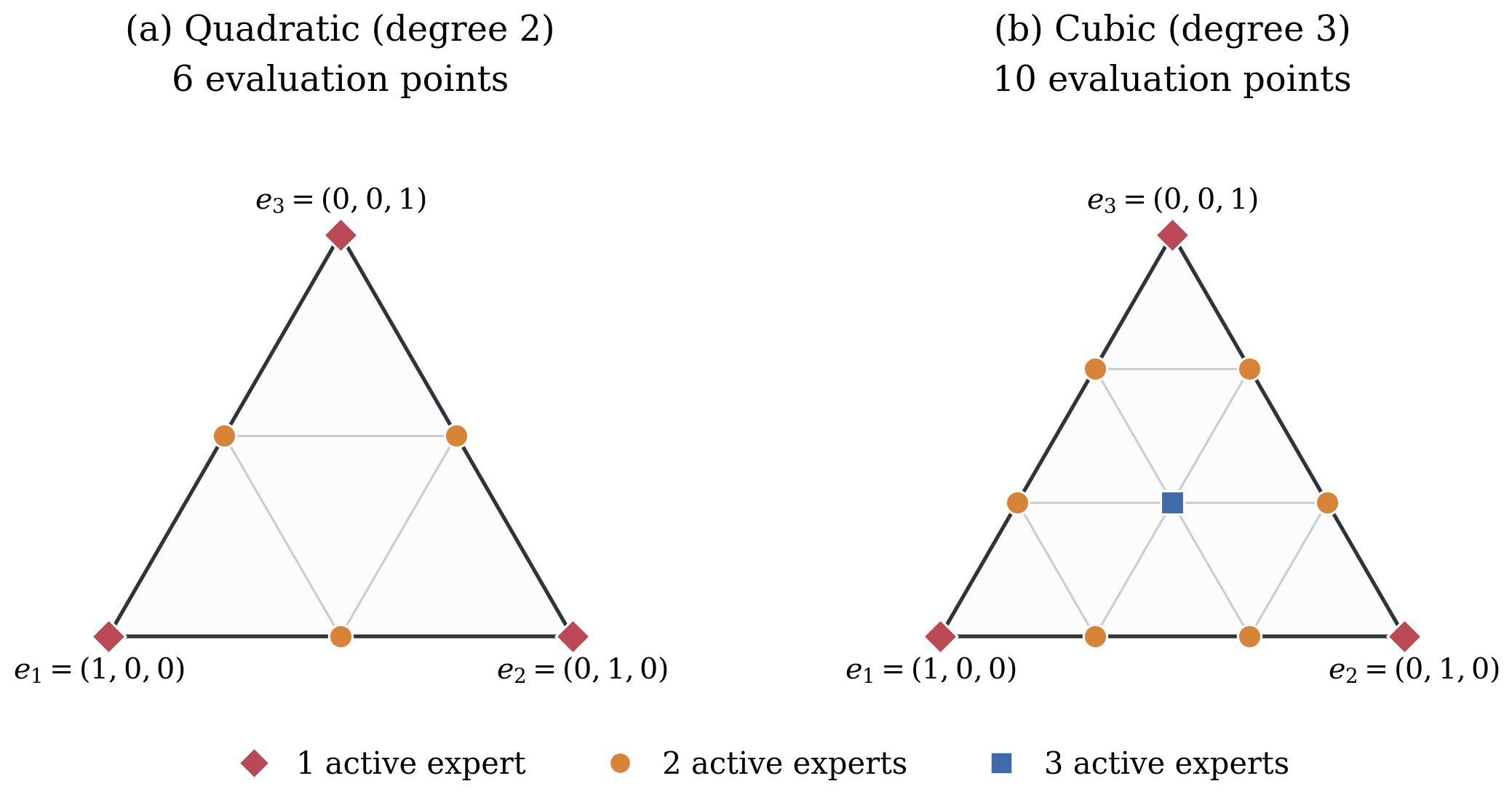}
\caption{Simplex-lattice designs for $T=3$. Each point is a coefficient vector with nonnegative entries summing to one. The quadratic design contains three vertices and three edge midpoints; the cubic design contains the vertices, two points on each edge with weights $(2/3,1/3)$ and $(1/3,2/3)$, and the barycenter $(1/3,1/3,1/3)$. Colors and marker shapes indicate the number of experts with nonzero coefficients. For $T=8$, the corresponding designs contain 36 and 120 points.}
\label{fig:simplex-lattice}
\end{figure}

Figure~\ref{fig:simplex-lattice} illustrates the quadratic design for $T=3$ and its cubic counterpart (Appendix~\ref{app:cubic-extension}).

\subsection{Quadratic Interpolation}
\label{sec:quadratic-interpolation}

Write $M_k=M(e_k)$ and $M_{ij}=M((e_i+e_j)/2)$.
The unique quadratic satisfying these measurements is
\begin{equation}
\widehat{M}(\lambda)
=\sum_{k=1}^{T}\lambda_k(2\lambda_k-1)M_k
+\sum_{i<j}4\lambda_i\lambda_jM_{ij}.
\label{eq:interpolant}
\end{equation}

Each basis function equals one at its associated design point
and zero at the others, so $\widehat{M}$ reproduces all
measurements on $\mathcal{D}$.
Appendix~\ref{app:interpolant-derivation} derives the coefficients and proves uniqueness and the identification minimum.

By linearity, interpolating AVG directly is equivalent to averaging
the domain-wise interpolants (Appendix~\ref{app:aggregation}).
Agreement with $M$ is guaranteed at the design points;
prediction accuracy at unseen mixtures is assessed in
Section~\ref{sec:fidelity}.

\subsection{Coefficient Selection}
\label{sec:coefficient-selection}

A constrained numerical optimizer returns $\widehat{\lambda}$
by approximately solving
\begin{equation}
\underset{\lambda\in\Delta^{T-1}}{\operatorname{maximize}}
\quad \widehat{M}(\lambda).
\label{eq:coefficient-selection}
\end{equation}
Because $\widehat{M}$ need not be concave, the returned solution
is not guaranteed to be a global maximizer.

Surrogate optimization requires no model evaluations beyond
the $T(T+1)/2$ design-point measurements.
The final merged model is $\theta(\widehat{\lambda})$
from Eq.~\ref{eq:merge}.

\section{Experiments}
\label{sec:experiments}

\subsection{Experimental Setup}
\label{sec:experimental-setup}

We use Gemma-2-9B~\citep{team2024gemma} and Llama-3.1-8B~\citep{grattafiori2024llama}, each with eight experts fine-tuned from a common base: two each for instruction following, mathematics, and coding, and one each for multilingual translation and safety.
All methods use the same experts and linear task-vector merging.
The target metric is AVG, the unweighted mean of five domain scores on a 0--100 scale.

Table~\ref{tab:benchmarks} lists the benchmarks; metrics and evaluation settings are given in Appendix~\ref{app:benchmarks} (Table~\ref{tab:benchmark-details}).

Search methods use ID AVG for coefficient selection, and the reported ID scores use these same benchmarks; OOD benchmarks are held out from selection.
All merged candidates, including simplex vertices, follow a common evaluation protocol within each architecture and benchmark.
Model identifiers, benchmark sources, and implementation details are provided in Appendix~\ref{app:experiments}.

\begin{table}[t]
\centering
\small
\begin{tabular}{@{}lll@{}}
\toprule
Domain & ID (used for selection) & OOD (held out) \\
\midrule
Instruction  & IFEval       & AlpacaEval 2.0 \\
Math         & GSM8K        & MATH-500 \\
Coding       & MBPP         & HumanEval \\
Multilingual & FLORES+ dev  & FLORES+ devtest \\
Safety       & HarmBench    & WildGuardTest \\
\bottomrule
\end{tabular}
\caption{ID benchmarks for coefficient selection and held-out OOD benchmarks across five domains. Reported ID scores use the selection benchmarks. Metrics, sample counts, and prompting settings appear in Table~\ref{tab:benchmark-details}.}
\label{tab:benchmarks}
\end{table}

\paragraph{Comparisons.}
Weight averaging~\citep{wortsman2022model} sets $\lambda_k=1/T$, while MetaGPT~\citep{zhou2024metagpt} uses
$\lambda_k=\|\tau_k\|_2^2/\sum_{j=1}^{T}\|\tau_j\|_2^2$.
Both determine $\lambda\in\Delta^{T-1}$ without target-metric evaluations.
Task Arithmetic~\citep{ilharco2023editing} evaluates 21 shared coefficients, $\lambda_k=c$ with $c\in\{0,0.05,\ldots,1\}$, and selects the candidate with the highest measured ID AVG.

MAP~\citep{li2024map} samples and optimizes coefficients in $[0,1]^8$.
For each seed, one sequence starts with the eight vertices $e_k$, followed by 127 independent vectors sampled uniformly from $[0,1]^8$.
Budgets $n\in\{45,90,135\}$ use the first $n$ points of this sequence.
MAP fits domain-wise sigmoid-transformed quadratics, uses NSGA-III to approximate a Pareto front, and selects the vector with the highest predicted AVG.

Evolutionary merging~\citep{akiba2025evolutionary} uses CMA-ES to optimize measured ID AVG over $[0,1]^8$, evaluating 10 candidates per generation and retaining the best observed candidate at each budget.
Task Arithmetic, MAP, and evolutionary merging do not impose a sum-to-one constraint.

SLIM constructs a quadratic surrogate of ID AVG from 36 evaluations and optimizes it over $\Delta^{T-1}$ using SLSQP.
A cubic extension uses 120 evaluations as a degree ablation (Table~\ref{tab:selection_ablation}; Appendix~\ref{app:cubic-extension}).
Neither variant requires further target-metric evaluations during surrogate optimization.
Implementation details and optimizer settings are given in Appendix~\ref{app:coefficient-implementations}.

\paragraph{Evaluation cost and runs.}
We report \#Evals as the hardware-independent coefficient-selection budget and GPU-hours as the measured cost in our environment.
Each evaluation constructs, saves, loads, and scores a candidate on all five ID domains using one H100 GPU; GPU-hours sum the elapsed time of these evaluations.
Final-model reporting and surrogate-fidelity measurements are excluded.
MAP and evolutionary merging are reported as mean $\pm$ standard deviation over seeds 42, 43, and 44; other methods use a single run.
SLIM uses a fixed evaluation design and a fixed set of optimizer initializations.

\subsection{Surrogate Fidelity}
\label{sec:fidelity}

For each architecture, we test the quadratic surrogate on 150 coefficient vectors sampled from $\operatorname{Dirichlet}(1,\ldots,1)$ with seed 42.
Each vector assigns nonzero coefficients to all eight experts, whereas every construction point uses at most two.
These points are held out from surrogate construction and coefficient selection; their measured scores use the same ID benchmarks.
Since coefficient selection depends on the relative performance of mixtures, we assess agreement between predicted and measured AVG rankings using Spearman's $\rho$.
We also report mean absolute error (MAE) to quantify score prediction errors.

Figure~\ref{fig:fidelity} compares predicted and measured scores.
For AVG, Spearman's $\rho$ is 0.936 on Gemma-2-9B and 0.959 on Llama-3.1-8B, with MAEs of 1.93 and 1.07 points, respectively.
Fidelity varies across domains: for example, Gemma's math scores have $\rho=-0.24$. Domain-wise results are provided in Appendix~\ref{app:domain-fidelity}.
The evidence therefore supports approximation of the aggregate objective rather than uniformly accurate prediction in every domain.

We also evaluate the quadratic surrogate at two held-out coefficient vectors on each of the 28 simplex edges.
Across these 56 points, MAE is 1.74 on Gemma and 1.38 on Llama; the full edge curves are shown in Appendix~\ref{app:edge-fidelity}.

\subsection{Merge Performance and Evaluation Cost}
\label{sec:merge-performance}
\begin{figure}[t]
\centering
\includegraphics[width=\textwidth]{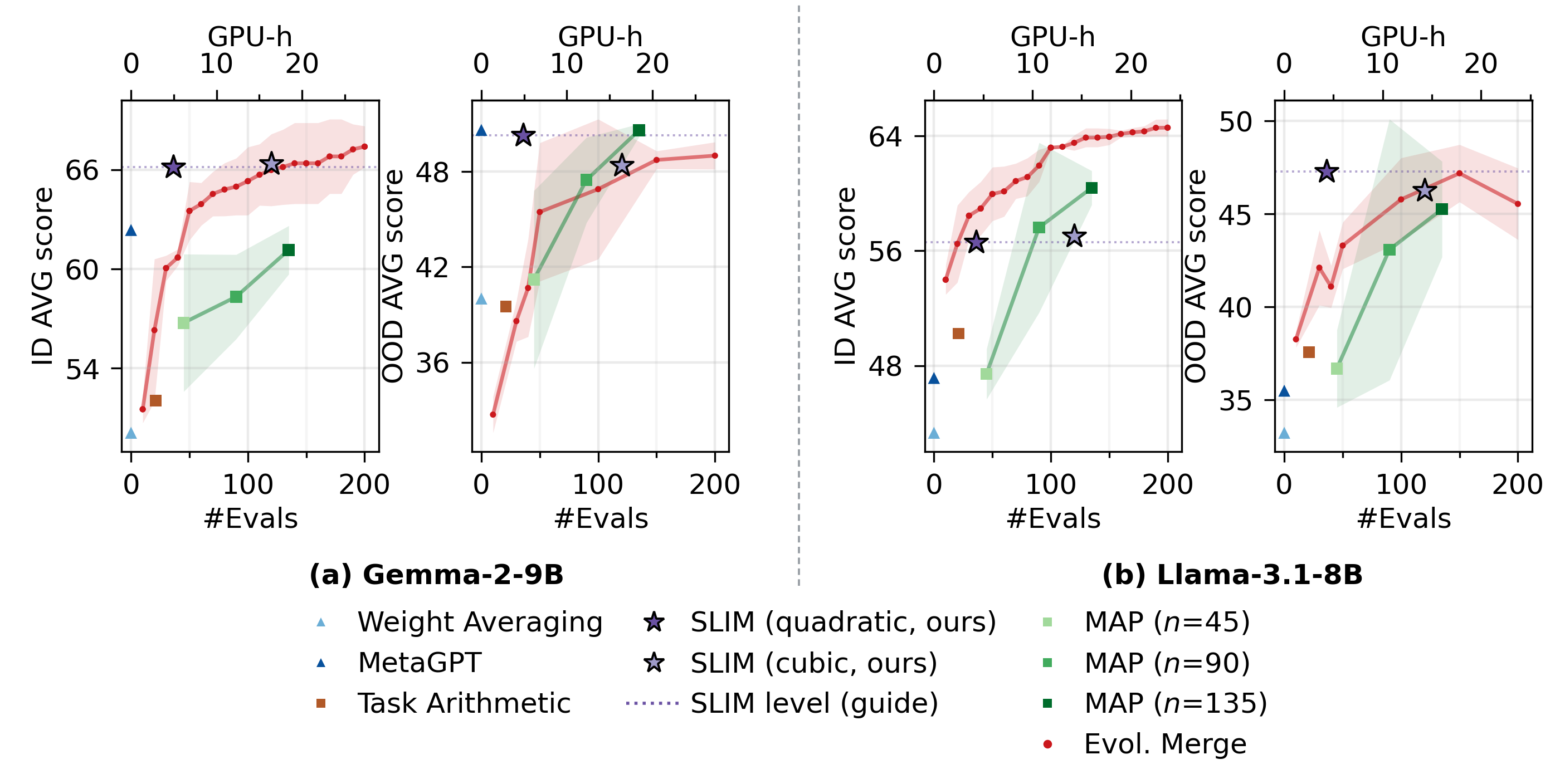}
\caption{ID and held-out OOD AVG versus the evaluation budget for coefficient selection (ID left, OOD right within each architecture).
\#Evals counts cumulative ID evaluations per run; upper axes estimate GPU-hours from measured per-evaluation costs.
Evolutionary merging evaluates 10 candidates per generation.
ID scores are shown after every generation and OOD scores after 10, 30, 40, 50, 100, 150, and 200 evaluations, always for the candidate with the highest ID AVG observed so far.
Quadratic and cubic SLIM use 36 and 120 evaluations; dotted lines mark quadratic SLIM's AVG.
MAP and evolutionary merging show means over three seeds; shaded bands indicate $\pm1$ standard deviation.
Appendix~\ref{app:cost-performance-results} gives the numerical results, including all domain scores and the scores of the base model and the eight experts.}
\label{fig:cost_perf}
\end{figure}

Figure~\ref{fig:cost_perf} shows the cost--performance trade-off; numerical results at selected budgets are provided in Table~\ref{tab:main}.
Quadratic SLIM uses 36 evaluations, costing 5.2 GPU-hours on Gemma-2-9B and 3.9 on Llama-3.1-8B.
It achieves higher AVG than weight averaging in all four architecture-by-distribution settings and exceeds MetaGPT in three; on Gemma OOD, MetaGPT scores slightly higher (50.6 versus 50.3).

SLIM with 36 evaluations outperforms MAP with 45 evaluations in all four settings.
On Gemma ID and Llama OOD, SLIM retains higher AVG at every tested MAP budget.
On Gemma OOD, SLIM also exceeds MAP at 90 evaluations, while MAP at 135 evaluations reaches $50.6\pm0.4$, close to SLIM's 50.3.
On Llama ID, MAP attains higher mean AVG at 90 and 135 evaluations ($57.6\pm5.9$ and $60.4\pm1.2$, respectively) than SLIM's 56.6.

At comparable budgets of 30 and 40 evaluations, evolutionary merging scores below SLIM on Gemma ID, Gemma OOD, and Llama OOD.
On Llama ID, however, it achieves 58.4 and 58.9, respectively, exceeding SLIM's 56.6.
With evolutionary merging extended to 150--200 evaluations, SLIM remains comparable on Gemma ID, higher on Gemma OOD, and comparable on Llama OOD.
These results demonstrate competitive merge performance with a limited evaluation budget in three of the four settings, while also reflecting the methods' different feasible coefficient domains.

\begin{figure}[t]
\centering
\includegraphics[width=0.95\linewidth]{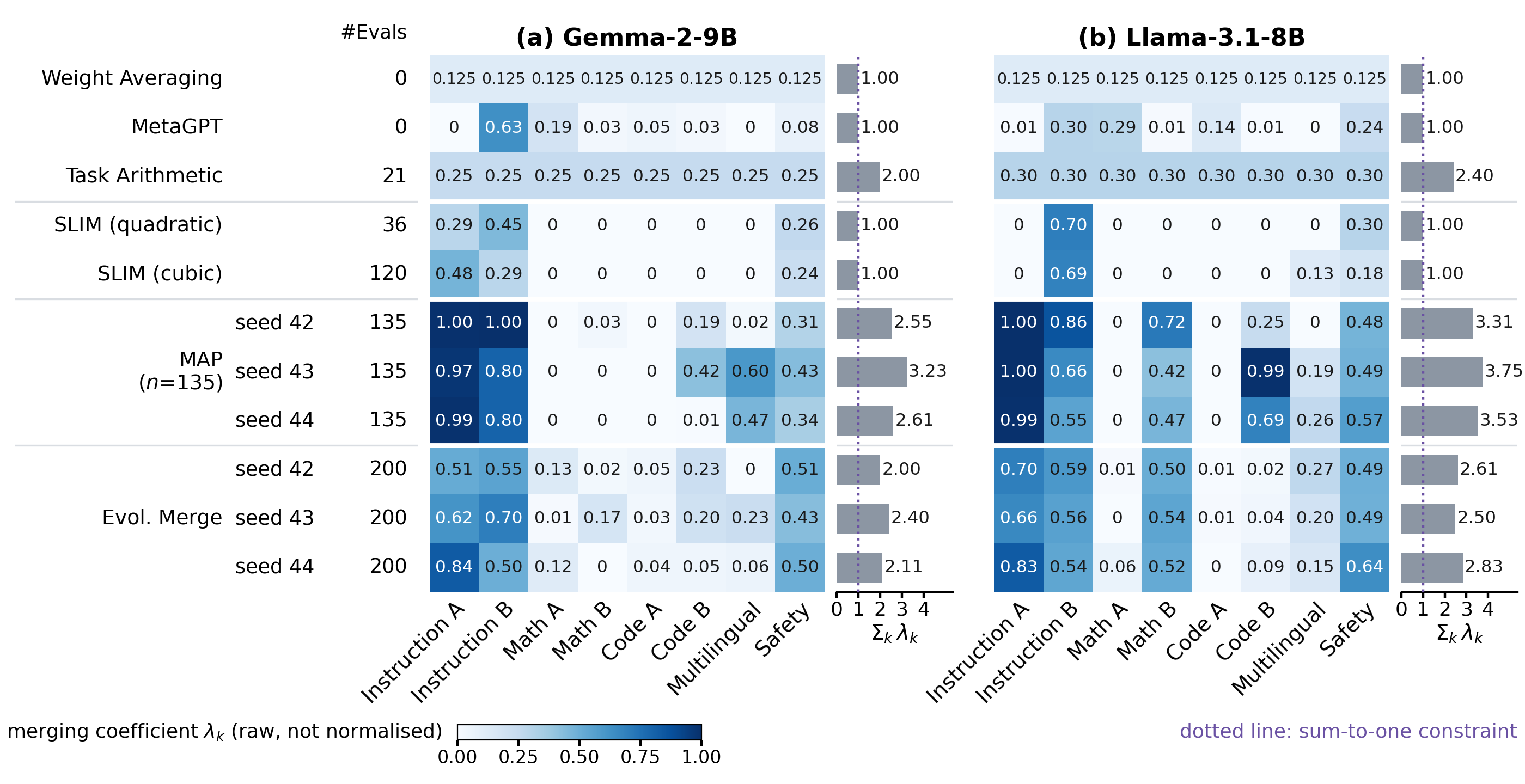}
\caption{Selected merging coefficients reveal differences in expert allocation and overall scale.
Cells show raw coefficients; bars show their sums, with dotted lines marking the sum-to-one constraint.
MAP uses 135 evaluations per seed.
Evolutionary merging shows the candidate with the highest ID AVG observed within 200 evaluations.
Cell labels are rounded, so displayed zeros need not be exact zeros; sums use unrounded values.
Expert identities are listed in Table~\ref{tab:models}.}
\label{fig:lambda-profile}
\end{figure}

\section{Ablation Studies}
\label{sec:ablations}

\subsection{Effect of Evaluation-Point Placement}
\label{sec:structured}

We compare lattice and random evaluation points using the same quadratic function class on the simplex.
For each budget $n$, we draw $n$ of the 150 interior points from Section~\ref{sec:fidelity} without replacement and fit ID AVG by unregularized least squares (LS) or ridge regression.
We repeat this procedure 1{,}000 times, evaluating all surrogates on the remaining $150-n$ points in each resample and keeping the 36-point lattice surrogate fixed.
Ridge strength is selected by cross-validation using only the fitting points; scaling and fitting details appear in Appendix~\ref{app:random-design-fits}.

At the matched budget of 36 evaluations, the lattice design achieves higher mean rank correlation and lower mean MAE than either random-design fit on both architectures (Table~\ref{tab:point_placement}).
Ridge reduces the random design's mean MAE from 33.64 to 2.90 on Gemma and from 30.15 to 2.24 on Llama, compared with approximately 1.93 and 1.07 for the lattice.
Among the tested budgets, the lattice's win rate first falls below 50\% at $n=60$ for Gemma and $n=105$ for Llama, against random designs fitted with either LS or ridge.

\begin{table}[t]
\centering
\small
\setlength{\tabcolsep}{2.3pt}
\begin{tabular}{@{}r rrrr @{\hspace{10pt}} rrrr@{}}
\toprule
 & \multicolumn{4}{c}{\textbf{Gemma-2-9B}} & \multicolumn{4}{c}{\textbf{Llama-3.1-8B}} \\
\cmidrule(lr){2-5}\cmidrule(lr){6-9}
$n$ & $\rho_{\mathrm{lattice}}$ & $\rho_{\mathrm{random}}$ & $\mathrm{MAE}_{\mathrm{random}}$ & Lattice win
    & $\rho_{\mathrm{lattice}}$ & $\rho_{\mathrm{random}}$ & $\mathrm{MAE}_{\mathrm{random}}$ & Lattice win \\
\midrule
 36 & 0.935 & 0.465\,/\,0.847 & 33.64\,/\,2.90 & 99.5\,/\,99.5 & 0.958 & 0.330\,/\,0.758 & 30.15\,/\,2.24 & 100.0\,/\,100.0 \\
 45 & 0.935 & 0.902\,/\,0.884 & 2.46\,/\,2.50 & 68.9\,/\,86.2 & 0.958 & 0.819\,/\,0.800 & 1.81\,/\,1.99 & 99.9\,/\,99.9 \\
 60 & 0.934 & 0.954\,/\,0.948 & 1.63\,/\,1.70 & 12.1\,/\,18.2 & 0.957 & 0.918\,/\,0.915 & 1.11\,/\,1.16 & 95.6\,/\,95.9 \\
 75 & 0.933 & 0.966\,/\,0.965 & 1.38\,/\,1.39 & 1.7\,/\,2.4 & 0.956 & 0.941\,/\,0.941 & 0.93\,/\,0.94 & 79.1\,/\,79.3 \\
 90 & 0.931 & 0.970\,/\,0.970 & 1.27\,/\,1.27 & 0.5\,/\,0.5 & 0.954 & 0.950\,/\,0.950 & 0.83\,/\,0.84 & 57.9\,/\,57.1 \\
105 & 0.930 & 0.972\,/\,0.972 & 1.19\,/\,1.19 & 0.5\,/\,0.6 & 0.953 & 0.953\,/\,0.954 & 0.80\,/\,0.80 & 44.2\,/\,43.8 \\
120 & 0.924 & 0.970\,/\,0.970 & 1.14\,/\,1.14 & 1.1\,/\,1.0 & 0.949 & 0.953\,/\,0.953 & 0.75\,/\,0.76 & 39.5\,/\,39.0 \\
\bottomrule
\end{tabular}
\caption{The lattice improves AVG prediction at the matched budget of 36 evaluations.
Random budgets are $n$; the lattice always uses 36.
Values average 1{,}000 resamples with shared held-out points; pairs show LS / ridge.
Lattice win is the percentage with $\rho_{\mathrm{lattice}}>\rho_{\mathrm{random}}$.
Lattice MAEs average approximately 1.93 (Gemma) and 1.07 (Llama).}
\label{tab:point_placement}
\label{tab:nsweep}
\end{table}

\subsection{Effect of Surrogate Optimization}
\label{sec:surrogate_selection}

The best lattice point is the lattice point with the highest measured ID AVG among the same 36 evaluations.
We compare it with quadratic SLIM in AVG and in the stability of the selected coefficients.
\paragraph{Differences in AVG.}
Relative to the best lattice point, SLIM changes ID AVG by $+0.8$ (Gemma) and $-0.8$ (Llama) points and OOD AVG by $-1.8$ and $+4.0$ points (Table~\ref{tab:selection_ablation}).
In the experiment of Table~\ref{tab:matched-budget}, SLIM and the best lattice point are evaluated on the same GPU node, and we compare them with an item-level paired bootstrap (Appendix~\ref{app:reb-protocol}, Table~\ref{tab:reb-avg-ci}).
SLIM minus the best lattice point, with 95\% intervals, is $+1.08$ $[-0.13, 2.35]$ (ID) and $-1.58$ $[-2.93, -0.14]$ (OOD) AVG points on Gemma and $-1.04$ $[-2.35, 0.33]$ (ID) and $+4.05$ $[2.53, 5.56]$ (OOD) on Llama.

\paragraph{Stability of the selected coefficients.}
We resample the benchmark items of the 36 lattice evaluations 1{,}000 times (Appendix~\ref{app:reb-stability}), refit SLIM and reselect the best lattice point in each resample $b$, and measure the $\ell_1$ distance $\|\lambda^{(b)}-\lambda^{(0)}\|_1=\sum_{k=1}^{T}|\lambda^{(b)}_k-\lambda^{(0)}_k|$ between the selected coefficients $\lambda^{(b)}$ and those selected with all items, $\lambda^{(0)}$.
Because both sum to one, this distance is twice the coefficient mass moved between experts, so distances above 0.1 and 0.3 mean that more than 0.05 (a small reweighting) and 0.15 (a substantial change) of the mass moved; replacing one expert of an equal-weight pair gives 1.0.
Table~\ref{tab:stability-main} shows the results: on Gemma, SLIM keeps its three experts in every resample, whereas the best lattice point switches between two expert pairs in 40.9\%.
On Llama, the mean distance is comparable, and SLIM makes small reweightings more often but substantial changes less often.

\begin{table}[t]
\noindent
\begin{minipage}[t]{0.535\textwidth}
\vspace{0pt}
\centering
\footnotesize
\setlength{\tabcolsep}{2.5pt}
\begin{tabular}{lrrrrr}
\toprule
 & & \multicolumn{2}{c}{Gemma-2-9B} & \multicolumn{2}{c}{Llama-3.1-8B} \\
\cmidrule(lr){3-4}\cmidrule(lr){5-6}
Variant & \#Evals & ID & OOD & ID & OOD \\
\midrule
Best vertex        & 8   & 58.51 & 49.75 & 54.80 & 46.50 \\
Best lattice point & 36  & 65.34 & \textbf{52.13} & \textbf{57.40} & 43.23 \\
SLIM (quadratic)   & 36  & 66.15 & 50.30 & 56.61 & \textbf{47.27} \\
\midrule
SLIM (cubic)       & 120 & \textbf{66.35} & 48.41 & 57.04 & 46.27 \\
\bottomrule
\end{tabular}
\caption{AVG of the selection variants. The first two rows select by measured ID AVG among eight vertices or 36 lattice points; SLIM optimizes its surrogate. \#Evals counts ID evaluations for selection; OOD is held out. Bold marks column maxima.}
\label{tab:selection_ablation}
\end{minipage}\hfill
\begin{minipage}[t]{0.445\textwidth}
\vspace{0pt}
\centering
\footnotesize
\setlength{\tabcolsep}{3pt}
\begin{tabular}{@{}lrrrr@{}}
\toprule
 & \multicolumn{2}{c}{Gemma-2-9B} & \multicolumn{2}{c}{Llama-3.1-8B} \\
\cmidrule(lr){2-3}\cmidrule(l){4-5}
 & SLIM & BL & SLIM & BL \\
\midrule
Experts changed & 0.0 & 40.9 & 25.8 & 7.8 \\
$\ell_1>0.1$ & 0.1 & 40.9 & 17.5 & 7.8 \\
$\ell_1>0.3$ & 0.0 & 40.9 & 4.0 & 7.8 \\
\midrule
Mean $\ell_1$ & 0.034 & 0.409 & 0.065 & 0.078 \\
\bottomrule
\end{tabular}

\caption{Stability of the selected coefficients over 1{,}000 item-level resamples (excerpt of Table~\ref{tab:reb-stability}): \% of resamples in which the selected experts change or the $\ell_1$ distance exceeds 0.1 or 0.3, and the mean distance (last row). BL: best lattice point.}
\label{tab:stability-main}
\end{minipage}
\end{table}


\subsection{Matched-Budget Comparison on the Simplex}
\label{sec:matched-budget}

To isolate the evaluation design from the coefficient domain and budget, we compare selection methods that share the eight experts, the simplex, and 36 ID evaluations (Table~\ref{tab:matched-budget}).
MAP and evolutionary merging run on the simplex, and we add three simple baselines (Appendix~\ref{app:coefficient-implementations}).
Random design + ridge fits the same quadratic class to 36 random interior mixtures and maximizes it like SLIM, differing in where the evaluations are placed and in the regularized fit that random placement requires (Section~\ref{sec:structured}).
GP-BO (Gaussian-process Bayesian optimization;~\citealp{snoek2012practical}) chooses its evaluations sequentially, and random best-of-36 selects the best of 36 random mixtures by measured ID AVG.

\paragraph{Comparison with the baselines.}
SLIM exceeds the three-seed means of evolutionary merging, MAP, and random best-of-36 in all four settings and that of random design + ridge in three, with paired 95\% intervals excluding zero (Appendix~\ref{app:reb-protocol}, Table~\ref{tab:reb-avg-ci}; Llama ID: $+0.37$ $[-0.85, 1.65]$); the intervals fix the seeds, omitting seed variability.

\paragraph{Comparison with GP-BO.}
SLIM, GP-BO, and the best lattice point are the top three in every setting; SLIM leads GP-BO on Gemma ID and Llama OOD ($+1.17$, $+4.47$) and trails it on Gemma OOD and Llama ID ($-2.14$, $-2.05$), all intervals excluding zero.
SLIM and GP-BO both select mixtures of two or three experts (Figure~\ref{fig:lambda-simplex36}); unlike GP-BO's 28 sequential queries, SLIM's evaluations can run in parallel, and its optimization takes under two seconds on a CPU.

\begin{table}[t]
\centering
\small
\setlength{\tabcolsep}{4pt}
\begin{tabular}{@{}lcrrrr@{}}
\toprule
 & & \multicolumn{2}{c}{Gemma-2-9B} & \multicolumn{2}{c}{Llama-3.1-8B} \\
\cmidrule(lr){3-4}\cmidrule(lr){5-6}
Method (36 evaluations on $\Delta^{T-1}$) & Runs & ID & OOD & ID & OOD \\
\midrule
SLIM (quadratic) & 1 & \textbf{66.43} & 50.56 & 56.37 & \textbf{47.28} \\
Best lattice point & 1 & 65.34 & 52.13 & 57.40 & 43.23 \\
\midrule
GP-BO & 3 & $65.25_{\pm0.58}$ & $\mathbf{52.69}_{\pm0.72}$ & $\mathbf{58.41}_{\pm1.24}$ & $42.81_{\pm1.89}$ \\
Random design + ridge & 3 & $64.73_{\pm1.76}$ & $47.11_{\pm2.91}$ & $56.00_{\pm1.50}$ & $41.48_{\pm1.01}$ \\
Evol.\ Merge (simplex) & 3 & $60.88_{\pm0.92}$ & $46.41_{\pm2.10}$ & $50.66_{\pm1.11}$ & $38.82_{\pm1.28}$ \\
MAP (simplex) & 3 & $58.85_{\pm5.89}$ & $44.00_{\pm5.45}$ & $48.97_{\pm7.24}$ & $36.61_{\pm4.65}$ \\
Random best-of-36 & 3 & $59.25_{\pm0.81}$ & $44.23_{\pm2.29}$ & $50.44_{\pm0.18}$ & $38.88_{\pm1.31}$ \\
\bottomrule
\end{tabular}

\caption{AVG of methods sharing the eight experts ($T=8$), the simplex, and 36 ID evaluations. All models of an architecture are evaluated on the same GPU node for a fair comparison, so SLIM's values differ by at most 0.28 points from Table~\ref{tab:selection_ablation}, measured on another GPU cluster (Appendix~\ref{app:reb-protocol}). Seeded methods show mean $\pm$ standard deviation (seeds 42/43/44). Bold marks column maxima; Tables~\ref{tab:reb-avg-ci}--\ref{tab:reb-domain-ci} give intervals and all domain scores.}
\label{tab:matched-budget}
\end{table}

\section{Discussion and Limitations}
\label{sec:discussion}

\paragraph{Sufficiency of a Quadratic Surrogate.}
Evaluations of individual experts and equal-weight pairs support AVG prediction at mixtures of all eight experts on both architectures (Figure~\ref{fig:fidelity}), although domain-wise fidelity varies (Table~\ref{tab:fidelity_domain}).
Increasing the degree from quadratic to cubic raises the budget from 36 to 120 evaluations with the same simplex-constrained optimization (Appendix~\ref{app:cubic-extension}) but produces small ID gains and lower OOD AVG on both architectures (Table~\ref{tab:selection_ablation}; Figure~\ref{fig:cost_perf}).
These results support quadratic approximation in the tested settings and show limited benefit from increasing the degree.
When greater surrogate expressiveness improves generalization remains an open question.

\paragraph{Evaluation Design as the Main Source of Gains.}
At 36 evaluations, the lattice yields higher mean rank correlation and lower MAE than random designs, including ridge-regularized fits (Table~\ref{tab:point_placement}), and at a matched budget SLIM and the best lattice point both exceed random design + ridge, evolutionary merging, MAP, and random best-of-36 (Section~\ref{sec:matched-budget}).
Evaluation placement therefore accounts for most of SLIM's advantage and matters even with the surrogate class, coefficient domain, and budget fixed.
The $T(T+1)/2$ count is the minimum for identifying a general quadratic on the simplex, not a universal lower bound for finding a good merge.

\paragraph{The Net Value of Optimization.}
Most ID improvement over the best individual expert comes from adding equal-weight pairs to the candidates (Table~\ref{tab:selection_ablation}).
Optimizing beyond the best lattice point leaves ID AVG unchanged within uncertainty, lowering OOD AVG on Gemma and raising it on Llama (Section~\ref{sec:surrogate_selection}).
The surrogate enables continuous coefficient selection without further target-metric evaluations; its additional performance benefit depends on the setting.
It also stabilizes the selection on Gemma (Table~\ref{tab:stability-main}), predicts the sparse mixtures where solutions lie better than random design + ridge (Appendix~\ref{app:reb-sparse}), and splits the four settings with GP-BO without sequential queries (Section~\ref{sec:matched-budget}).

\paragraph{The Simplex: Predictability versus Scale.}
The simplex contains competitive merges that a quadratic surrogate locates from 36 evaluations and that generalize beyond the selection benchmarks (Figure~\ref{fig:cost_perf}).
SLIM concentrates weight on instruction and safety experts, whereas on Llama ID, MAP and evolutionary merging select candidates scoring higher at 135 and 200 evaluations, with coefficient sums of 3.31--3.75 and 2.50--2.83 (Figure~\ref{fig:lambda-profile}; Table~\ref{tab:main}), differing in allocation and scale.
With evolutionary merging (CMA-ES) and 36 evaluations fixed, restricting the search from $[0,1]^8$ to the simplex raises Gemma's OOD AVG by 4.3--10.8 points but lowers Llama's ID and OOD AVG by 7.3--8.4 and 1.0--2.9 points in all three seeds, so the domain effect differs between architectures (Table~\ref{tab:reb-e11}).
Scaling SLIM's coefficients as $s\widehat{\lambda}$ with $s\in\{1,1.5,\ldots,3.5\}$ chosen on ID AVG (six evaluations, including $s=1$) selects $s=1.5$ on Gemma, raising OOD AVG from 50.56 to 53.65 on an ID margin of only 0.43 (point estimates), and $s=1$ on Llama (Appendix~\ref{app:reb-scale}).
Although only the rescaling isolates the effect of coefficient scale, these results motivate varying both mixture proportions and overall task-vector scale while controlling the cost of surrogate construction.

\paragraph{Limitations.}
Each architecture uses one set of eight experts; AVG prediction remains comparable on subsets of five to seven (Spearman's $\rho$ of 0.88--0.98; Appendix~\ref{app:reb-subsets}), but larger sets are untested, and merging scaling laws, measured with cross-entropy loss and equal-weight merging, find that most of the gain from nine experts arrives with five or six~\citep{wang-2026-merging-scaling-law}.
At its maximizer, the surrogate overestimates AVG by 2.17 (Gemma) and 1.41 (Llama) points (Appendix~\ref{app:reb-local}).
Safety scores are refusal rates on harmful prompts; on Llama, SLIM's OOD lead relies on safety, but penalizing over-refusal leaves its rank among the 36-evaluation methods unchanged (Appendix~\ref{app:reb-safety}).

\section{Conclusion}
\label{sec:conclusion}

SLIM selects LLM merging coefficients using a quadratic surrogate built from individual experts and equal-weight pairs, with the minimum evaluations required for quadratic identification on the simplex.
Experiments on two architectures show accurate prediction of aggregate performance at unseen mixtures and competitive merging under limited evaluation budgets.
Comparisons at a matched budget highlight evaluation-point placement as an important design choice for surrogate-based coefficient selection.

\clearpage
\subsection*{AI use statement}
We used generative AI tools (ChatGPT, Claude, and Claude Code) for research
ideation and discussion, literature search, writing and debugging code for
experiments and figures, running experiments and analyzing results, drafting
the manuscript, English editing and paraphrasing, creating figures and tables,
and checking the mathematical derivations in Appendix~\ref{app:surrogate-details}.
We have reviewed all AI-assisted work. LLM-generated code was tested by the
authors, and the resulting numerical values were verified by manual
calculation or against an independent implementation. All references
identified with AI assistance were checked against the original sources.
We take responsibility for the final content of this work, including text,
claims, and artifacts produced with the aid of generative AI.

\subsection*{Ethics statement}
This work does not involve human subjects, personally identifiable
information, or newly collected data. We are not aware of ethical concerns
raised by this work beyond those associated with the publicly available
models and benchmarks it uses.

\subsection*{Reproducibility statement}
Section~\ref{sec:method} and Algorithm~\ref{alg:slim} describe SLIM, and
Appendices~~\ref{app:interpolant-derivation}--~\ref{app:cubic-extension} derive the quadratic interpolant, its uniqueness, its
equivalence under domain aggregation, and the cubic extension.
Appendix~\ref{app:models} (Table~\ref{tab:models}) lists all base and expert checkpoints;
Appendix~\ref{app:benchmarks} (Table~\ref{tab:benchmark-details}) specifies the benchmarks, metrics, sample counts, and
prompting settings; and Appendix~\ref{app:coefficient-implementations} details the implementations of SLIM and
all baselines, including random seeds, optimizer settings, and
initializations. Upon acceptance, we will release our code together with the
measured scores used in our experiments, including those at the lattice
design points and the 150 held-out mixtures. These measurements allow the
surrogate-fidelity results and SLIM's coefficient selection to be reproduced
without re-evaluating merged models.

\subsubsection*{Acknowledgments}
This work was supported by JST SPRING (Grant Number JPMJSP2108) and JSPS KAKENHI (Grant Number 23H04974), and in part by KIOXIA Corporation.

\bibliography{iclr2027_conference}

@article{song2026model,
  title={Model Merging in the Era of Large Language Models: Methods, Applications, and Future Directions},
  author={Song, Mingyang and Zheng, Mao},
  journal={arXiv preprint arXiv:2603.09938},
  year={2026}
}

@inproceedings{ilharco2023editing,
title={Editing models with task arithmetic},
author={Gabriel Ilharco and Marco Tulio Ribeiro and Mitchell Wortsman and Ludwig Schmidt and Hannaneh Hajishirzi and Ali Farhadi},
booktitle={The Eleventh International Conference on Learning Representations },
year={2023},
url={https://openreview.net/forum?id=6t0Kwf8-jrj}
}

@inproceedings{wortsman2022model,
  title={Model soups: averaging weights of multiple fine-tuned models improves accuracy without increasing inference time},
  author={Wortsman, Mitchell and Ilharco, Gabriel and Gadre, Samir Ya and Roelofs, Rebecca and Gontijo-Lopes, Raphael and Morcos, Ari S and Namkoong, Hongseok and Farhadi, Ali and Carmon, Yair and Kornblith, Simon and others},
  booktitle={International conference on machine learning},
  pages={23965--23998},
  year={2022},
  organization={PMLR}
}

@article{yadav2023ties,
  title={Ties-merging: Resolving interference when merging models},
  author={Yadav, Prateek and Tam, Derek and Choshen, Leshem and Raffel, Colin A and Bansal, Mohit},
  journal={Advances in neural information processing systems},
  volume={36},
  pages={7093--7115},
  year={2023}
}

@inproceedings{yu2024language,
  title={Language models are super mario: Absorbing abilities from homologous models as a free lunch},
  author={Yu, Le and Yu, Bowen and Yu, Haiyang and Huang, Fei and Li, Yongbin},
  booktitle={Forty-first International Conference on Machine Learning},
  year={2024}
}

@article{AdaMerging_ICLR_2024,
  title={AdaMerging: Adaptive Model Merging for Multi-Task Learning},
  author={Yang, Enneng and Wang, Zhenyi and Shen, Li and Liu, Shiwei and Guo, Guibing and Wang, Xingwei and Tao, Dacheng},
  journal={The Twelfth International Conference on Learning Representations},
  year={2024}
}

@inproceedings{zhou2024metagpt,
  title={Metagpt: Merging large language models using model exclusive task arithmetic},
  author={Zhou, Yuyan and Song, Liang and Wang, Bingning and Chen, Weipeng},
  booktitle={Proceedings of the 2024 Conference on Empirical Methods in Natural Language Processing},
  pages={1711--1724},
  year={2024}
}

@article{akiba2025evolutionary,
  title={Evolutionary optimization of model merging recipes},
  author={Akiba, Takuya and Shing, Makoto and Tang, Yujin and Sun, Qi and Ha, David},
  journal={Nature Machine Intelligence},
  volume={7},
  number={2},
  pages={195--204},
  year={2025},
  publisher={Nature Publishing Group UK London}
}

@inproceedings{li2024map,
 author = {Li, Lu and Zhang, Tianyu and Bu, Zhiqi and Wang, Suyuchen and He, Huan and Fu, Jie and Wu, Yonghui and Bian, Jiang and Chen, Yong and Bengio, Yoshua},
 booktitle = {International Conference on Learning Representations},
 editor = {Y. Yue and A. Garg and N. Peng and F. Sha and R. Yu},
 pages = {65032--65064},
 title = {MAP: Low-compute Model Merging with Amortized Pareto Fronts via Quadratic Approximation},
 url = {https://proceedings.iclr.cc/paper_files/paper/2025/file/a3724221b623bf1966e92cd266ac6177-Paper-Conference.pdf},
 volume = {2025},
 year = {2025}
}

@ARTICLE{li2025morphing,
  author={Li, Bingdong and Di, Zixiang and Yang, Yanting and Qian, Hong and Yang, Peng and Hao, Hao and Tang, Ke and Zhou, Aimin},
  journal={IEEE Transactions on Evolutionary Computation}, 
  title={It’s Morphing Time: Unleashing the Potential of Multiple LLMs via Multiobjective Optimization}, 
  year={2026},
  volume={30},
  number={4},
  pages={1699-1713},
  doi={10.1109/TEVC.2025.3613937}}

@article{su2025fine,
  title={Fine, I'll Merge It Myself: A Multi-Fidelity Framework for Automated Model Merging},
  author={Su, Guinan and Geiping, Jonas},
  journal={arXiv preprint arXiv:2502.04030},
  year={2025}
}

@article{snoek2012practical,
  title={Practical bayesian optimization of machine learning algorithms},
  author={Snoek, Jasper and Larochelle, Hugo and Adams, Ryan P},
  journal={Advances in neural information processing systems},
  volume={25},
  year={2012}
}

@article{yang2026model,
  title={Model Merging in LLMs, MLLMs, and Beyond: Methods, Theories, Applications, and Opportunities},
  author={Yang, Enneng and Shen, Li and Guo, Guibing and Wang, Xingwei and Cao, Xiaochun and Zhang, Jie and Tao, Dacheng},
  journal={ACM Comput. Surv.},
  volume={58},
  number={8},
  pages={216--257},
  year={2026}
}

@article{cobbe2021training,
  title={Training verifiers to solve math word problems},
  author={Cobbe, Karl and Kosaraju, Vineet and Bavarian, Mohammad and Chen, Mark and Jun, Heewoo and Kaiser, Lukasz and Plappert, Matthias and Tworek, Jerry and Hilton, Jacob and Nakano, Reiichiro and others},
  journal={arXiv preprint arXiv:2110.14168},
  year={2021}
}

@article{hendrycks2021measuring,
  title={Measuring mathematical problem solving with the math dataset},
  author={Hendrycks, Dan and Burns, Collin and Kadavath, Saurav and Arora, Akul and Basart, Steven and Tang, Eric and Song, Dawn and Steinhardt, Jacob},
  journal={arXiv preprint arXiv:2103.03874},
  year={2021}
}

@article{austin2021program,
  title={Program synthesis with large language models},
  author={Austin, Jacob and Odena, Augustus and Nye, Maxwell and Bosma, Maarten and Michalewski, Henryk and Dohan, David and Jiang, Ellen and Cai, Carrie and Terry, Michael and Le, Quoc and others},
  journal={arXiv preprint arXiv:2108.07732},
  year={2021}
}

@article{zhou2023instruction,
  title={Instruction-following evaluation for large language models},
  author={Zhou, Jeffrey and Lu, Tianjian and Mishra, Swaroop and Brahma, Siddhartha and Basu, Sujoy and Luan, Yi and Zhou, Denny and Hou, Le},
  journal={arXiv preprint arXiv:2311.07911},
  year={2023}
}

@article{dubois2024length,
  title={Length-controlled alpacaeval: A simple way to debias automatic evaluators},
  author={Dubois, Yann and Galambosi, Bal{\'a}zs and Liang, Percy and Hashimoto, Tatsunori B},
  journal={arXiv preprint arXiv:2404.04475},
  year={2024}
}

@article{chen2021evaluating,
  title={Evaluating large language models trained on code},
  author={Chen, Mark and Tworek, Jerry and Jun, Heewoo and Yuan, Qiming and Pinto, Henrique Ponde De Oliveira and Kaplan, Jared and Edwards, Harri and Burda, Yuri and Joseph, Nicholas and Brockman, Greg and others},
  journal={arXiv preprint arXiv:2107.03374},
  year={2021}
}

@article{scheffe1958experiments,
  title={Experiments with mixtures},
  author={Scheff{\'e}, Henry},
  journal={Journal of the Royal Statistical Society: Series B (Methodological)},
  volume={20},
  number={2},
  pages={344--360},
  year={1958},
  publisher={Wiley Online Library}
}

@article{kiefer1961optimum,
  title={Optimum designs in regression problems, II},
  author={Kiefer, J},
  journal={The Annals of Mathematical Statistics},
  volume={32},
  number={1},
  pages={298--325},
  year={1961},
  publisher={Institute of Mathematical Statistics}
}

@article{mao2026datamixing,
  title={Data Mixing as Mixture Experiment: Response Surface Methodology and Optimal Design for Large Language Model Pretraining},
  author={Mao, Yicheng and Du, Hongru},
  journal={arXiv preprint arXiv:2608.23922},
  year={2026}
}

@book{cornell2002experiments,
  author    = {Cornell, John A.},
  title     = {Experiments with Mixtures: Designs, Models, and the Analysis of Mixture Data},
  edition   = {3rd},
  publisher = {John Wiley \& Sons},
  year      = {2002}
}

@inproceedings{burchell2024flores,
  author    = {Burchell, Laurie and Maillard, Jean and Anastasopoulos, Antonios and Federmann, Christian and Koehn, Philipp and Wang, Skyler},
  title     = {Findings of the {WMT} 2024 Shared Task of the Open Language Data Initiative},
  booktitle = {Proceedings of the Ninth Conference on Machine Translation},
  pages     = {110--117},
  year      = {2024}
}

@article{nllb2024scaling,
  author  = {{NLLB Team}},
  title   = {Scaling Neural Machine Translation to 200 Languages},
  journal = {Nature},
  volume  = {630},
  pages   = {841--846},
  year    = {2024},
  doi     = {https://doi.org/10.1038/s41586-024-07335-x}
}

@InProceedings{mazeika2024harmbench,
  title = 	 {{H}arm{B}ench: A Standardized Evaluation Framework for Automated Red Teaming and Robust Refusal},
  author =       {Mazeika, Mantas and Phan, Long and Yin, Xuwang and Zou, Andy and Wang, Zifan and Mu, Norman and Sakhaee, Elham and Li, Nathaniel and Basart, Steven and Li, Bo and Forsyth, David and Hendrycks, Dan},
  booktitle = 	 {Proceedings of the 41st International Conference on Machine Learning},
  pages = 	 {35181--35224},
  year = 	 {2024},
  editor = 	 {Salakhutdinov, Ruslan and Kolter, Zico and Heller, Katherine and Weller, Adrian and Oliver, Nuria and Scarlett, Jonathan and Berkenkamp, Felix},
  volume = 	 {235},
  series = 	 {Proceedings of Machine Learning Research},
  month = 	 {21--27 Jul},
  publisher =    {PMLR},
  url = 	 {https://proceedings.mlr.press/v235/mazeika24a.html}
}

@article{han2024wildguard,
  title={Wildguard: Open one-stop moderation tools for safety risks, jailbreaks, and refusals of llms},
  author={Han, Seungju and Rao, Kavel and Ettinger, Allyson and Jiang, Liwei and Lin, Bill Yuchen and Lambert, Nathan and Choi, Yejin and Dziri, Nouha},
  journal={Advances in neural information processing systems},
  volume={37},
  pages={8093--8131},
  year={2024}
}

@inproceedings{mencattini2025merge3,
  title     = {MERGE$^3$: Efficient Evolutionary Merging on Consumer-grade GPUs},
  author    = {Mencattini, Tommaso and Minut, Adrian Robert and Crisostomi, Donato and Santilli, Andrea and Rodolà, Emanuele},
  year      = 2025,
  booktitle = {Proceedings of the 42nd International Conference on Machine Learning (ICML)},
  url       = {https://openreview.net/pdf?id=qFXDv0X4yc},
}

@article{team2024gemma,
  title={Gemma 2: Improving open language models at a practical size},
  author={{Gemma Team} and Riviere, Morgane and Pathak, Shreya and Sessa, Pier Giuseppe and Hardin, Cassidy and Bhupatiraju, Surya and Hussenot, L{\'e}onard and Mesnard, Thomas and Shahriari, Bobak and Ram{\'e}, Alexandre and others},
  journal={arXiv preprint arXiv:2408.00118},
  year={2024}
}

@article{grattafiori2024llama,
  title={The llama 3 herd of models},
  author={Grattafiori, Aaron and Dubey, Abhimanyu and Jauhri, Abhinav and Pandey, Abhinav and Kadian, Abhishek and Al-Dahle, Ahmad and Letman, Aiesha and Mathur, Akhil and Schelten, Alan and Vaughan, Alex and others},
  journal={arXiv preprint arXiv:2407.21783},
  year={2024}
}

@inproceedings{
    he2025mergebench,
    title={MergeBench: A Benchmark for Merging Domain-Specialized {LLM}s},
    author={Yifei He and Siqi Zeng and Yuzheng Hu and Rui Yang and Tong Zhang and Han Zhao},
    booktitle={The Thirty-ninth Annual Conference on Neural Information Processing Systems Datasets and Benchmarks Track},
    year={2025},
    url={https://openreview.net/forum?id=rw50iUoyLu}
}

@inproceedings{lightman2024let,
  title={Let's verify step by step},
  author={Lightman, Hunter and Kosaraju, Vineet and Burda, Yuri and Edwards, Harrison and Baker, Bowen and Lee, Teddy and Leike, Jan and Schulman, John and Sutskever, Ilya and Cobbe, Karl},
  booktitle={International Conference on Learning Representations},
  volume={2024},
  pages={39578--39601},
  year={2024}
}

@inproceedings{wang-2026-merging-scaling-law,
      title={Model Merging Scaling Laws in Large Language Models},
      author={Yuanyi Wang and Yanggan Gu and Yiming Zhang and Qi Zhou and Zhaoyi Yan and Congkai Xie and Xinyao Wang and Jianbo Yuan and Hongxia Yang},
      booktitle={Proceedings of the 43rd International Conference on Machine Learning},
      year={2026},
      url={https://github.com/InfiXAI/Merging-Scaling-Law},
}
\bibliographystyle{iclr2027_conference}

\clearpage
\appendix

\startcontents[appendix]
\section*{Appendix Contents}
\printcontents[appendix]{}{1}{\setcounter{tocdepth}{2}}
\clearpage

\section{Notation}
\label{app:notation}

\begin{table}[htbp]
\centering
\small
\renewcommand{\arraystretch}{1.12}
\setlength{\tabcolsep}{5pt}
\begin{tabular}{@{}p{0.23\linewidth}p{0.73\linewidth}@{}}
\toprule
Symbol & Meaning \\
\midrule
$T$
& Number of expert models. \\

$\theta_{\mathrm{pre}},\theta_k$
& Parameters of the shared base model and expert $k$. \\

$\tau_k$
& Task vector $\theta_k-\theta_{\mathrm{pre}}$. \\

$\lambda,\lambda_k$
& Merging coefficient vector and its $k$-th component. \\

$\theta(\lambda)$
& Merged parameters
$\theta_{\mathrm{pre}}+\sum_k\lambda_k\tau_k$. \\

$\Lambda$
& General feasible set of merging coefficients. \\

$\Delta^{T-1}$
& Coefficient simplex:
$\lambda_k\geq0$ and $\sum_k\lambda_k=1$. \\

$e_k$
& $k$-th standard basis vector; the vertex corresponding to expert $k$. \\

$\mathcal{D}$
& Evaluation design consisting of the vertices and edge midpoints. \\

\addlinespace
$M(\lambda)$
& Measured target metric of $\theta(\lambda)$; AVG in the experiments. \\

$M_k,M_{ij}$
& Measurements $M(e_k)$ and $M((e_i+e_j)/2)$. \\

$\widehat{M}(\lambda)$
& Quadratic interpolant constructed from measurements on $\mathcal{D}$. \\

$q,\beta_k,\beta_{ij}$
& A quadratic in canonical mixture form and its linear and pairwise coefficients. \\

$\varphi_a(\lambda)$
& Interpolation basis function associated with design point $a\in\mathcal{D}$. \\

$\lambda^\star$
& A maximizer of $M$ over $\Lambda$. \\

$\widehat{\lambda}$
& Coefficients returned by numerical optimization of
$\widehat{M}$ over $\Delta^{T-1}$. \\

\addlinespace
$D,M_d,w_d$
& Number of domains, domain-$d$ metric, and its fixed aggregation weight;
$w_d=1/D$ for AVG. \\

$\widehat{M}_d$
& Interpolant of domain-$d$ measurements on $\mathcal{D}$. \\

$n$
& Number of evaluation points used by MAP or a random-design surrogate. \\

\addlinespace
$\lambda^{(b)},\lambda^{(0)}$
& Coefficients selected in item-resampling replicate $b$ and with all items (Section~\ref{sec:surrogate_selection}; Appendix~\ref{app:reb-stability}). \\

SLIM$_{\mathrm{R}}$
& SLIM refit on the 36 lattice points evaluated on the GPU node (Appendix~\ref{app:reb-stability}). \\

$\mathcal{S}_5^A,\mathcal{S}_5^B,\mathcal{S}_6,\mathcal{S}_7$
& Expert subsets of Appendix~\ref{app:reb-subsets}. \\

$\epsilon$
& Weight moved toward a vertex for the local validation points of Appendix~\ref{app:reb-local}. \\

$s$
& Global scale of SLIM's coefficients, $\lambda=s\widehat{\lambda}$ (Appendix~\ref{app:reb-scale}). \\

$R_h,R_b$
& Refusal rates on harmful and benign prompts (Appendix~\ref{app:reb-overrefusal}). \\

$S_{\mathrm{bal}}$
& Balanced safety score $(R_h+100-R_b)/2$ (Appendix~\ref{app:reb-overrefusal}). \\

AVG$_{\mathrm{bal}}$, AVG$_4$
& OOD AVG with the safety score replaced by $S_{\mathrm{bal}}$ or removed (Appendix~\ref{app:reb-overrefusal}). \\

$w_s$
& Weight of the safety domain in the selection objective of Appendix~\ref{app:reb-ws}; $w_s=1$ gives AVG. \\
\bottomrule
\end{tabular}
\caption{Symbols defining the merging coefficients, target metric, and surrogate construction; the last block lists symbols of the additional analyses in Section~\ref{sec:surrogate_selection} and Appendices~\ref{app:reb-surrogate}--\ref{app:reb-safety}.}
\label{tab:notation}
\end{table}

\section{Additional Details on the Methods}
\label{app:surrogate-details}

\subsection{Derivation and Uniqueness}
\label{app:interpolant-derivation}

On the simplex, $1=\sum_k\lambda_k$ and
$\lambda_k^2=\lambda_k-\sum_{j\ne k}\lambda_k\lambda_j$.
Thus, constant and squared terms can be absorbed into linear
and pairwise terms, giving the canonical mixture form in
Eq.~\ref{eq:quadratic-mixture}~\citep{scheffe1958experiments}.

At each vertex, interpolation requires
$\beta_k=q(e_k)=M_k$.
At the midpoint of experts $i$ and $j$,
\begin{equation}
q\left(\frac{e_i+e_j}{2}\right)
=\frac{\beta_i+\beta_j}{2}+\frac{\beta_{ij}}{4}
=M_{ij},
\label{eq:midpoint-condition}
\end{equation}
so
\begin{equation}
\beta_{ij}=4M_{ij}-2M_i-2M_j.
\label{eq:pairwise-coefficient}
\end{equation}
The vertex and midpoint values uniquely determine all
$T+\binom{T}{2}$ coefficients.
Conversely, each evaluation supplies one linear constraint
on these coefficients, so fewer evaluations cannot identify
an unrestricted member of this quadratic space.

Substituting $\beta_k=M_k$ and
Eq.~\ref{eq:pairwise-coefficient}
into Eq.~\ref{eq:quadratic-mixture} gives
\begin{align*}
q(\lambda)
&=\sum_k M_k\lambda_k
  +\sum_{i<j}(4M_{ij}-2M_i-2M_j)\lambda_i\lambda_j\\
&=\sum_k M_k\lambda_k
  +\sum_{i<j}4M_{ij}\lambda_i\lambda_j
  -2\sum_{i<j}(M_i+M_j)\lambda_i\lambda_j\\
&=\sum_k M_k\lambda_k+\sum_{i<j}4M_{ij}\lambda_i\lambda_j-2\left(\sum_{i<j}M_i\lambda_i\lambda_j+\sum_{i<j}M_j\lambda_i\lambda_j\right).\\
\intertext{
Regrouping the pairwise terms by expert index $k$ gives
}
q(\lambda)
&=\sum_k M_k\lambda_k+\sum_{i<j}4M_{ij}\lambda_i\lambda_j-2\sum_k M_k\lambda_k\left(\sum_{\ell>k}\lambda_\ell+\sum_{\ell<k}\lambda_\ell\right)\\
&=\sum_k M_k\lambda_k
  +\sum_{i<j}4M_{ij}\lambda_i\lambda_j
  -2\sum_k M_k\lambda_k
    \sum_{\ell\neq k}\lambda_\ell.\\
\intertext{
The simplex constraint implies
$\sum_{\ell\neq k}\lambda_\ell
=\sum_\ell \lambda_\ell-\lambda_k
=1-\lambda_k$.
Therefore,
}
q(\lambda)
&=\sum_k M_k\lambda_k
  +\sum_{i<j}4M_{ij}\lambda_i\lambda_j
  -2\sum_k M_k\lambda_k(1-\lambda_k)\\
&=\sum_k \lambda_k(2\lambda_k-1)M_k
  +\sum_{i<j}4\lambda_i\lambda_jM_{ij},
\end{align*}
which is Eq.~\ref{eq:interpolant}.

\subsection{Linearity under Domain Aggregation}
\label{app:aggregation}

For domain metrics $M_d$ and fixed weights $w_d$, define
\begin{equation}
M(\lambda)=\sum_{d=1}^{D}w_dM_d(\lambda).
\label{eq:aggregate-metric}
\end{equation}
AVG corresponds to $w_d=1/D$.
Let $\varphi_a$ denote the basis function in
Eq.~\ref{eq:interpolant} associated with design point
$a\in\mathcal{D}$.
Since these functions do not depend on the measured values,
\begin{align*}
\widehat{M}(\lambda)
&=\sum_{a\in\mathcal{D}}\varphi_a(\lambda)
  \sum_{d=1}^{D}w_dM_d(a)\\
&=\sum_{d=1}^{D}w_d
  \sum_{a\in\mathcal{D}}\varphi_a(\lambda)M_d(a)
=\sum_{d=1}^{D}w_d\widehat{M}_d(\lambda).
\end{align*}

Thus, interpolating aggregated measurements and aggregating
domain-wise interpolants give the same surrogate.
No additional model evaluations are needed when the domain
scores are recorded at the same design points.

\subsection{SLIM Algorithm}
\label{app:algorithm}

\begin{algorithm}[H]
\caption{SLIM: Simplex-Lattice Interpolation Merging}
\label{alg:slim}
\begin{algorithmic}[1]
\Require Base parameters $\theta_{\mathrm{pre}}$;
task vectors $\{\tau_k\}_{k=1}^{T}$; target metric $M$
\Ensure Merge coefficients $\widehat{\lambda}$
\State Construct $\mathcal{D}$ using Eq.~\ref{eq:design}
\ForAll{$\lambda\in\mathcal{D}$}
    \State Form $\theta(\lambda)$ using Eq.~\ref{eq:merge}
    \State Evaluate the model and store $M(\lambda)$
\EndFor
\State Construct $\widehat{M}$ using Eq.~\ref{eq:interpolant}
\State Numerically maximize $\widehat{M}$ over
$\Delta^{T-1}$ to obtain $\widehat{\lambda}$
\State \Return $\widehat{\lambda}$
\end{algorithmic}
\end{algorithm}

\subsection{Cubic Extension for the Degree Ablation}
\label{app:cubic-extension}

For the polynomial-degree ablation, we construct a cubic interpolant on the same coefficient simplex using the $\{T,3\}$ simplex-lattice design~\citep{scheffe1958experiments}:
\begin{equation}
\mathcal{A}_3
=\left\{\alpha\in\mathbb{Z}_{\geq0}^{T}\,\middle|\,\sum_{k=1}^{T}\alpha_k=3\right\},
\qquad
\mathcal{D}_3
=\left\{\frac{\alpha}{3}\,\middle|\,\alpha\in\mathcal{A}_3\right\}.
\label{eq:cubic-design}
\end{equation}

Figure~\ref{fig:simplex-lattice} illustrates the quadratic and cubic designs for $T=3$.

The design contains the vertices, two points per expert pair with weights $(2/3,1/3)$ and $(1/3,2/3)$, and the equal-weight mixture of each expert triple.
Its size is
\begin{equation}
|\mathcal{D}_3|
=T+2\binom{T}{2}+\binom{T}{3}
=\binom{T+2}{3},
\label{eq:cubic-design-size}
\end{equation}
giving $8+56+56=120$ evaluations for $T=8$.

Using the measured ID AVG at each design point, we construct
\begin{equation}
\begin{aligned}
\widehat{M}_3(\lambda)
&=\sum_{\alpha\in\mathcal{A}_3}
  \varphi^{(3)}_{\alpha}(\lambda)\,M(\alpha/3),\\
\varphi^{(3)}_{\alpha}(\lambda)
&=\prod_{k=1}^{T}\prod_{\ell=0}^{\alpha_k-1}
  \frac{3\lambda_k-\ell}{\alpha_k-\ell},
\end{aligned}
\label{eq:cubic-interpolant}
\end{equation}
where an empty product equals one.

At a design point $\gamma/3$, the basis function equals one if $\gamma=\alpha$.
Otherwise, some $\gamma_k<\alpha_k$, making a factor in the product zero.
Thus, $\widehat{M}_3$ reproduces all measured values.
The number of cardinal basis functions equals the dimension $\binom{T+2}{3}$ of polynomials of degree at most three on the simplex, so the interpolant is unique in this space.

We numerically maximize $\widehat{M}_3$ over $\Delta^{T-1}$ using the same SLSQP settings and initializations as quadratic SLIM (Appendix~\ref{app:coefficient-implementations}).
This optimization requires no further model evaluations, leaving the budget at $|\mathcal{D}_3|=120$ for $T=8$.

\section{Additional Details on Experiments}
\label{app:experiments}

\subsection{Models and Checkpoints}
\label{app:models}

Table~\ref{tab:models} lists the checkpoints used for Gemma-2-9B~\citep{team2024gemma} and Llama-3.1-8B~\citep{grattafiori2024llama}.
Each expert set contains five domain-specific checkpoints from MergeBench~\citep{he2025mergebench} and three additional checkpoints: Instruction B, Math B, and Code B.
Instruction B is the official instruction-tuned model.

\begin{table}[t]
\centering
\small
\setlength{\tabcolsep}{5pt}
\begin{tabular}{@{}ll@{}}
\toprule
Role & Hugging Face identifier \\
\midrule
\multicolumn{2}{l}{\emph{Gemma-2-9B}} \\
Base & \href{https://huggingface.co/google/gemma-2-9b}{\nolinkurl{google/gemma-2-9b}} \\
Instruction A & \href{https://huggingface.co/MergeBench/gemma-2-9b_instruction}{\nolinkurl{MergeBench/gemma-2-9b_instruction}} \\
Instruction B & \href{https://huggingface.co/google/gemma-2-9b-it}{\nolinkurl{google/gemma-2-9b-it}} \\
Math A & \href{https://huggingface.co/MergeBench/gemma-2-9b_math}{\nolinkurl{MergeBench/gemma-2-9b_math}} \\
Math B & \href{https://huggingface.co/POME-OPTIM/Gemma2-9B-MetaMathQA}{\nolinkurl{POME-OPTIM/Gemma2-9B-MetaMathQA}} \\
Code A & \href{https://huggingface.co/MergeBench/gemma-2-9b_coding}{\nolinkurl{MergeBench/gemma-2-9b_coding}} \\
Code B & \href{https://huggingface.co/testmoto/gemma-2-9b-synthetic_coding}{\nolinkurl{testmoto/gemma-2-9b-synthetic_coding}} \\
Multilingual & \href{https://huggingface.co/MergeBench/gemma-2-9b_multilingual}{\nolinkurl{MergeBench/gemma-2-9b_multilingual}} \\
Safety & \href{https://huggingface.co/MergeBench/gemma-2-9b_safety}{\nolinkurl{MergeBench/gemma-2-9b_safety}} \\
\midrule
\multicolumn{2}{l}{\emph{Llama-3.1-8B}} \\
Base & \href{https://huggingface.co/meta-llama/Llama-3.1-8B}{\nolinkurl{meta-llama/Llama-3.1-8B}} \\
Instruction A & \href{https://huggingface.co/MergeBench/Llama-3.1-8B_instruction}{\nolinkurl{MergeBench/Llama-3.1-8B_instruction}} \\
Instruction B & \href{https://huggingface.co/meta-llama/Llama-3.1-8B-Instruct}{\nolinkurl{meta-llama/Llama-3.1-8B-Instruct}} \\
Math A & \href{https://huggingface.co/MergeBench/Llama-3.1-8B_math}{\nolinkurl{MergeBench/Llama-3.1-8B_math}} \\
Math B & \href{https://huggingface.co/pmahdavi/Llama-3.1-8B-math-reasoning}{\nolinkurl{pmahdavi/Llama-3.1-8B-math-reasoning}} \\
Code A & \href{https://huggingface.co/MergeBench/Llama-3.1-8B_coding}{\nolinkurl{MergeBench/Llama-3.1-8B_coding}} \\
Code B & \href{https://huggingface.co/pmahdavi/Llama-3.1-8B-coding}{\nolinkurl{pmahdavi/Llama-3.1-8B-coding}} \\
Multilingual & \href{https://huggingface.co/MergeBench/Llama-3.1-8B_multilingual}{\nolinkurl{MergeBench/Llama-3.1-8B_multilingual}} \\
Safety & \href{https://huggingface.co/MergeBench/Llama-3.1-8B_safety}{\nolinkurl{MergeBench/Llama-3.1-8B_safety}} \\
\midrule
\multicolumn{2}{l}{\emph{External evaluator}} \\
Refusal classifier & \href{https://huggingface.co/allenai/wildguard}{\nolinkurl{allenai/wildguard}} \\
\bottomrule
\end{tabular}
\caption{Shared bases and eight experts per architecture, with identifiers linked to their releases. A and B distinguish experts in the same domain. WildGuard is the external evaluator and is not merged.}
\label{tab:models}
\end{table}

\subsection{Benchmarks and Evaluation Protocol}
\label{app:benchmarks}

Table~\ref{tab:benchmark-details} provides benchmark sources and evaluation settings.
ID benchmarks are used for coefficient selection and the reported ID results; OOD benchmarks are held out from selection.

\begin{table}[t]
\centering
\scriptsize
\setlength{\tabcolsep}{4pt}
\begin{tabular}{@{}llrrl@{}}
\toprule
Benchmark & Role & Examples & Shots & Scoring \\
\midrule
IFEval~\citep{zhou2023instruction} & ID & 541 & 0 & Prompt-level strict accuracy \\
AlpacaEval 2.0~\citep{dubois2024length} & OOD & 805 & 0 & Length-controlled win rate \\
\addlinespace
GSM8K-CoT~\citep{cobbe2021training} & ID & 1,319 & 8 & Strict-match exact match \\
MATH-500~\citep{hendrycks2021measuring,lightman2024let} & OOD & 500 & 4 & Verified accuracy (\texttt{math\_verify}) \\
\addlinespace
MBPP~\citep{austin2021program} & ID & 500 & 3 & pass@1 \\
HumanEval~\citep{chen2021evaluating} & OOD & 164 & 0 & pass@1 \\
\addlinespace
FLORES+ dev~\citep{nllb2024scaling,burchell2024flores} & ID & $100\times8$ & 0 & chrF2 \\
FLORES+ devtest & OOD & $1{,}012\times8$ & 0 & chrF2 \\
\addlinespace
HarmBench~\citep{mazeika2024harmbench} & ID & 320 & 0 & Refuse-to-answer rate \\
WildGuardTest~\citep{han2024wildguard} & OOD & 749 & 0 & Refuse-to-answer rate \\
\bottomrule
\end{tabular}
\caption{Sample counts, prompting, and scoring for ID selection and held-out OOD evaluation. Shots denotes in-context examples. FLORES+ counts are sentences per target language times eight languages. Refusal rates use WildGuard.}
\label{tab:benchmark-details}
\end{table}

We use \href{https://huggingface.co/datasets/openlanguagedata/flores_plus}{\nolinkurl{openlanguagedata/flores_plus}} for translation from English into German, French, Spanish, Chinese, Russian, Arabic, Hindi, and Japanese.
ID evaluation reuses a fixed subset of 100 dev sentences per target language (seed 42); OOD evaluation uses all 1,012 devtest sentences per language.
The multilingual score is the mean chrF2 across these eight languages.
WildGuardTest has 749 evaluated examples after five entries with missing values are excluded.
All 749 are prompts labeled harmful (412 vanilla and 337 adversarial), and HarmBench likewise contains only harmful behaviors; both safety scores are therefore refusal rates on harmful requests and do not measure over-refusal of benign requests, which we measure separately in Appendix~\ref{app:reb-safety}.
Accuracy and pass@1 scores are multiplied by 100; chrF2, refusal rates, and the AlpacaEval win rate are expressed on the same 0--100 scale.

Merged candidates inherit the base model's configuration and tokenizer, with benchmark-specific prompt formatting applied consistently.
The lm-eval tasks and FLORES+ use raw prompts; AlpacaEval uses the architecture's Instruct chat template, and safety evaluation uses a shared chat format.
Simplex vertices follow this pipeline: their weights equal those of the corresponding experts, but their scores can differ from evaluations using each expert's native tokenizer, prompts, and generation settings.

\subsection{Coefficient-Selection Implementations}
\label{app:coefficient-implementations}

\paragraph{MetaGPT.}
MetaGPT~\citep{zhou2024metagpt} computes $\lambda_k=\|\tau_k\|_2^2/\sum_j\|\tau_j\|_2^2$ from the full task vectors.
This rule gives nonnegative coefficients summing to one and requires no target-metric evaluations.

\paragraph{MAP.}
For each seed, we construct one sequence of 135 coefficient vectors.
With zero-based indexing, indices 0--7 are $e_1,\ldots,e_8$; indices 8--134 are 127 independent draws from $\mathcal{U}([0,1]^8)$, with all coordinates sampled independently.
We initialize \texttt{np.random.default\_rng(seed)} and generate these coordinates with \texttt{uniform(0, 1)}.
The fit at budget $n\in\{45,90,135\}$ uses the first $n$ points, so all three budgets contain the eight vertices and form nested samples.

For $T=8$, each surrogate's quadratic component has $(T+1)(T+2)/2=45$ coefficients.
The budgets of 45, 90, and 135 evaluations are one, two, and three times this count; 90 is the released implementation's default.
We adapt the released implementation in three ways.
First, we divide scores by 100 to match the sigmoid output range, corresponding to a bounded-response model on the original 0--100 score scale.
Second, we replace all-ones surrogate parameter initialization with zeros to avoid initial sigmoid saturation.
Third, we replace the released code's search bounds $[-3,3]^8$ with $[0,1]^8$ so that optimization remains within the domain from which the surrogate-fitting points are sampled.
All reported MAP results use these adjustments to the released implementation.
NSGA-III in DEAP uses a population of 100 for 150 generations over $x\in[0,1]^8$.
We evaluate the surrogates at $x$, select the vector with the highest predicted AVG from the approximate Pareto front, and use that same vector directly as the merging coefficients, $\lambda=x$.
The sigmoid is applied only to the surrogate outputs, not to the merging coefficients.
This selection requires no additional target-metric evaluations.

\paragraph{Evolutionary merging.}
For $T=8$, CMA-ES uses a population size of $4+\lfloor3\ln T\rfloor=10$, an initial mean of 0.5 in every coordinate, an initial step size of $1/6$, and bounds $[0,1]^8$.
For each seed, we use a single run of 20 generations and report ID scores after every generation.
The OOD evaluation checkpoints of 10, 30, 40, 50, 100, 150, and 200 cumulative evaluations correspond to generations 1, 3, 4, 5, 10, 15, and 20, respectively.
At each checkpoint, we report the ID and OOD scores of the candidate with the highest measured ID AVG among all candidates evaluated so far.

\paragraph{SLIM.}
We interpolate the measured ID AVG directly; by linearity, this is equivalent to averaging the domain-wise interpolants (Appendix~\ref{app:aggregation}).
For both quadratic and cubic variants, SLSQP uses analytic gradients, $\sum_k\lambda_k=1$, bounds $0\leq\lambda_k\leq1$, \texttt{ftol}$=10^{-12}$, and at most 1,000 iterations per initialization.
The 20 initializations comprise the barycenter, eight points near the vertices, and 11 random simplex points generated with seed 42.
We retain the converged solution with the highest surrogate value.
Given the measurements and these fixed initializations, the procedure is deterministic and uses no additional target-metric evaluations.
The cubic extension is defined in Appendix~\ref{app:cubic-extension}.

\paragraph{Matched-budget methods on the simplex.}
All methods use the same eight experts, linear merging, coefficients on the simplex, and 36 ID evaluations; seeded methods use seeds 42, 43, and 44.
\emph{Random best-of-36} draws 36 of the 150 held-out interior mixtures of Section~\ref{sec:fidelity} (\texttt{default\_rng(seed).choice}) and selects the one with the highest measured ID AVG.
\emph{Random design + ridge} fits the quadratic Scheff\'{e} model of Section~\ref{sec:structured} to the same 36 mixtures by ridge regression, selecting $\alpha$ by five-fold cross-validation on them (Appendix~\ref{app:random-design-fits}), and maximizes it over the simplex with SLIM's SLSQP procedure.
\emph{MAP (simplex)} fits MAP's domain-wise sigmoid-transformed quadratics, with 36 parameters each, to the eight vertices and the first 28 mixtures of the same draw, runs NSGA-III over $x\in[0,1]^8$ with the surrogates evaluated at $\lambda=x/\sum_k x_k$, and merges with the normalized coefficients of the Pareto point with the highest predicted AVG; other settings follow the MAP paragraph above.
\emph{Evol.\ Merge (simplex)} runs CMA-ES with the configuration above (initial mean 0.5, step size $1/6$, population 10, bounds $[0,1]^8$) but evaluates $\lambda=x/\sum_k x_k$; the 36 evaluations comprise three generations and six candidates of the fourth, and the candidate with the highest measured ID AVG is selected.
\emph{GP-BO}~\citep{snoek2012practical} starts from the eight vertices and makes 28 sequential queries using a Gaussian process with a constant $\times$ Mat\'{e}rn-5/2 kernel with automatic relevance determination plus white noise, refit on all data at every step, and expected improvement ($\xi=0.01$), maximized over the simplex by refining the best 10 of 10{,}000 $\operatorname{Dirichlet}(1,\ldots,1)$ candidates with SLSQP; it selects the evaluated point with the highest measured ID AVG.
SLIM, the best lattice point, random best-of-36, random design + ridge, and MAP select from the lattice and held-out measurements of Section~\ref{sec:experiments}, whereas evolutionary merging and GP-BO select from measurements taken during their runs; refitting SLIM on lattice values evaluated on the comparison GPU node leaves its OOD AVG nearly unchanged (Appendix~\ref{app:reb-stability}).
All selected models were evaluated on the ID and OOD benchmarks on the same GPU node per architecture (Appendix~\ref{app:reb-protocol}); for evolutionary merging and GP-BO, the reported ID values are the in-run measurements used for selection.
SLIM's surrogate optimization takes 0.25\,s (Gemma) and 1.62\,s (Llama) on a CPU.

\subsection{Random-Design Fits}
\label{app:random-design-fits}

The evaluation-placement ablation uses the same Scheff\'{e} basis for LS and ridge: eight linear terms and 28 pairwise terms, with no explicit intercept.
For ridge, each feature is divided by its root mean square without centering.
We select $\alpha\in\{10^{-4},10^{-3},\ldots,10^2\}$ by five-fold cross-validation on the $n$ fitting points, minimizing mean squared error and recomputing feature scales within each training fold.
The remaining $150-n$ points are used only to assess prediction fidelity.

\subsection{Hardware and Statistical Intervals}
\label{app:reb-protocol}

\paragraph{Hardware.}
Except for the best lattice point in Table~\ref{tab:selection_ablation} and the paired intervals and coefficient-stability analysis of Section~\ref{sec:surrogate_selection}, which use the GPU node below, the results of Sections~\ref{sec:experiments}, \ref{sec:structured}, and~\ref{sec:surrogate_selection}, including the cubic variant, were measured on a cluster of H100 NVL GPUs.
The models compared in pairs in Section~\ref{sec:matched-budget} and Tables~\ref{tab:reb-e21}, \ref{tab:reb-avg-ci}--\ref{tab:reb-domain-ci}, \ref{tab:reb-e11}, and~\ref{tab:reb-overrefusal} were evaluated on the same H200 NVL GPU node per architecture, so the two models in each such comparison share hardware.
On this GPU node, the AVG values of SLIM, MAP ($n=45$), and evolutionary merging (gen$=$20) differ from those in Table~\ref{tab:main} by at most 0.34 points before rounding.
Auxiliary measurements used other GPU nodes, as noted in each table.
After a GPU node failure, several expert-subset groups were split across two GPU nodes of the same type (H200 NVL), on which a canary run produced bit-identical generations; three Llama subset groups and the Llama local validation, scale grid, and re-checks were measured on an H100 PCIe GPU node.

\paragraph{Intervals.}
Intervals are item-level paired bootstrap percentile intervals (95\%, $B=10{,}000$, seed 20261109) without multiplicity adjustment.
Each replicate applies the same item resampling to both models and recomputes all domain scores and AVG; FLORES+ is resampled by source sentence and AlpacaEval by instruction, refitting the length-controlled (LC) model in every replicate.
For seeded methods, the interval is for the difference from the three-seed mean and resamples items with the three models fixed; it excludes seed variability, so we also report per-seed differences.
ID intervals do not correct for the optimism of selecting on the same benchmarks; the held-out OOD comparisons carry the main evidence.
AlpacaEval preferences were annotated once per model by gpt-4o-2024-08-06 and are held fixed in the intervals, which therefore exclude annotator variability.
The LC fit depends on the CPU (differences of up to 1.1 points on other CPUs), so all LC values of the H200 NVL comparisons, including those in the intervals, were computed on one CPU; the LC values of the H100 NVL results above were computed on a different machine.

\paragraph{Deviations from the pre-specified protocol.}
We fixed the protocol of the experiments in Section~\ref{sec:matched-budget} and Appendices~\ref{app:reb-surrogate}--\ref{app:reb-safety} before running them and note the following deviations.
(i) Before any expert-subset result was available, the number of held-out mixtures per expert subset was increased from 30 to 50.
(ii) After seeing the results, we decided not to run MAP on the simplex with 45 evaluations, because the 36-evaluation version already scored higher than MAP ($n=45$) on $[0,1]^8$ in three settings and within 0.40 points in the fourth.
(iii) The over-refusal measurement and the safety-weight sensitivity of Appendix~\ref{app:reb-safety} were added after seeing the analysis with safety removed (AVG$_4$).
(iv) Measurements from one GPU node with an unstable network connection were discarded because of the GPU node's reliability, not their values, and repeated on another GPU node.
(v) The following analyses were not in the protocol and were added after the measurements: the sparse-mixture comparison of Table~\ref{tab:reb-predictor}, the AVG$_4$ ranks of Table~\ref{tab:reb-overrefusal}, the threshold shares of Table~\ref{tab:reb-stability}(b) (excerpted in Table~\ref{tab:stability-main}), and the centered MAE in Appendix~\ref{app:reb-sparse}.

\subsection{Numerical Results for Figure~\ref{fig:cost_perf}}
\label{app:cost-performance-results}

Table~\ref{tab:main} gives the numerical results underlying Figure~\ref{fig:cost_perf}.
All coefficients are selected using ID evaluations only.
Tables~\ref{tab:full_gemma} and~\ref{tab:full_llama} report the domain scores behind Table~\ref{tab:main}, together with those of the base model and the eight experts.

\begin{table}[t]
\centering
\small
\setlength{\tabcolsep}{3pt}
\begin{tabular}{@{}lrrrrrrr@{}}
\toprule
 & & \multicolumn{3}{c}{Gemma-2-9B} & \multicolumn{3}{c}{Llama-3.1-8B} \\
\cmidrule(lr){3-5}\cmidrule(lr){6-8}
Method & \#Evals & GPU-h & ID AVG & OOD AVG & GPU-h & ID AVG & OOD AVG \\
\midrule
Weight Averaging & 0 & 0 & 50.1 & 40.0 & 0 & 43.3 & 33.2 \\
MetaGPT & 0 & 0 & 62.4 & \textbf{50.6} & 0 & 47.1 & 35.5 \\
Task Arithmetic & 21 & 3.0 & 52.0 & 39.5 & 2.7 & 50.2 & 37.6 \\
\midrule
Evol.\ Merge (gen$=$1) & 10 & 1.4 & $51.5_{\pm0.8}$ & $32.7_{\pm1.1}$ & 1.3 & $54.0_{\pm1.0}$ & $38.3_{\pm0.4}$ \\
Evol.\ Merge (gen$=$3) & 30 & 4.1 & $60.1_{\pm0.8}$ & $38.6_{\pm1.3}$ & 3.9 & $58.4_{\pm1.7}$ & $42.1_{\pm2.0}$ \\
Evol.\ Merge (gen$=$4) & 40 & 5.4 & $60.7_{\pm0.5}$ & $40.7_{\pm3.1}$ & 5.0 & $58.9_{\pm1.9}$ & $41.1_{\pm1.1}$ \\
Evol.\ Merge (gen$=$5) & 50 & 6.7 & $63.5_{\pm1.8}$ & $45.5_{\pm4.4}$ & 6.2 & $60.0_{\pm1.9}$ & $43.3_{\pm1.3}$ \\
Evol.\ Merge (gen$=$10) & 100 & 13.8 & $65.3_{\pm2.1}$ & $46.9_{\pm4.4}$ & 11.9 & $63.2_{\pm0.1}$ & $45.8_{\pm2.2}$ \\
Evol.\ Merge (gen$=$15) & 150 & 20.4 & $66.4_{\pm2.4}$ & $48.7_{\pm0.6}$ & 17.6 & $63.9_{\pm0.6}$ & $47.2_{\pm1.5}$ \\
Evol.\ Merge (gen$=$20) & 200 & 26.8 & $\mathbf{67.4}_{\pm1.2}$ & $49.0_{\pm0.9}$ & 23.2 & $\mathbf{64.6}_{\pm0.6}$ & $45.5_{\pm1.9}$ \\
\midrule
MAP ($n=45$) & 45 & 6.3 & $56.7_{\pm4.2}$ & $41.2_{\pm5.6}$ & 5.5 & $47.4_{\pm1.8}$ & $36.7_{\pm2.1}$ \\
MAP ($n=90$) & 90 & 12.4 & $58.3_{\pm2.6}$ & $47.5_{\pm2.7}$ & 11.1 & $57.6_{\pm5.9}$ & $43.1_{\pm7.0}$ \\
MAP ($n=135$) & 135 & 18.5 & $61.2_{\pm1.5}$ & $50.6_{\pm0.4}$ & 16.7 & $60.4_{\pm1.2}$ & $45.3_{\pm2.6}$ \\
\midrule
SLIM (quadratic) & 36 & 5.2 & 66.2 & 50.3 & 3.9 & 56.6 & \textbf{47.3} \\
SLIM (cubic) & 120 & 16.4 & 66.4 & 48.4 & 12.9 & 57.0 & 46.3 \\
\bottomrule
\end{tabular}
\caption{ID and held-out OOD performance across coefficient-selection budgets ($T=8$; Figure~\ref{fig:cost_perf}).
AVG is the unweighted mean of five domain scores on a 0--100 scale.
MAP and evolutionary merging report mean $\pm$ standard deviation over three seeds; their GPU-hours are means.
Costs cover coefficient selection per run, excluding final reporting and fidelity measurements.
Zero cost denotes no target-metric evaluations.
Bold marks the highest AVG in each column before rounding.
On Gemma OOD, MetaGPT (50.62) and MAP with $n=135$ (50.60) both round to 50.6.
The cubic variant is a degree ablation.}
\label{tab:main}
\end{table}

\subsection{Domain-wise Surrogate Fidelity}
\label{app:domain-fidelity}

Table~\ref{tab:fidelity_domain} reports fidelity on the same 150 held-out mixtures as Figure~\ref{fig:fidelity}.
The interquartile range (IQR) of measured scores provides context for the MAE.
The AVG row evaluates predictions of the aggregate objective; it is not an average of the domain-wise fidelity statistics.

\begin{table}[t]
\centering
\small
\setlength{\tabcolsep}{5pt}
\begin{tabular}{@{}lrrrrrr@{}}
\toprule
 & \multicolumn{3}{c}{Gemma-2-9B} & \multicolumn{3}{c}{Llama-3.1-8B} \\
\cmidrule(lr){2-4}\cmidrule(lr){5-7}
Domain & $\rho$ & MAE & IQR & $\rho$ & MAE & IQR \\
\midrule
Instruction & 0.97 & 2.60 & 15.9 & 0.96 & 2.46 & 10.9 \\
Math & $-0.24$ & 4.88 & 5.0 & 0.85 & 2.33 & 6.5 \\
Coding & 0.86 & 1.89 & 4.2 & 0.70 & 1.93 & 3.9 \\
Multilingual & 0.63 & 1.40 & 1.1 & 0.56 & 1.22 & 0.6 \\
Safety & 0.94 & 9.20 & 47.2 & 0.95 & 4.39 & 19.9 \\
\midrule
AVG & 0.936 & 1.93 & 10.1 & 0.959 & 1.07 & 4.6 \\
\bottomrule
\end{tabular}
\caption{AVG is accurately predicted despite variation in domain-wise fidelity. On 150 held-out mixtures, Spearman's $\rho$ compares predicted and measured rankings; MAE and the IQR of measured scores are in score points. All interpolants use the same 36 construction evaluations.}
\label{tab:fidelity_domain}
\end{table}

\begin{sidewaystable}
\centering
\small
\colorlet{avgshade}{black!7}
\setlength{\tabcolsep}{3.5pt}
\renewcommand{\arraystretch}{1.12}
\begin{tabular}{@{}lrrrrr>{\columncolor{avgshade}}rrr@{\hspace{12pt}}rrrrr>{\columncolor{avgshade}}r@{}}
\toprule
 & \multicolumn{8}{c}{ID (used for coefficient selection, with its cost)} & \multicolumn{6}{c}{OOD (held out from selection)} \\
\cmidrule(lr){2-9}\cmidrule(l){10-15}
Method & Instr. & Math & Coding & Multi. & Safety & AVG & \#Evals & GPU-h & Instr. & Math & Coding & Multi. & Safety & AVG \\
\midrule
Base & 15.0 & 66.2 & 53.4 & 38.7 & 9.1 & 36.5 & -- & -- & 0.0 & 33.2 & 37.2 & 38.4 & 18.6 & 25.5 \\
\quad Instruction A & 65.4 & 66.2 & 54.2 & 52.3 & 12.2 & 50.1 & -- & -- & 9.6 & 39.0 & 53.7 & 50.7 & 20.2 & 34.6 \\
\quad Instruction B & 51.2 & 66.7 & 57.6 & 44.5 & 72.5 & 58.5 & -- & -- & 45.9 & 40.6 & 54.9 & 41.7 & 65.7 & 49.7 \\
\quad Math A & 8.9 & 5.8$^{\dagger}$ & 34.0 & 12.7 & 3.4 & 13.0 & -- & -- & 0.7 & 40.2 & 3.0 & 12.2 & 10.0 & 13.2 \\
\quad Math B & 16.6 & 58.0 & 27.6 & 14.7 & 7.2 & 24.8 & -- & -- & 0.3 & 36.6 & 8.5 & 13.9 & 11.1 & 14.1 \\
\quad Code A & 24.2 & 45.6 & 48.8 & 51.8 & 5.3 & 35.2 & -- & -- & 3.1 & 23.4 & 67.1 & 50.4 & 22.4 & 33.3 \\
\quad Code B & 17.0 & 61.6 & 52.4 & 43.7 & 18.4 & 38.6 & -- & -- & 0.1 & 35.0 & 36.6 & 42.2 & 20.0 & 26.8 \\
\quad Multilingual & 14.8 & 65.8 & 55.4 & 47.2 & 35.0 & 43.6 & -- & -- & 3.1 & 34.0 & 43.3 & 46.3 & 18.3 & 29.0 \\
\quad Safety & 15.2 & 57.5 & 44.0 & 49.9 & 75.6 & 48.4 & -- & -- & 4.7 & 28.4 & 40.2 & 48.9 & 62.5 & 37.0 \\
\midrule
Weight Averaging & 29.0 & 68.4 & 54.8 & 54.5 & 43.8 & 50.1 & 0 & 0 & 8.1 & 44.8 & 55.5 & 53.4 & 38.3 & 40.0 \\
MetaGPT & 51.2 & \textbf{76.6} & 58.0 & 56.0 & 70.0 & 62.4 & 0 & 0 & 39.2 & \textbf{47.4} & 57.3 & 54.3 & 54.9 & \textbf{50.6} \\
Task Arithmetic & 36.4 & 72.3 & 50.4 & 51.3 & 49.7 & 52.0 & 21 & 3.0 & 12.3 & 46.0 & 48.2 & 49.8 & 41.3 & 39.5 \\
\midrule
Evol.\ Merge (gen$=$1) & $36.7_{\pm3.4}$ & $64.8_{\pm4.3}$ & $43.8_{\pm0.4}$ & $41.0_{\pm0.8}$ & $71.2_{\pm6.6}$ & $51.5_{\pm0.8}$ & 10 & 1.4 & $13.6_{\pm2.3}$ & $42.3_{\pm0.3}$ & $18.5_{\pm7.0}$ & $39.9_{\pm0.7}$ & $49.4_{\pm2.2}$ & $32.7_{\pm1.1}$ \\
Evol.\ Merge (gen$=$3) & $46.2_{\pm2.4}$ & $67.0_{\pm3.9}$ & $48.6_{\pm1.1}$ & $48.5_{\pm1.4}$ & $90.0_{\pm2.3}$ & $60.1_{\pm0.8}$ & 30 & 4.1 & $23.2_{\pm3.5}$ & $40.9_{\pm0.5}$ & $27.8_{\pm10.7}$ & $47.0_{\pm1.3}$ & $54.1_{\pm0.5}$ & $38.6_{\pm1.3}$ \\
Evol.\ Merge (gen$=$4) & $47.1_{\pm3.0}$ & $70.6_{\pm1.5}$ & $48.8_{\pm1.6}$ & $49.8_{\pm1.5}$ & $87.2_{\pm7.7}$ & $60.7_{\pm0.5}$ & 40 & 5.4 & $23.4_{\pm2.8}$ & $41.5_{\pm1.4}$ & $35.4_{\pm15.3}$ & $48.3_{\pm1.5}$ & $54.8_{\pm2.4}$ & $40.7_{\pm3.1}$ \\
Evol.\ Merge (gen$=$5) & $49.8_{\pm8.6}$ & $68.7_{\pm4.7}$ & $51.1_{\pm3.6}$ & $52.0_{\pm2.1}$ & $\mathbf{95.9}_{\pm1.4}$ & $63.5_{\pm1.8}$ & 50 & 6.7 & $30.5_{\pm9.3}$ & $40.1_{\pm2.3}$ & $46.7_{\pm10.3}$ & $50.5_{\pm2.1}$ & $\mathbf{59.4}_{\pm1.4}$ & $45.5_{\pm4.4}$ \\
Evol.\ Merge (gen$=$10) & $58.0_{\pm7.0}$ & $68.7_{\pm4.9}$ & $55.4_{\pm4.1}$ & $53.0_{\pm2.7}$ & $91.5_{\pm0.5}$ & $65.3_{\pm2.1}$ & 100 & 13.8 & $34.2_{\pm7.7}$ & $42.0_{\pm0.4}$ & $52.2_{\pm13.6}$ & $51.5_{\pm2.8}$ & $54.6_{\pm2.2}$ & $46.9_{\pm4.4}$ \\
Evol.\ Merge (gen$=$15) & $61.6_{\pm0.7}$ & $71.9_{\pm2.2}$ & $54.9_{\pm1.7}$ & $53.9_{\pm1.4}$ & $89.7_{\pm7.3}$ & $66.4_{\pm2.4}$ & 150 & 20.4 & $36.0_{\pm1.3}$ & $44.4_{\pm2.1}$ & $55.1_{\pm0.4}$ & $52.1_{\pm1.6}$ & $56.0_{\pm1.8}$ & $48.7_{\pm0.6}$ \\
Evol.\ Merge (gen$=$20) & $\mathbf{63.6}_{\pm2.3}$ & $70.4_{\pm5.3}$ & $54.6_{\pm0.7}$ & $54.2_{\pm0.1}$ & $94.3_{\pm2.0}$ & $\mathbf{67.4}_{\pm1.2}$ & 200 & 26.8 & $38.1_{\pm3.5}$ & $42.9_{\pm1.2}$ & $54.5_{\pm2.9}$ & $52.6_{\pm0.2}$ & $57.0_{\pm0.3}$ & $49.0_{\pm0.9}$ \\
\midrule
MAP ($n=45$) & $43.8_{\pm6.4}$ & $63.0_{\pm0.5}$ & $48.5_{\pm4.9}$ & $50.4_{\pm1.7}$ & $78.0_{\pm16.3}$ & $56.7_{\pm4.2}$ & 45 & 6.3 & $24.7_{\pm13.2}$ & $39.0_{\pm6.5}$ & $38.4_{\pm5.0}$ & $48.8_{\pm1.5}$ & $55.1_{\pm10.4}$ & $41.2_{\pm5.6}$ \\
MAP ($n=90$) & $58.5_{\pm7.2}$ & $43.8_{\pm9.1}$ & $54.2_{\pm0.5}$ & $52.9_{\pm0.8}$ & $82.3_{\pm7.1}$ & $58.3_{\pm2.6}$ & 90 & 12.4 & $44.7_{\pm3.5}$ & $38.6_{\pm3.3}$ & $45.5_{\pm11.1}$ & $50.8_{\pm0.3}$ & $57.8_{\pm1.7}$ & $47.5_{\pm2.7}$ \\
MAP ($n=135$) & $63.2_{\pm4.4}$ & $46.5_{\pm4.5}$ & $54.9_{\pm1.1}$ & $54.0_{\pm0.5}$ & $87.1_{\pm3.1}$ & $61.2_{\pm1.5}$ & 135 & 18.5 & $\mathbf{46.7}_{\pm1.7}$ & $37.7_{\pm0.8}$ & $\mathbf{59.8}_{\pm3.4}$ & $51.9_{\pm0.5}$ & $57.0_{\pm2.0}$ & $50.6_{\pm0.4}$ \\
\midrule
SLIM (quadratic) & 60.3 & 71.3 & \textbf{58.4} & \textbf{56.5} & 84.4 & 66.2 & 36 & 5.2 & 38.7 & 40.6 & 59.1 & \textbf{54.6} & 58.5 & 50.3 \\
SLIM (cubic) & 58.4 & 70.7 & 56.4 & 55.6 & 90.6 & 66.4 & 120 & 16.4 & 29.6 & 41.0 & 57.9 & 54.2 & 59.3 & 48.4 \\
\bottomrule
\end{tabular}
\caption{Per-domain ID and held-out OOD scores on Gemma-2-9B ($T=8$), extending Table~\ref{tab:main} with the base model and the eight experts. Experts are evaluated through the same merging pipeline as the merged models, with one-hot coefficients; their ID scores are the vertices of the lattice used to construct quadratic SLIM. Domain metrics follow Table~\ref{tab:benchmark-details}; AVG, the mean $\pm$ standard deviation over three seeds, \#Evals, and GPU-hours (means) follow Table~\ref{tab:main}. Bold marks the highest value among merging methods in each score column before rounding. In the OOD AVG column, MetaGPT (50.62) and MAP with $n=135$ (50.60) both round to 50.6. $^{\dagger}$Math~A answers in a format that GSM8K strict matching does not extract.}
\label{tab:full_gemma}
\end{sidewaystable}

\begin{sidewaystable}
\centering
\small
\colorlet{avgshade}{black!7}
\setlength{\tabcolsep}{3.5pt}
\renewcommand{\arraystretch}{1.12}
\begin{tabular}{@{}lrrrrr>{\columncolor{avgshade}}rrr@{\hspace{12pt}}rrrrr>{\columncolor{avgshade}}r@{}}
\toprule
 & \multicolumn{8}{c}{ID (used for coefficient selection, with its cost)} & \multicolumn{6}{c}{OOD (held out from selection)} \\
\cmidrule(lr){2-9}\cmidrule(l){10-15}
Method & Instr. & Math & Coding & Multi. & Safety & AVG & \#Evals & GPU-h & Instr. & Math & Coding & Multi. & Safety & AVG \\
\midrule
Base & 8.3 & 54.1 & 48.4 & 48.9 & 2.8 & 32.5 & -- & -- & 0.3 & 21.4 & 37.8 & 47.5 & 5.6 & 22.5 \\
\quad Instruction A & 47.0 & 57.9 & 52.4 & 50.7 & 8.4 & 43.3 & -- & -- & 8.7 & 23.4 & 45.7 & 49.5 & 13.4 & 28.1 \\
\quad Instruction B & 44.4 & 76.9 & 58.6 & 22.6 & 71.6 & 54.8 & -- & -- & 31.3 & 36.2 & 64.0 & 22.1 & 78.9 & 46.5 \\
\quad Math A & 13.1 & 7.1$^{\dagger}$ & 38.0 & 17.5 & 9.7 & 17.1 & -- & -- & 1.0 & 38.4 & 28.7 & 17.4 & 10.9 & 19.3 \\
\quad Math B & 19.4 & 75.4 & 48.4 & 48.9 & 7.8 & 40.0 & -- & -- & 6.9 & 33.0 & 44.5 & 47.8 & 1.9 & 26.8 \\
\quad Code A & 22.4 & 45.0 & 46.2 & 49.9 & 9.1 & 34.5 & -- & -- & 3.8 & 16.4 & 63.4 & 48.3 & 18.8 & 30.1 \\
\quad Code B & 14.0 & 55.3 & 46.2 & 50.8 & 5.3 & 34.3 & -- & -- & 4.5 & 21.2 & 53.0 & 49.3 & 12.0 & 28.0 \\
\quad Multilingual & 4.1 & 51.5 & 39.6 & 49.2 & 20.9 & 33.0 & -- & -- & 3.2 & 22.0 & 41.5 & 48.4 & 10.3 & 25.1 \\
\quad Safety & 7.4 & 46.9 & 44.8 & 47.8 & 54.7 & 40.3 & -- & -- & 3.9 & 17.6 & 41.5 & 46.1 & 46.1 & 31.0 \\
\midrule
Weight Averaging & 20.3 & 69.4 & 53.4 & \textbf{51.6} & 21.9 & 43.3 & 0 & 0 & 6.9 & 30.8 & 52.4 & \textbf{50.5} & 25.5 & 33.2 \\
MetaGPT & 25.5 & 73.7 & 53.8 & 49.9 & 32.8 & 47.1 & 0 & 0 & 1.1 & 38.0 & 54.9 & 49.2 & 34.3 & 35.5 \\
Task Arithmetic & 37.3 & 72.6 & 53.8 & 50.3 & 37.2 & 50.2 & 21 & 2.7 & 8.7 & \textbf{39.0} & 57.3 & 48.8 & 34.0 & 37.6 \\
\midrule
Evol.\ Merge (gen$=$1) & $38.4_{\pm4.6}$ & $73.1_{\pm1.1}$ & $51.2_{\pm2.0}$ & $46.3_{\pm1.0}$ & $60.9_{\pm11.7}$ & $54.0_{\pm1.0}$ & 10 & 1.3 & $5.8_{\pm5.3}$ & $37.1_{\pm2.1}$ & $62.4_{\pm2.5}$ & $44.2_{\pm1.0}$ & $41.8_{\pm5.9}$ & $38.3_{\pm0.4}$ \\
Evol.\ Merge (gen$=$3) & $41.1_{\pm4.3}$ & $70.0_{\pm3.5}$ & $53.4_{\pm1.6}$ & $48.8_{\pm1.4}$ & $79.0_{\pm10.2}$ & $58.4_{\pm1.7}$ & 30 & 3.9 & $19.2_{\pm8.0}$ & $30.5_{\pm4.4}$ & $\mathbf{66.9}_{\pm2.7}$ & $47.4_{\pm1.3}$ & $46.6_{\pm1.8}$ & $42.1_{\pm2.0}$ \\
Evol.\ Merge (gen$=$4) & $42.0_{\pm3.4}$ & $72.7_{\pm4.5}$ & $52.9_{\pm0.1}$ & $48.4_{\pm1.8}$ & $78.8_{\pm10.8}$ & $58.9_{\pm1.9}$ & 40 & 5.0 & $13.8_{\pm8.6}$ & $32.9_{\pm3.0}$ & $64.0_{\pm1.1}$ & $46.8_{\pm1.4}$ & $47.9_{\pm4.2}$ & $41.1_{\pm1.1}$ \\
Evol.\ Merge (gen$=$5) & $44.0_{\pm1.2}$ & $71.7_{\pm2.4}$ & $54.4_{\pm1.8}$ & $48.8_{\pm0.3}$ & $80.8_{\pm7.7}$ & $60.0_{\pm1.9}$ & 50 & 6.2 & $22.6_{\pm7.2}$ & $33.3_{\pm1.9}$ & $65.9_{\pm4.4}$ & $47.4_{\pm0.2}$ & $47.4_{\pm3.8}$ & $43.3_{\pm1.3}$ \\
Evol.\ Merge (gen$=$10) & $46.9_{\pm4.8}$ & $77.3_{\pm0.9}$ & $55.1_{\pm1.1}$ & $49.5_{\pm0.9}$ & $87.2_{\pm3.6}$ & $63.2_{\pm0.1}$ & 100 & 11.9 & $29.4_{\pm5.2}$ & $36.3_{\pm1.5}$ & $61.2_{\pm2.5}$ & $48.2_{\pm0.7}$ & $53.8_{\pm8.0}$ & $45.8_{\pm2.2}$ \\
Evol.\ Merge (gen$=$15) & $46.8_{\pm1.7}$ & $78.0_{\pm2.1}$ & $54.3_{\pm0.2}$ & $50.2_{\pm0.5}$ & $90.4_{\pm2.2}$ & $63.9_{\pm0.6}$ & 150 & 17.6 & $31.5_{\pm5.1}$ & $37.5_{\pm1.7}$ & $64.2_{\pm1.3}$ & $48.7_{\pm0.4}$ & $54.0_{\pm0.6}$ & $47.2_{\pm1.5}$ \\
Evol.\ Merge (gen$=$20) & $\mathbf{49.2}_{\pm2.2}$ & $\mathbf{79.0}_{\pm1.1}$ & $53.8_{\pm0.9}$ & $50.1_{\pm0.8}$ & $\mathbf{90.8}_{\pm2.0}$ & $\mathbf{64.6}_{\pm0.6}$ & 200 & 23.2 & $27.4_{\pm6.7}$ & $37.1_{\pm1.5}$ & $61.8_{\pm2.0}$ & $48.6_{\pm0.7}$ & $52.7_{\pm0.5}$ & $45.5_{\pm1.9}$ \\
\midrule
MAP ($n=45$) & $32.9_{\pm8.4}$ & $57.3_{\pm12.5}$ & $48.6_{\pm0.9}$ & $43.1_{\pm8.5}$ & $55.2_{\pm7.6}$ & $47.4_{\pm1.8}$ & 45 & 5.5 & $19.2_{\pm12.3}$ & $25.1_{\pm5.4}$ & $58.7_{\pm8.2}$ & $41.5_{\pm7.9}$ & $38.8_{\pm1.5}$ & $36.7_{\pm2.1}$ \\
MAP ($n=90$) & $42.0_{\pm4.3}$ & $70.3_{\pm10.9}$ & $52.0_{\pm4.0}$ & $47.9_{\pm2.5}$ & $75.9_{\pm10.3}$ & $57.6_{\pm5.9}$ & 90 & 11.1 & $29.9_{\pm11.4}$ & $31.0_{\pm7.3}$ & $63.0_{\pm8.1}$ & $46.7_{\pm2.2}$ & $44.7_{\pm6.9}$ & $43.1_{\pm7.0}$ \\
MAP ($n=135$) & $43.3_{\pm4.4}$ & $73.4_{\pm2.2}$ & $54.5_{\pm0.6}$ & $49.1_{\pm0.4}$ & $81.7_{\pm4.0}$ & $60.4_{\pm1.2}$ & 135 & 16.7 & $29.0_{\pm8.5}$ & $35.0_{\pm0.5}$ & $65.2_{\pm1.6}$ & $47.8_{\pm0.4}$ & $49.3_{\pm6.7}$ & $45.3_{\pm2.6}$ \\
\midrule
SLIM (quadratic) & 38.1 & 74.2 & 57.0 & 32.5 & 81.2 & 56.6 & 36 & 3.9 & 33.8 & 36.8 & 58.5 & 32.1 & \textbf{75.2} & \textbf{47.3} \\
SLIM (cubic) & 40.1 & 75.1 & \textbf{58.4} & 32.8 & 78.8 & 57.0 & 120 & 12.9 & \textbf{34.0} & 35.4 & 58.5 & 32.0 & 71.4 & 46.3 \\
\bottomrule
\end{tabular}
\caption{Per-domain ID and held-out OOD scores on Llama-3.1-8B ($T=8$), extending Table~\ref{tab:main} with the base model and the eight experts. Experts are evaluated through the same merging pipeline as the merged models, with one-hot coefficients; their ID scores are the vertices of the lattice used to construct quadratic SLIM. Domain metrics follow Table~\ref{tab:benchmark-details}; AVG, the mean $\pm$ standard deviation over three seeds, \#Evals, and GPU-hours (means) follow Table~\ref{tab:main}. Bold marks the highest value among merging methods in each score column before rounding. $^{\dagger}$Math~A answers in a format that GSM8K strict matching does not extract.}
\label{tab:full_llama}
\end{sidewaystable}

\subsection{Quadratic Interpolation along Simplex Edges}
\label{app:edge-fidelity}

For each expert pair $(i,j)$, we consider
$\lambda(t)=(1-t)e_i+t e_j$ for $t\in[0,1]$.
The quadratic interpolant uses measurements at $t=0,1/2,1$.
We assess its predictions at $t=1/3,2/3$, giving 56 held-out points across 28 edges per architecture.
These points are excluded from quadratic surrogate construction and coefficient selection, although they belong to the cubic design.
All scores are measured on the ID benchmarks.

\begin{figure}[p]
\centering
\includegraphics[
  width=\textwidth,
  height=0.78\textheight,
  keepaspectratio
]{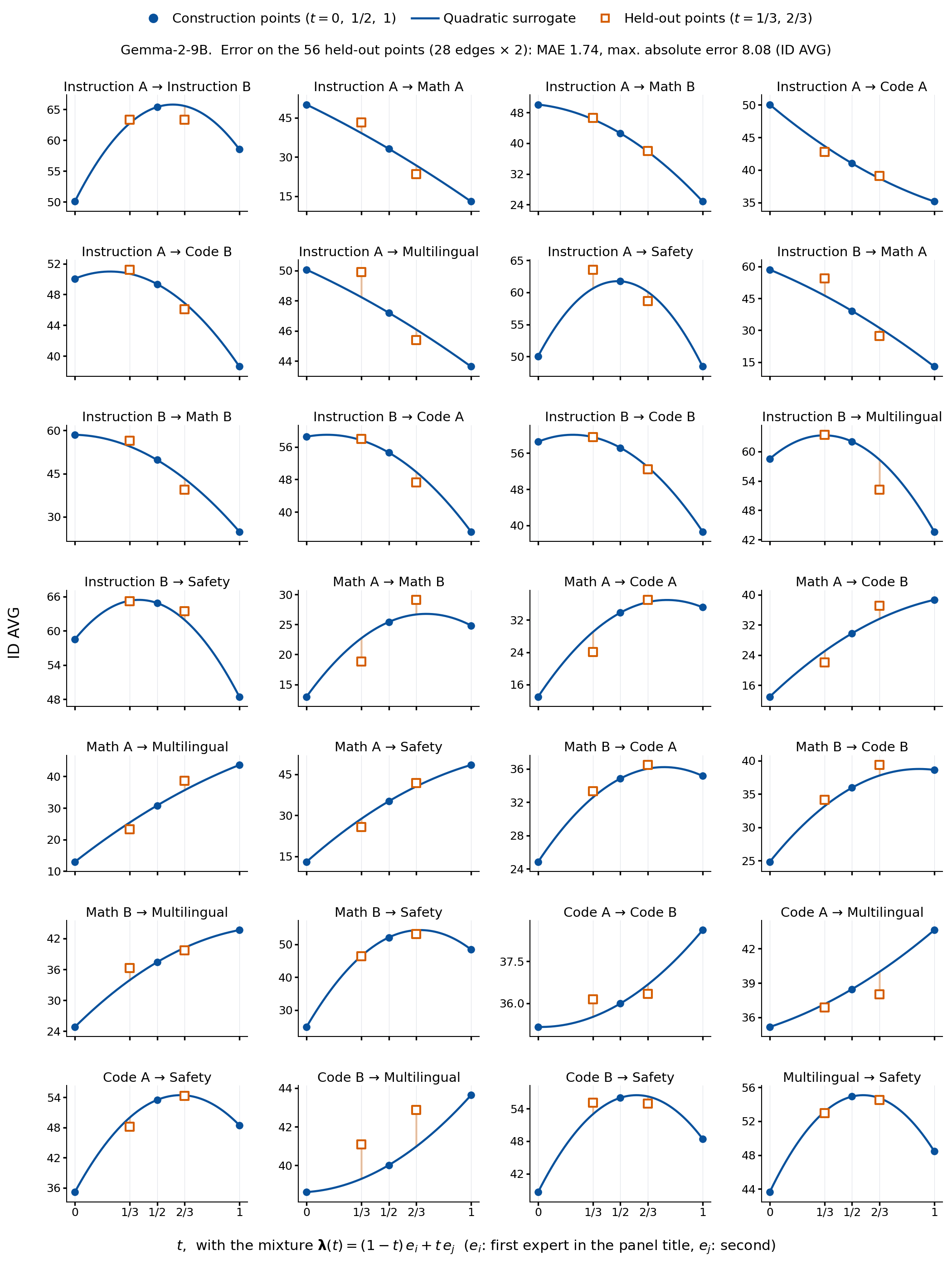}
\caption{Quadratic predictions and measured ID AVG along all 28 simplex edges for Gemma-2-9B.
Blue circles mark construction points and blue curves show the quadratic surrogate.
Orange squares mark points held out from quadratic construction; orange segments show prediction errors.
Panel titles list the experts at $t=0$ and $t=1$.
Y-axis ranges vary across panels.}
\label{fig:edge-curves-gemma}
\end{figure}

\begin{figure}[p]
\centering
\includegraphics[
  width=\textwidth,
  height=0.78\textheight,
  keepaspectratio
]{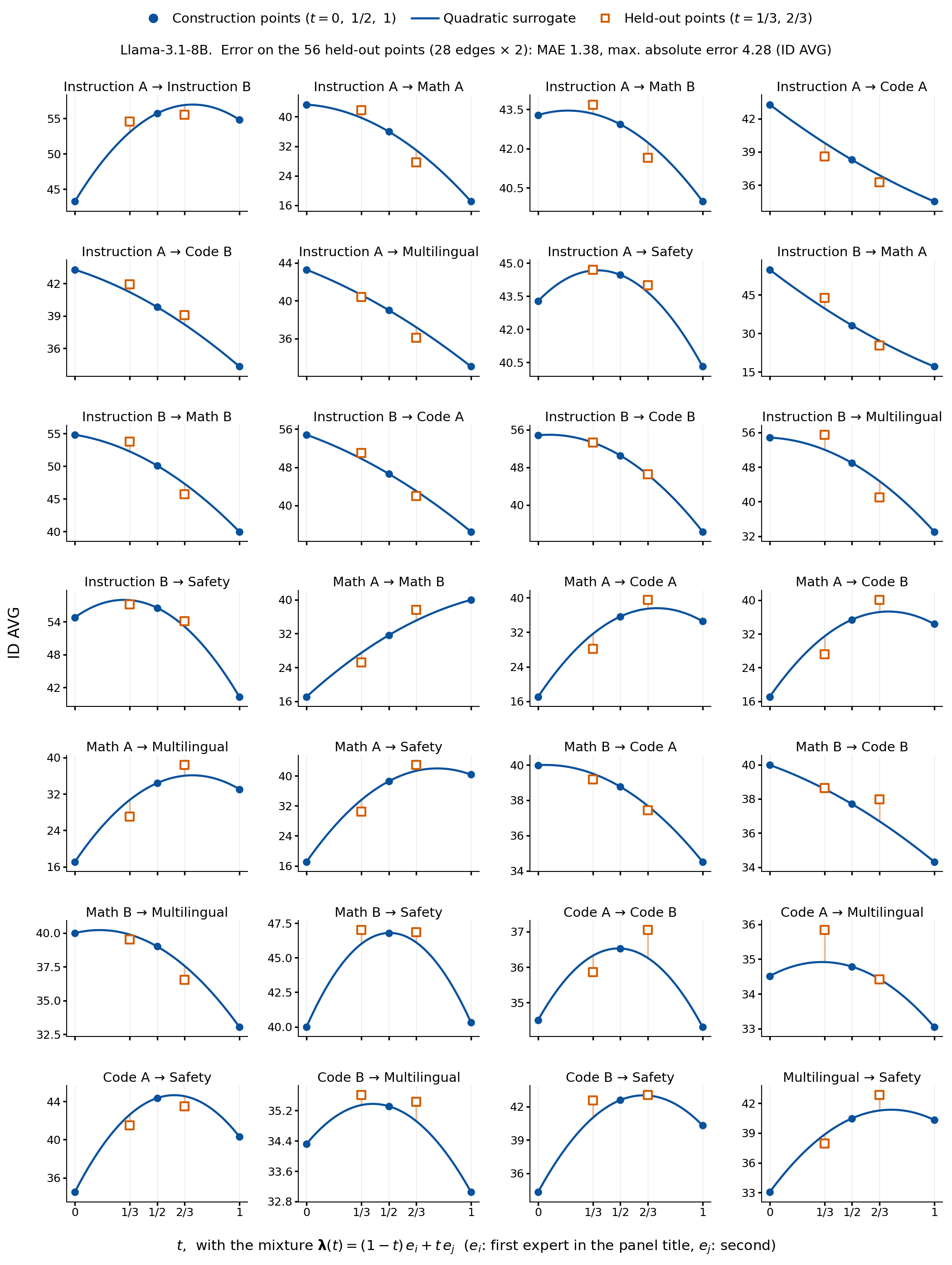}
\caption{Quadratic predictions and measured ID AVG along all 28 simplex edges for Llama-3.1-8B.
Blue circles mark construction points and blue curves show the quadratic surrogate.
Orange squares mark points held out from quadratic construction; orange segments show prediction errors.
Panel titles list the experts at $t=0$ and $t=1$.
Y-axis ranges vary across panels.}
\label{fig:edge-curves-llama}
\end{figure}

Figures~\ref{fig:edge-curves-gemma} and~\ref{fig:edge-curves-llama} show the quadratic predictions and held-out measurements along all simplex edges for Gemma-2-9B and Llama-3.1-8B, respectively.
MAE is 1.74 on Gemma and 1.38 on Llama, with maximum absolute errors of 8.08 and 4.28, respectively.
The curves show useful interpolation overall, with localized deviations.
This analysis complements the interior-mixture evaluation in Section~\ref{sec:fidelity}.

\section{Additional Analyses of the Surrogate}
\label{app:reb-surrogate}

This appendix extends the fidelity analysis of Section~\ref{sec:fidelity} to fewer experts (including selection within subsets), to the selected solutions and their neighborhoods, and to sparse mixtures.

\subsection[\texorpdfstring{Fidelity and Selection with Fewer Experts ($T=5$--$7$)}{Fidelity and Selection with Fewer Experts (T=5-7)}]{Fidelity and Selection with Fewer Experts ($T=5$--$7$)}
\label{app:reb-subsets}

We repeat the fidelity test of Section~\ref{sec:fidelity} on four expert subsets: $\mathcal{S}_5^A=\{$Instruction A, Math A, Code A, Multilingual, Safety$\}$, $\mathcal{S}_5^B=\{$Instruction B, Math B, Code B, Multilingual, Safety$\}$, $\mathcal{S}_6=\mathcal{S}_5^A\cup\{$Instruction B$\}$, and $\mathcal{S}_7=\mathcal{S}_6\cup\{$Math B$\}$.
Each subset surrogate uses only the $T(T+1)/2$ lattice measurements of Section~\ref{sec:experiments} on the subset's face and is tested on 50 new mixtures sampled from $\operatorname{Dirichlet}(1,\ldots,1)$ over the subset (seed 42); for all eight experts, we re-measure the first 50 of the 150 mixtures of Section~\ref{sec:fidelity}.
Table~\ref{tab:reb-e20} reports Spearman's $\rho$ and MAE with intervals from resampling the held-out mixtures ($B=10{,}000$).
For $T=5$--7, $\rho$ ranges from 0.88 to 0.98, at a level comparable to all eight experts: the intervals of the differences from $T=8$ include zero, except for a larger MAE on Llama $\mathcal{S}_5^B$ ($+0.38$ $[0.06, 0.70]$) and a higher $\rho$ on Gemma $\mathcal{S}_5^A$ ($+0.048$ $[0.003, 0.113]$).
Because the number and the identity of the experts change together, these comparisons do not isolate the effect of $T$.
Predictions use the lattice measurements of Section~\ref{sec:experiments}, whereas the held-out mixtures were measured on other GPU nodes; on the 50 mixtures of all eight experts, the new measurements change $\rho$ by $+0.003$ (Gemma) and $+0.011$ (Llama) and MAE by $+0.01$ and $-0.15$ relative to those of Section~\ref{sec:fidelity}.
Each Gemma group and the Llama $\mathcal{S}_7$ group were split across two GPU nodes; restricted to the first 30 mixtures of each Gemma group, all measured on the same GPU node, $\rho$ is 0.92--0.97 for $T=5$--7 and 0.90 for all eight experts (MAE 1.41--2.09 and 2.06).
The Llama comparisons of $\mathcal{S}_5^A$, $\mathcal{S}_5^B$, and $\mathcal{S}_6$ with all eight experts also cross GPU types (H100 PCIe versus H200 NVL).

Table~\ref{tab:reb-e21} repeats the comparison of SLIM and the best lattice point within each subset.
In five of the eight cases, the subset contains the experts used by both solutions for all eight experts, so the same models are selected; in the three independent cases, SLIM is ahead on OOD in two with intervals excluding zero ($+1.70$ and $+2.12$) and within uncertainty in the third ($+0.44$), and its ID differences range from $-0.91$ to $+1.03$.
We did not test more than eight experts. Merging scaling-law results~\citep{wang-2026-merging-scaling-law}, measured with cross-entropy loss for experts merged with equal weights, show that about 90\% of the improvement from merging up to nine experts is reached with six, and their fitted law implies that the gain from each additional expert decays roughly as $k^{-2}$; in a 16-expert extension, gains flatten from about ten experts, and benchmark scores, evaluated for up to five experts, plateau earlier.

\begin{table}[htbp]
\centering
\footnotesize
\setlength{\tabcolsep}{3.5pt}
\begin{tabular}{@{}lccrrrr@{}}
\toprule
 & & & \multicolumn{2}{c}{Gemma-2-9B} & \multicolumn{2}{c}{Llama-3.1-8B} \\
\cmidrule(lr){4-5}\cmidrule(l){6-7}
Expert set & $T$ & \#Evals & Spearman's $\rho$ & MAE & Spearman's $\rho$ & MAE \\
\midrule
$\mathcal{S}_{5}^{A}$ & 5 & 15 & 0.974\,{\scriptsize $[0.939, 0.984]$} & 2.20\,{\scriptsize $[1.75, 2.67]$} & 0.881\,{\scriptsize $[0.770, 0.941]$} & 0.79\,{\scriptsize $[0.60, 1.03]$} \\
$\mathcal{S}_{5}^{B}$ & 5 & 15 & 0.964\,{\scriptsize $[0.918, 0.979]$} & 1.61\,{\scriptsize $[1.27, 1.98]$} & 0.969\,{\scriptsize $[0.930, 0.983]$} & 1.39\,{\scriptsize $[1.13, 1.66]$} \\
$\mathcal{S}_{6}$ & 6 & 21 & 0.945\,{\scriptsize $[0.875, 0.977]$} & 2.07\,{\scriptsize $[1.60, 2.58]$} & 0.978\,{\scriptsize $[0.945, 0.989]$} & 0.90\,{\scriptsize $[0.66, 1.18]$} \\
$\mathcal{S}_{7}$ & 7 & 28 & 0.954\,{\scriptsize $[0.895, 0.978]$} & 1.82\,{\scriptsize $[1.45, 2.21]$} & 0.956\,{\scriptsize $[0.903, 0.978]$} & 1.02\,{\scriptsize $[0.78, 1.29]$} \\
\midrule
All eight & 8 & 36 & 0.926\,{\scriptsize $[0.858, 0.956]$} & 1.79\,{\scriptsize $[1.43, 2.16]$} & 0.951\,{\scriptsize $[0.893, 0.974]$} & 1.01\,{\scriptsize $[0.83, 1.19]$} \\
Sec.~\ref{sec:fidelity}, 150 mixtures & 8 & 36 & 0.936 & 1.93 & 0.959 & 1.07 \\
\bottomrule
\end{tabular}

\caption{AVG prediction fidelity on expert subsets with 50 held-out mixtures each (95\% intervals from resampling mixtures); \#Evals is the number of lattice points. The last row is the result on 150 mixtures in Section~\ref{sec:fidelity}. Each Gemma group was split across two H200 NVL GPU nodes; Llama $\mathcal{S}_5^A$, $\mathcal{S}_5^B$, and $\mathcal{S}_6$ were measured on an H100 PCIe GPU node and the other Llama groups on H200 NVL GPU nodes.}
\label{tab:reb-e20}
\end{table}

\begin{table}[htbp]
\centering
\scriptsize
\setlength{\tabcolsep}{2.2pt}
\begin{tabular}{@{}llclcrr@{}}
\toprule
 & & & \multicolumn{2}{c}{Selected coefficients} & \multicolumn{2}{c}{SLIM $-$ best lattice point (AVG)} \\
\cmidrule(lr){4-5}\cmidrule(l){6-7}
Arch. & Set & \#Evals & SLIM & Best lattice point & ID & OOD \\
\midrule
Gemma & $\mathcal{S}_{5}^{A}$ & 15 & IA\,0.52, S\,0.48 & IA\,0.50, S\,0.50 & $+1.03$$^{*}$\,{\scriptsize $[0.41, 1.65]$} & $+0.44$\phantom{$^{*}$}\,{\scriptsize $[-0.35, 1.22]$} \\
Gemma & $\mathcal{S}_{5}^{B}$ & 15 & IB\,0.55, ML\,0.14, S\,0.30 & IB\,0.50, S\,0.50 & $-0.91$\phantom{$^{*}$}\,{\scriptsize $[-2.05, 0.22]$} & $+1.70$$^{*}$\,{\scriptsize $[0.29, 3.21]$} \\
Gemma & $\mathcal{S}_{6}$ & 21 & IA\,0.29, IB\,0.45, S\,0.26 & IA\,0.50, IB\,0.50 & \multicolumn{2}{c}{same models as with all eight experts} \\
Gemma & $\mathcal{S}_{7}$ & 28 & IA\,0.29, IB\,0.45, S\,0.26 & IA\,0.50, IB\,0.50 & \multicolumn{2}{c}{same models as with all eight experts} \\
\addlinespace[2pt]
Llama & $\mathcal{S}_{5}^{A}$ & 15 & MA\,0.08, CA\,0.36, S\,0.56 & IA\,0.50, S\,0.50 & $-0.08$\phantom{$^{*}$}\,{\scriptsize $[-1.54, 1.37]$} & $+2.12$$^{*}$\,{\scriptsize $[0.49, 3.78]$} \\
Llama & $\mathcal{S}_{5}^{B}$ & 15 & IB\,0.70, S\,0.30 & IB\,0.50, S\,0.50 & \multicolumn{2}{c}{same models as with all eight experts} \\
Llama & $\mathcal{S}_{6}$ & 21 & IB\,0.70, S\,0.30 & IB\,0.50, S\,0.50 & \multicolumn{2}{c}{same models as with all eight experts} \\
Llama & $\mathcal{S}_{7}$ & 28 & IB\,0.70, S\,0.30 & IB\,0.50, S\,0.50 & \multicolumn{2}{c}{same models as with all eight experts} \\
\bottomrule
\end{tabular}

\caption{Selection within expert subsets with $T(T+1)/2$ evaluations. Coefficients use IA/IB (Instruction A/B), MA (Math A), CA (Code A), ML (Multilingual), and S (Safety). Differences are in AVG points with paired 95\% intervals ($^{*}$: excluding zero); both models are measured on the same GPU node.}
\label{tab:reb-e21}
\end{table}

\FloatBarrier
\subsection{Prediction Error at and near the Selected Solutions}
\label{app:reb-local}

\paragraph{Selected point.}
Table~\ref{tab:reb-selerr} compares the surrogate prediction $\widehat{M}(\widehat{\lambda})$ with the measured $M(\widehat{\lambda})$.
The AVG prediction is optimistic by 2.17 (Gemma) and 1.41 (Llama) points, slightly more than the held-out MAE of Section~\ref{sec:fidelity} (1.93 and 1.07), as expected when selecting the maximizer of a surrogate; for SLIM$_{\mathrm{R}}$ (Appendix~\ref{app:reb-stability}), it is optimistic by 2.99 and 1.76 points.
The largest domain errors are Gemma safety ($+11.59$) and Llama multilingual ($+10.09$), with the same sign for SLIM$_{\mathrm{R}}$ ($+12.25$ and $+7.60$).

\paragraph{Local validation.}
We evaluate 16 mixtures per architecture around SLIM's solution, $(1-\epsilon)\widehat{\lambda}+\epsilon e_j$ for $j=1,\ldots,8$ and $\epsilon\in\{0.05,0.10\}$, none of which is used for construction or selection (Table~\ref{tab:reb-e22}).
The AVG MAE is 1.79 on Gemma (maximum 2.45) and 0.75 on Llama (maximum 2.05), similar to the held-out MAE; on Gemma, the errors are mostly optimistic (mean $+1.55$), mainly through safety.
At $\widehat{\lambda}$ itself, measured on the same GPU nodes, the prediction is optimistic by 1.89 (Gemma) and 1.32 (Llama) points.

\begin{table}[htbp]
\centering
\footnotesize
\setlength{\tabcolsep}{4pt}
\begin{tabular}{@{}lrrrrrr@{}}
\toprule
 & \multicolumn{3}{c}{Gemma-2-9B} & \multicolumn{3}{c}{Llama-3.1-8B} \\
\cmidrule(lr){2-4}\cmidrule(l){5-7}
Domain & $\widehat{M}(\widehat{\lambda})$ & $M(\widehat{\lambda})$ & Error & $\widehat{M}(\widehat{\lambda})$ & $M(\widehat{\lambda})$ & Error \\
\midrule
Instruction & 57.25 & 60.26 & $-3.01$ & 41.22 & 38.08 & $+3.14$ \\
Math & 72.56 & 71.27 & $+1.30$ & 74.13 & 74.22 & $-0.09$ \\
Coding & 58.43 & 58.40 & $+0.03$ & 53.05 & 57.00 & $-3.95$ \\
Multilingual & 57.40 & 56.45 & $+0.95$ & 42.58 & 32.48 & $+10.09$ \\
Safety & 95.96 & 84.38 & $+11.59$ & 79.09 & 81.25 & $-2.16$ \\
\midrule
AVG & 68.32 & 66.15 & $+2.17$ & 58.01 & 56.61 & $+1.41$ \\
AVG, SLIM$_{\mathrm{R}}$ & 68.54 & 65.55 & $+2.99$ & 58.74 & 56.98 & $+1.76$ \\
\bottomrule
\end{tabular}

\caption{Surrogate prediction $\widehat{M}(\widehat{\lambda})$ and measured score $M(\widehat{\lambda})$ at SLIM's selected coefficients (measured as in Table~\ref{tab:main}), with the error $\widehat{M}(\widehat{\lambda})-M(\widehat{\lambda})$. The last row is SLIM$_{\mathrm{R}}$ (Appendix~\ref{app:reb-stability}), measured on the GPU node of Appendix~\ref{app:reb-protocol}.}
\label{tab:reb-selerr}
\end{table}

\begin{table}[htbp]
\centering
\footnotesize
\setlength{\tabcolsep}{3pt}
\begin{tabular}{@{}lrrrrrrrr@{}}
\toprule
 & \multicolumn{4}{c}{Gemma-2-9B} & \multicolumn{4}{c}{Llama-3.1-8B} \\
\cmidrule(lr){2-5}\cmidrule(l){6-9}
Points & \#Points & MAE & Max & Mean error & \#Points & MAE & Max & Mean error \\
\midrule
All points & 16 & 1.79 & 2.45 & $+1.55$ & 16 & 0.75 & 2.05 & $-0.07$ \\
\midrule
$\epsilon=0.05$ & 8 & 1.84 & 2.45 & $+1.84$ & 8 & 0.60 & 1.81 & $+0.29$ \\
$\epsilon=0.10$ & 8 & 1.73 & 2.15 & $+1.26$ & 8 & 0.90 & 2.05 & $-0.42$ \\
Toward an expert in the support & 6 & 1.97 & 2.45 & $+1.97$ & 4 & 1.00 & 1.81 & $+0.89$ \\
Toward an expert outside the support & 10 & 1.68 & 2.08 & $+1.30$ & 12 & 0.67 & 2.05 & $-0.38$ \\
\bottomrule
\end{tabular}

\caption{AVG prediction error at 16 local points around SLIM's selected coefficients (Gemma measured on an H200 NVL GPU node, Llama on an H100 PCIe GPU node). The mean error is prediction minus measurement. Directions toward an expert in the support of $\widehat{\lambda}$ shift weight among the selected experts; the others add a small weight to an unselected expert.}
\label{tab:reb-e22}
\end{table}

\FloatBarrier
\subsection{Prediction Error on Sparse Mixtures}
\label{app:reb-sparse}

SLIM's solutions use two or three experts, so we compare the lattice surrogate with the random-design ridge surrogates of Section~\ref{sec:matched-budget} on sparse held-out mixtures (Table~\ref{tab:reb-predictor}).
Edge points, equal-weight mixtures of three experts, and the local points lie outside the interior region covered by random designs, and the lattice surrogate has lower MAE than the ridge surrogates' three-seed mean on all of them.
On the Gemma local points, one ridge seed is more accurate (1.19 versus 1.79), mainly because of a smaller mean offset (centered MAE 0.66 for SLIM versus 0.76--0.97 for ridge).
On interior mixtures of all eight experts, where both surrogates interpolate, the lattice surrogate remains more accurate (1.81 versus 3.06 on Gemma and 0.99 versus 1.98 on Llama).

\begin{table}[htbp]
\centering
\footnotesize
\setlength{\tabcolsep}{3pt}
\begin{tabular}{@{}lcrrrr@{}}
\toprule
 & & \multicolumn{2}{c}{Gemma-2-9B} & \multicolumn{2}{c}{Llama-3.1-8B} \\
\cmidrule(lr){3-4}\cmidrule(l){5-6}
Held-out points & \#Points & SLIM & Random + ridge & SLIM & Random + ridge \\
\midrule
Edge points ($t=1/3,2/3$), 2 experts & 56 & 1.74 & $11.04_{\pm1.43}$ & 1.38 & $10.88_{\pm0.17}$ \\
Equal-weight mixtures of 3 experts & 56 & 1.96 & $5.84_{\pm1.73}$ & 1.34 & $4.70_{\pm0.18}$ \\
Local points around $\widehat{\lambda}$ (Table~\ref{tab:reb-e22}) & 16 & 1.79 & $2.36_{\pm1.53}$ & 0.75 & $9.36_{\pm6.62}$ \\
Interior mixtures of 8 experts$^{\ddagger}$ & 37--42 & 1.81 & $3.06_{\pm0.49}$ & 0.99 & $1.98_{\pm0.21}$ \\
\bottomrule
\end{tabular}

\caption{AVG MAE of the lattice surrogate (SLIM) and random design + ridge (mean $\pm$ standard deviation over the three seeds of Table~\ref{tab:matched-budget}) on held-out mixtures. Edge points and three-expert mixtures are measurements from the cubic-lattice run (Appendix~\ref{app:cubic-extension}). Local and interior points are the measurements of Appendices~\ref{app:reb-local} and~\ref{app:reb-subsets}. $^{\ddagger}$The first 50 mixtures of Section~\ref{sec:fidelity}, re-measured, excluding each seed's ridge training points; SLIM is averaged over the same per-seed point sets.}
\label{tab:reb-predictor}
\end{table}

\FloatBarrier

\clearpage
\section{Additional Details on Ablation Studies}
\label{app:reb-ablation}

This appendix details the comparisons of Sections~\ref{sec:surrogate_selection}, \ref{sec:matched-budget}, and~\ref{sec:discussion}; the methods are described in Appendix~\ref{app:coefficient-implementations}, and the hardware and intervals in Appendix~\ref{app:reb-protocol}.

\subsection{Matched-Budget Comparison on the Simplex}
\label{app:reb-simplex}

\begin{figure}[htbp]
\centering
\includegraphics[width=\linewidth]{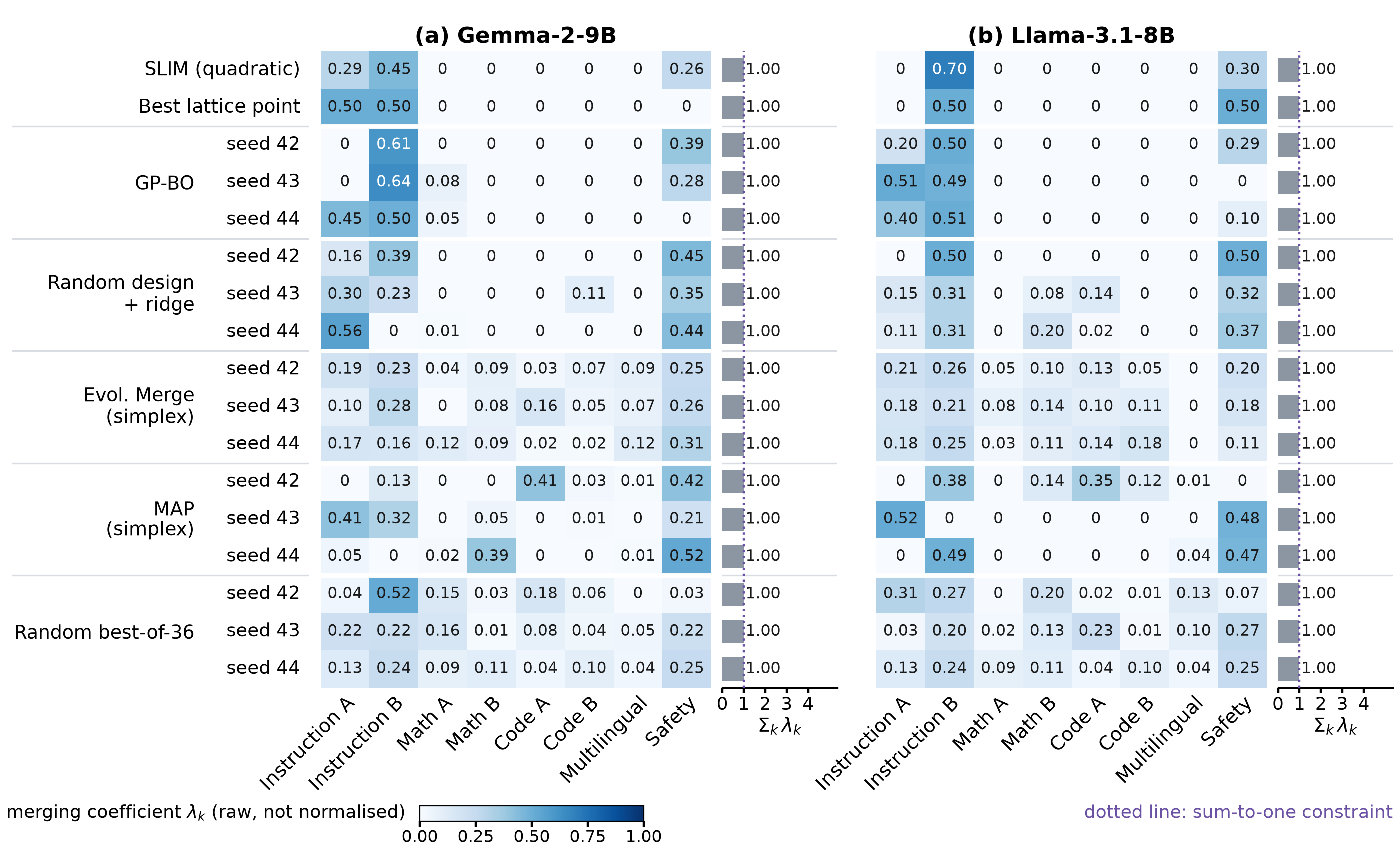}
\caption{Coefficients selected by the methods of Table~\ref{tab:matched-budget} (eight experts, the simplex, and 36 ID evaluations), drawn as in Figure~\ref{fig:lambda-profile}; all coefficient sums are one. Seeded methods show seeds 42/43/44. Cell labels are rounded, so displayed zeros need not be exact zeros.}
\label{fig:lambda-simplex36}
\end{figure}

Table~\ref{tab:reb-avg-ci} gives the AVG differences from SLIM with intervals and per-seed values, Table~\ref{tab:reb-domain-scores} the domain scores of all compared models, and Table~\ref{tab:reb-domain-ci} the domain-wise differences with intervals.
The differences between SLIM and the best lattice point reflect trade-offs across domains.
On Gemma, SLIM is higher on safety ($+14.38$ ID, $+5.34$ OOD) and lower on instruction following ($-11.28$ IFEval, $-9.51$ AlpacaEval LC).
On Llama, it is higher on ID math and coding ($+5.31$ and $+3.40$) and lower on multilingual ($-17.86$ ID, $-17.62$ OOD); its OOD lead comes mainly from AlpacaEval LC ($+24.76$; the best lattice point scores 10.03) and safety ($+8.28$).
MAP (simplex) varies strongly across seeds, with SLIM's per-seed lead ranging from $+0.98$ to $+12.31$ on Gemma ID and from $-0.93$ to $+12.22$ on Llama ID.
GP-BO selects mixtures of two or three experts in all six runs, with largest coefficients of 0.50--0.64 (Figure~\ref{fig:lambda-simplex36}).

\paragraph{Re-checks of selected points.}
Table~\ref{tab:reb-recheck} re-measures the points selected by evolutionary merging and GP-BO once on another GPU node.
The re-checked ID AVG differs from the in-run value by 0.00 to 0.04 points on Gemma and by $-0.23$ to $+0.37$ points on Llama.
This shows that the measurements are reproducible; because the same benchmark items are reused, it does not remove the optimism of selecting the best of 36 measured values.

\begin{table}[htbp]
\centering
\scriptsize
\setlength{\tabcolsep}{2pt}
\begin{tabular}{@{}lrrrr@{}}
\toprule
 & \multicolumn{2}{c}{Gemma-2-9B} & \multicolumn{2}{c}{Llama-3.1-8B} \\
\cmidrule(lr){2-3}\cmidrule(lr){4-5}
SLIM $-$ method & ID & OOD & ID & OOD \\
\midrule
Best lattice point & $+1.08$\phantom{$^{*}$}\,{\scriptsize $[-0.13, 2.35]$} & $-1.58$$^{*}$\,{\scriptsize $[-2.93, -0.14]$} & $-1.04$\phantom{$^{*}$}\,{\scriptsize $[-2.35, 0.33]$} & $+4.05$$^{*}$\,{\scriptsize $[2.53, 5.56]$} \\
\addlinespace[2pt]
GP-BO & $+1.17$$^{*}$\,{\scriptsize $[0.28, 2.10]$} & $-2.14$$^{*}$\,{\scriptsize $[-3.19, -1.00]$} & $-2.05$$^{*}$\,{\scriptsize $[-3.23, -0.81]$} & $+4.47$$^{*}$\,{\scriptsize $[3.27, 5.74]$} \\
\quad {\scriptsize per seed} & {\scriptsize $+0.57$\,/\,$+1.23$\,/\,$+1.72$} & {\scriptsize $-2.39$\,/\,$-2.69$\,/\,$-1.33$} & {\scriptsize $-2.26$\,/\,$-0.71$\,/\,$-3.17$} & {\scriptsize $+2.90$\,/\,$+6.57$\,/\,$+3.96$} \\
\addlinespace[2pt]
Random design + ridge & $+1.70$$^{*}$\,{\scriptsize $[0.55, 2.85]$} & $+3.44$$^{*}$\,{\scriptsize $[2.03, 4.87]$} & $+0.37$\phantom{$^{*}$}\,{\scriptsize $[-0.85, 1.65]$} & $+5.80$$^{*}$\,{\scriptsize $[4.43, 7.26]$} \\
\quad {\scriptsize per seed} & {\scriptsize $+0.51$\,/\,$+0.87$\,/\,$+3.72$} & {\scriptsize $+0.64$\,/\,$+3.23$\,/\,$+6.46$} & {\scriptsize $-1.36$\,/\,$+1.33$\,/\,$+1.13$} & {\scriptsize $+4.80$\,/\,$+5.75$\,/\,$+6.83$} \\
\addlinespace[2pt]
Evol.\ Merge (simplex) & $+5.55$$^{*}$\,{\scriptsize $[4.35, 6.75]$} & $+4.15$$^{*}$\,{\scriptsize $[2.69, 5.60]$} & $+5.70$$^{*}$\,{\scriptsize $[4.21, 7.24]$} & $+8.46$$^{*}$\,{\scriptsize $[6.84, 10.11]$} \\
\quad {\scriptsize per seed} & {\scriptsize $+4.50$\,/\,$+5.90$\,/\,$+6.23$} & {\scriptsize $+3.63$\,/\,$+2.36$\,/\,$+6.46$} & {\scriptsize $+4.44$\,/\,$+6.53$\,/\,$+6.14$} & {\scriptsize $+6.98$\,/\,$+9.13$\,/\,$+9.26$} \\
\addlinespace[2pt]
MAP (simplex) & $+7.58$$^{*}$\,{\scriptsize $[6.48, 8.68]$} & $+6.56$$^{*}$\,{\scriptsize $[5.21, 7.94]$} & $+7.39$$^{*}$\,{\scriptsize $[6.15, 8.71]$} & $+10.67$$^{*}$\,{\scriptsize $[9.26, 12.12]$} \\
\quad {\scriptsize per seed} & {\scriptsize $+9.44$\,/\,$+0.98$\,/\,$+12.31$} & {\scriptsize $+5.00$\,/\,$+2.05$\,/\,$+12.62$} & {\scriptsize $+10.88$\,/\,$+12.22$\,/\,$-0.93$} & {\scriptsize $+12.89$\,/\,$+13.79$\,/\,$+5.33$} \\
\addlinespace[2pt]
Random best-of-36 & $+7.18$$^{*}$\,{\scriptsize $[6.02, 8.38]$} & $+6.33$$^{*}$\,{\scriptsize $[4.88, 7.79]$} & $+5.93$$^{*}$\,{\scriptsize $[4.48, 7.43]$} & $+8.41$$^{*}$\,{\scriptsize $[6.84, 9.98]$} \\
\quad {\scriptsize per seed} & {\scriptsize $+8.04$\,/\,$+7.03$\,/\,$+6.45$} & {\scriptsize $+8.96$\,/\,$+5.28$\,/\,$+4.75$} & {\scriptsize $+5.75$\,/\,$+5.91$\,/\,$+6.12$} & {\scriptsize $+9.65$\,/\,$+7.04$\,/\,$+8.53$} \\
\bottomrule
\end{tabular}

\caption{SLIM minus each matched-budget method in AVG points (Table~\ref{tab:matched-budget}). Brackets give item-level paired bootstrap 95\% intervals; $^{*}$ marks intervals excluding zero. For seeded methods, the difference is from the three-seed mean, its interval resamples items with the seeds fixed, and the second row lists per-seed differences (seeds 42 / 43 / 44).}
\label{tab:reb-avg-ci}
\end{table}

\begin{table}[htbp]
\centering
\scriptsize
\setlength{\tabcolsep}{2.4pt}
\begin{tabular}{@{}lrrrrrrrrrrrrr@{}}
\toprule
 & & \multicolumn{6}{c}{ID (used for selection)} & \multicolumn{6}{c}{OOD (held out)} \\
\cmidrule(lr){3-8}\cmidrule(l){9-14}
Method & \#Evals & Instr. & Math & Coding & Multi. & Safety & AVG & Instr. & Math & Coding & Multi. & Safety & AVG \\
\midrule
\multicolumn{14}{@{}l}{\emph{Gemma-2-9B}} \\
SLIM (quadratic) & 36 & 61.0 & 71.6 & 59.0 & 56.5 & 84.1 & 66.43 & 38.8 & 41.6 & 59.1 & 54.6 & 58.6 & 50.56 \\
Best lattice point & 36 & 72.3 & 69.8 & 58.4 & 56.5 & 69.7 & 65.34 & 48.3 & 43.2 & 61.0 & 54.9 & 53.3 & 52.13 \\
GP-BO & 36 & 60.3 & 71.9 & 57.5 & 56.2 & 80.4 & 65.25 & 45.4 & 45.5 & 60.0 & 54.6 & 58.0 & 52.69 \\
Random design + ridge & 36 & 45.0 & 71.6 & 55.7 & 55.0 & 96.4 & 64.73 & 21.4 & 38.4 & 55.5 & 53.6 & 66.6 & 47.11 \\
Evol.\ Merge (simplex) & 36 & 39.1 & 67.5 & 56.8 & 55.4 & 85.6 & 60.88 & 17.9 & 43.5 & 59.1 & 54.1 & 57.5 & 46.41 \\
MAP (simplex) & 36 & 35.9 & 65.8 & 52.4 & 53.6 & 86.5 & 58.85 & 17.2 & 38.5 & 52.6 & 52.5 & 59.1 & 44.00 \\
Random best-of-36 & 36 & 42.1 & 70.9 & 56.9 & 55.8 & 70.6 & 59.25 & 24.1 & 45.1 & 45.3 & 54.1 & 52.6 & 44.23 \\
\cmidrule(l){1-14}
SLIM$_{\mathrm{R}}$ & 36 & 57.9 & 71.2 & 58.6 & 56.3 & 83.8 & 65.55 & 39.0 & 41.4 & 59.1 & 54.6 & 58.7 & 50.58 \\
Evol.\ Merge ($[0,1]^8$) & 36 & 46.0 & 67.0 & 48.5 & 48.4 & 89.9 & 59.99 & 23.7 & 41.3 & 28.3 & 46.9 & 54.0 & 38.84 \\
MAP ($n=45$) & 45 & 44.7 & 63.2 & 48.5 & 50.4 & 78.5 & 57.04 & 24.3 & 38.6 & 38.2 & 48.8 & 54.7 & 40.91 \\
Evol.\ Merge (gen$=$20) & 200 & 62.9 & 70.3 & 54.7 & 54.2 & 94.4 & 67.29 & 38.0 & 42.7 & 54.3 & 52.6 & 56.7 & 48.85 \\
\midrule
\multicolumn{14}{@{}l}{\emph{Llama-3.1-8B}} \\
SLIM (quadratic) & 36 & 37.9 & 74.6 & 57.2 & 31.8 & 80.3 & 56.37 & 34.8 & 36.2 & 57.9 & 31.9 & 75.6 & 47.28 \\
Best lattice point & 36 & 36.4 & 69.3 & 53.8 & 49.7 & 77.8 & 57.40 & 10.0 & 33.2 & 56.1 & 49.5 & 67.3 & 43.23 \\
GP-BO & 36 & 50.2 & 72.5 & 56.9 & 49.5 & 62.9 & 58.41 & 26.6 & 33.1 & 55.7 & 50.3 & 48.4 & 42.81 \\
Random design + ridge & 36 & 33.8 & 69.1 & 54.7 & 50.7 & 71.8 & 56.00 & 10.4 & 29.8 & 54.5 & 50.1 & 62.6 & 41.48 \\
Evol.\ Merge (simplex) & 36 & 32.7 & 69.2 & 54.6 & 51.9 & 44.9 & 50.66 & 11.6 & 31.7 & 56.5 & 50.8 & 43.4 & 38.82 \\
MAP (simplex) & 36 & 30.8 & 64.6 & 53.7 & 50.8 & 44.9 & 48.97 & 6.3 & 27.8 & 54.1 & 49.8 & 45.0 & 36.61 \\
Random best-of-36 & 36 & 30.6 & 68.8 & 54.6 & 51.5 & 46.7 & 50.44 & 13.1 & 29.9 & 54.5 & 50.6 & 46.3 & 38.88 \\
\cmidrule(l){1-14}
SLIM$_{\mathrm{R}}$ & 36 & 36.4 & 74.3 & 57.4 & 35.5 & 81.2 & 56.98 & 32.9 & 34.2 & 57.3 & 35.7 & 75.0 & 47.04 \\
Evol.\ Merge ($[0,1]^8$) & 36 & 41.2 & 71.2 & 52.3 & 48.8 & 78.9 & 58.48 & 13.8 & 30.6 & 65.7 & 47.3 & 47.2 & 40.90 \\
MAP ($n=45$) & 45 & 32.7 & 57.4 & 48.3 & 42.9 & 55.0 & 47.26 & 19.8 & 25.6 & 59.3 & 41.5 & 38.9 & 37.01 \\
Evol.\ Merge (gen$=$20) & 200 & 49.4 & 78.5 & 53.3 & 50.0 & 90.0 & 64.23 & 27.6 & 37.9 & 61.8 & 48.6 & 52.6 & 45.71 \\
\bottomrule
\end{tabular}

\caption{Domain scores (0--100) of the final models of the 11 methods compared in Section~\ref{sec:matched-budget} and Appendices~\ref{app:reb-ablation} and~\ref{app:reb-safety}. In each architecture, the upper block lists the methods of Table~\ref{tab:matched-budget}; the lower block lists SLIM$_{\mathrm{R}}$ (Appendix~\ref{app:reb-stability}) and models from the $[0,1]^8$ runs of Table~\ref{tab:main}. OOD instruction following is the AlpacaEval 2.0 length-controlled win rate. Seeded methods show three-seed means. All models of an architecture are evaluated on the same GPU node.}
\label{tab:reb-domain-scores}
\end{table}

\begin{table}[htbp]
\centering
\scriptsize
\setlength{\tabcolsep}{2pt}
\begin{tabular}{@{}lrrrr@{}}
\toprule
 & \multicolumn{2}{c}{Gemma-2-9B} & \multicolumn{2}{c}{Llama-3.1-8B} \\
\cmidrule(lr){2-3}\cmidrule(lr){4-5}
Domain & ID & OOD & ID & OOD \\
\midrule
\multicolumn{5}{@{}l}{\emph{SLIM $-$ Best lattice point}} \\
\quad Instruction & $-11.28$$^{*}$\,{\scriptsize $[-14.97, -7.58]$} & $-9.51$$^{*}$\,{\scriptsize $[-13.26, -5.49]$} & $+1.48$\phantom{$^{*}$}\,{\scriptsize $[-2.22, 5.18]$} & $+24.76$$^{*}$\,{\scriptsize $[21.74, 27.76]$} \\
\quad Math & $+1.74$\phantom{$^{*}$}\,{\scriptsize $[-0.15, 3.71]$} & $-1.60$\phantom{$^{*}$}\,{\scriptsize $[-4.20, 1.20]$} & $+5.31$$^{*}$\,{\scriptsize $[3.26, 7.43]$} & $+3.00$\phantom{$^{*}$}\,{\scriptsize $[-0.40, 6.60]$} \\
\quad Coding & $+0.60$\phantom{$^{*}$}\,{\scriptsize $[-1.80, 3.00]$} & $-1.83$\phantom{$^{*}$}\,{\scriptsize $[-6.71, 2.44]$} & $+3.40$$^{*}$\,{\scriptsize $[0.80, 6.00]$} & $+1.83$\phantom{$^{*}$}\,{\scriptsize $[-3.05, 7.32]$} \\
\quad Multilingual & $-0.03$\phantom{$^{*}$}\,{\scriptsize $[-0.61, 0.55]$} & $-0.28$$^{*}$\,{\scriptsize $[-0.46, -0.11]$} & $-17.86$$^{*}$\,{\scriptsize $[-20.04, -15.58]$} & $-17.62$$^{*}$\,{\scriptsize $[-18.34, -16.90]$} \\
\quad Safety & $+14.38$$^{*}$\,{\scriptsize $[10.62, 18.44]$} & $+5.34$$^{*}$\,{\scriptsize $[2.94, 7.74]$} & $+2.50$\phantom{$^{*}$}\,{\scriptsize $[-1.25, 6.25]$} & $+8.28$$^{*}$\,{\scriptsize $[5.34, 11.35]$} \\
\quad AVG & $+1.08$\phantom{$^{*}$}\,{\scriptsize $[-0.13, 2.35]$} & $-1.58$$^{*}$\,{\scriptsize $[-2.93, -0.14]$} & $-1.04$\phantom{$^{*}$}\,{\scriptsize $[-2.35, 0.33]$} & $+4.05$$^{*}$\,{\scriptsize $[2.53, 5.56]$} \\
\midrule
\multicolumn{5}{@{}l}{\emph{SLIM $-$ GP-BO}} \\
\quad Instruction & $+0.68$\phantom{$^{*}$}\,{\scriptsize $[-2.28, 3.64]$} & $-6.55$$^{*}$\,{\scriptsize $[-9.81, -3.09]$} & $-12.26$$^{*}$\,{\scriptsize $[-15.47, -9.12]$} & $+8.23$$^{*}$\,{\scriptsize $[6.16, 10.66]$} \\
\quad Math & $-0.30$\phantom{$^{*}$}\,{\scriptsize $[-1.82, 1.21]$} & $-3.93$$^{*}$\,{\scriptsize $[-6.53, -1.40]$} & $+2.07$$^{*}$\,{\scriptsize $[0.45, 3.69]$} & $+3.13$$^{*}$\,{\scriptsize $[0.27, 6.00]$} \\
\quad Coding & $+1.53$\phantom{$^{*}$}\,{\scriptsize $[-0.20, 3.33]$} & $-0.81$\phantom{$^{*}$}\,{\scriptsize $[-4.07, 2.24]$} & $+0.27$\phantom{$^{*}$}\,{\scriptsize $[-1.67, 2.20]$} & $+2.24$\phantom{$^{*}$}\,{\scriptsize $[-1.83, 6.50]$} \\
\quad Multilingual & $+0.32$\phantom{$^{*}$}\,{\scriptsize $[-0.08, 0.72]$} & $+0.02$\phantom{$^{*}$}\,{\scriptsize $[-0.10, 0.14]$} & $-17.70$$^{*}$\,{\scriptsize $[-19.90, -15.42]$} & $-18.38$$^{*}$\,{\scriptsize $[-19.08, -17.68]$} \\
\quad Safety & $+3.65$$^{*}$\,{\scriptsize $[1.15, 6.25]$} & $+0.58$\phantom{$^{*}$}\,{\scriptsize $[-1.11, 2.27]$} & $+17.40$$^{*}$\,{\scriptsize $[13.65, 21.25]$} & $+27.15$$^{*}$\,{\scriptsize $[24.39, 29.95]$} \\
\quad AVG & $+1.17$$^{*}$\,{\scriptsize $[0.28, 2.10]$} & $-2.14$$^{*}$\,{\scriptsize $[-3.19, -1.00]$} & $-2.05$$^{*}$\,{\scriptsize $[-3.23, -0.81]$} & $+4.47$$^{*}$\,{\scriptsize $[3.27, 5.74]$} \\
\midrule
\multicolumn{5}{@{}l}{\emph{SLIM $-$ Random design + ridge}} \\
\quad Instruction & $+15.96$$^{*}$\,{\scriptsize $[12.57, 19.41]$} & $+17.38$$^{*}$\,{\scriptsize $[13.71, 21.15]$} & $+4.13$$^{*}$\,{\scriptsize $[0.74, 7.46]$} & $+24.35$$^{*}$\,{\scriptsize $[21.89, 27.23]$} \\
\quad Math & $-0.03$\phantom{$^{*}$}\,{\scriptsize $[-1.72, 1.67]$} & $+3.20$$^{*}$\,{\scriptsize $[0.47, 5.93]$} & $+5.53$$^{*}$\,{\scriptsize $[3.69, 7.38]$} & $+6.40$$^{*}$\,{\scriptsize $[3.40, 9.40]$} \\
\quad Coding & $+3.33$$^{*}$\,{\scriptsize $[0.93, 5.67]$} & $+3.66$\phantom{$^{*}$}\,{\scriptsize $[-1.22, 8.74]$} & $+2.47$$^{*}$\,{\scriptsize $[0.07, 4.80]$} & $+3.46$\phantom{$^{*}$}\,{\scriptsize $[-1.42, 8.54]$} \\
\quad Multilingual & $+1.53$$^{*}$\,{\scriptsize $[0.89, 2.18]$} & $+0.99$$^{*}$\,{\scriptsize $[0.83, 1.14]$} & $-18.84$$^{*}$\,{\scriptsize $[-21.02, -16.47]$} & $-18.22$$^{*}$\,{\scriptsize $[-18.95, -17.48]$} \\
\quad Safety & $-12.29$$^{*}$\,{\scriptsize $[-15.94, -8.75]$} & $-8.01$$^{*}$\,{\scriptsize $[-10.15, -6.01]$} & $+8.54$$^{*}$\,{\scriptsize $[5.00, 12.19]$} & $+13.00$$^{*}$\,{\scriptsize $[10.01, 16.07]$} \\
\quad AVG & $+1.70$$^{*}$\,{\scriptsize $[0.55, 2.85]$} & $+3.44$$^{*}$\,{\scriptsize $[2.03, 4.87]$} & $+0.37$\phantom{$^{*}$}\,{\scriptsize $[-0.85, 1.65]$} & $+5.80$$^{*}$\,{\scriptsize $[4.43, 7.26]$} \\
\midrule
\multicolumn{5}{@{}l}{\emph{SLIM $-$ Evol.\ Merge (simplex)}} \\
\quad Instruction & $+21.93$$^{*}$\,{\scriptsize $[18.24, 25.69]$} & $+20.97$$^{*}$\,{\scriptsize $[17.04, 24.96]$} & $+5.24$$^{*}$\,{\scriptsize $[1.42, 9.12]$} & $+23.14$$^{*}$\,{\scriptsize $[20.43, 25.99]$} \\
\quad Math & $+4.07$$^{*}$\,{\scriptsize $[2.25, 5.94]$} & $-1.87$\phantom{$^{*}$}\,{\scriptsize $[-4.80, 1.00]$} & $+5.38$$^{*}$\,{\scriptsize $[3.34, 7.46]$} & $+4.47$$^{*}$\,{\scriptsize $[1.20, 7.73]$} \\
\quad Coding & $+2.20$\phantom{$^{*}$}\,{\scriptsize $[-0.33, 4.87]$} & $+0.00$\phantom{$^{*}$}\,{\scriptsize $[-4.88, 4.88]$} & $+2.60$\phantom{$^{*}$}\,{\scriptsize $[0.00, 5.20]$} & $+1.42$\phantom{$^{*}$}\,{\scriptsize $[-4.67, 7.52]$} \\
\quad Multilingual & $+1.09$$^{*}$\,{\scriptsize $[0.51, 1.71]$} & $+0.53$$^{*}$\,{\scriptsize $[0.37, 0.69]$} & $-20.12$$^{*}$\,{\scriptsize $[-22.50, -17.50]$} & $-18.88$$^{*}$\,{\scriptsize $[-19.63, -18.11]$} \\
\quad Safety & $-1.56$\phantom{$^{*}$}\,{\scriptsize $[-5.21, 1.98]$} & $+1.11$\phantom{$^{*}$}\,{\scriptsize $[-1.11, 3.29]$} & $+35.42$$^{*}$\,{\scriptsize $[30.42, 40.42]$} & $+32.13$$^{*}$\,{\scriptsize $[28.66, 35.47]$} \\
\quad AVG & $+5.55$$^{*}$\,{\scriptsize $[4.35, 6.75]$} & $+4.15$$^{*}$\,{\scriptsize $[2.69, 5.60]$} & $+5.70$$^{*}$\,{\scriptsize $[4.21, 7.24]$} & $+8.46$$^{*}$\,{\scriptsize $[6.84, 10.11]$} \\
\midrule
\multicolumn{5}{@{}l}{\emph{SLIM $-$ MAP (simplex)}} \\
\quad Instruction & $+25.08$$^{*}$\,{\scriptsize $[21.87, 28.34]$} & $+21.57$$^{*}$\,{\scriptsize $[18.19, 25.26]$} & $+7.09$$^{*}$\,{\scriptsize $[3.88, 10.29]$} & $+28.44$$^{*}$\,{\scriptsize $[25.58, 31.47]$} \\
\quad Math & $+5.74$$^{*}$\,{\scriptsize $[3.99, 7.53]$} & $+3.13$$^{*}$\,{\scriptsize $[0.33, 5.93]$} & $+9.98$$^{*}$\,{\scriptsize $[8.09, 11.88]$} & $+8.40$$^{*}$\,{\scriptsize $[5.27, 11.60]$} \\
\quad Coding & $+6.60$$^{*}$\,{\scriptsize $[4.07, 9.13]$} & $+6.50$$^{*}$\,{\scriptsize $[1.83, 11.18]$} & $+3.47$$^{*}$\,{\scriptsize $[1.20, 5.73]$} & $+3.86$\phantom{$^{*}$}\,{\scriptsize $[-1.02, 8.74]$} \\
\quad Multilingual & $+2.86$$^{*}$\,{\scriptsize $[2.19, 3.59]$} & $+2.10$$^{*}$\,{\scriptsize $[1.93, 2.27]$} & $-18.99$$^{*}$\,{\scriptsize $[-21.23, -16.55]$} & $-17.93$$^{*}$\,{\scriptsize $[-18.66, -17.18]$} \\
\quad Safety & $-2.40$\phantom{$^{*}$}\,{\scriptsize $[-5.52, 0.62]$} & $-0.53$\phantom{$^{*}$}\,{\scriptsize $[-2.71, 1.60]$} & $+35.42$$^{*}$\,{\scriptsize $[31.46, 39.48]$} & $+30.57$$^{*}$\,{\scriptsize $[27.68, 33.47]$} \\
\quad AVG & $+7.58$$^{*}$\,{\scriptsize $[6.48, 8.68]$} & $+6.56$$^{*}$\,{\scriptsize $[5.21, 7.94]$} & $+7.39$$^{*}$\,{\scriptsize $[6.15, 8.71]$} & $+10.67$$^{*}$\,{\scriptsize $[9.26, 12.12]$} \\
\midrule
\multicolumn{5}{@{}l}{\emph{SLIM $-$ Random best-of-36}} \\
\quad Instruction & $+18.85$$^{*}$\,{\scriptsize $[15.34, 22.49]$} & $+14.72$$^{*}$\,{\scriptsize $[11.02, 18.52]$} & $+7.27$$^{*}$\,{\scriptsize $[3.51, 11.09]$} & $+21.64$$^{*}$\,{\scriptsize $[18.99, 24.46]$} \\
\quad Math & $+0.71$\phantom{$^{*}$}\,{\scriptsize $[-1.19, 2.65]$} & $-3.47$$^{*}$\,{\scriptsize $[-6.33, -0.60]$} & $+5.79$$^{*}$\,{\scriptsize $[3.89, 7.68]$} & $+6.33$$^{*}$\,{\scriptsize $[3.27, 9.47]$} \\
\quad Coding & $+2.13$\phantom{$^{*}$}\,{\scriptsize $[-0.27, 4.67]$} & $+13.82$$^{*}$\,{\scriptsize $[8.74, 19.11]$} & $+2.60$$^{*}$\,{\scriptsize $[0.27, 5.00]$} & $+3.46$\phantom{$^{*}$}\,{\scriptsize $[-2.03, 9.15]$} \\
\quad Multilingual & $+0.75$$^{*}$\,{\scriptsize $[0.27, 1.24]$} & $+0.52$$^{*}$\,{\scriptsize $[0.36, 0.68]$} & $-19.67$$^{*}$\,{\scriptsize $[-22.00, -17.15]$} & $-18.65$$^{*}$\,{\scriptsize $[-19.40, -17.88]$} \\
\quad Safety & $+13.44$$^{*}$\,{\scriptsize $[9.90, 17.08]$} & $+6.05$$^{*}$\,{\scriptsize $[3.87, 8.23]$} & $+33.65$$^{*}$\,{\scriptsize $[28.96, 38.33]$} & $+29.24$$^{*}$\,{\scriptsize $[25.90, 32.44]$} \\
\quad AVG & $+7.18$$^{*}$\,{\scriptsize $[6.02, 8.38]$} & $+6.33$$^{*}$\,{\scriptsize $[4.88, 7.79]$} & $+5.93$$^{*}$\,{\scriptsize $[4.48, 7.43]$} & $+8.41$$^{*}$\,{\scriptsize $[6.84, 9.98]$} \\
\bottomrule
\end{tabular}

\caption{Domain-wise differences between SLIM and each matched-budget method with paired 95\% intervals ($^{*}$: excluding zero); seeded methods as in Table~\ref{tab:reb-avg-ci}. OOD instruction following is the AlpacaEval LC win rate.}
\label{tab:reb-domain-ci}
\end{table}

\begin{table}[htbp]
\centering
\footnotesize
\setlength{\tabcolsep}{4pt}
\begin{tabular}{@{}lcrrrrrr@{}}
\toprule
 & & \multicolumn{3}{c}{Gemma-2-9B} & \multicolumn{3}{c}{Llama-3.1-8B} \\
\cmidrule(lr){3-5}\cmidrule(l){6-8}
Method & Seed & In run & Re-check & Diff. & In run & Re-check & Diff. \\
\midrule
Evol.\ Merge (simplex) & 42 & 61.92 & 61.92 & $+0.00$ & 51.93 & 52.04 & $+0.11$ \\
Evol.\ Merge (simplex) & 43 & 60.52 & 60.52 & $+0.00$ & 49.83 & 50.11 & $+0.28$ \\
Evol.\ Merge (simplex) & 44 & 60.20 & 60.20 & $+0.00$ & 50.23 & 50.33 & $+0.11$ \\
GP-BO & 42 & 65.86 & 65.86 & $+0.00$ & 58.62 & 58.58 & $-0.04$ \\
GP-BO & 43 & 65.19 & 65.19 & $+0.00$ & 57.08 & 56.85 & $-0.23$ \\
GP-BO & 44 & 64.71 & 64.75 & $+0.04$ & 59.53 & 59.90 & $+0.37$ \\
\bottomrule
\end{tabular}

\caption{ID AVG of the points selected by evolutionary merging and GP-BO on the simplex: the in-run measurement used for selection and one re-measurement on another GPU node (H200 NVL for Gemma, H100 PCIe for Llama).}
\label{tab:reb-recheck}
\end{table}

\FloatBarrier
\subsection{Effect of the Coefficient Domain}
\label{app:reb-domain}

Table~\ref{tab:reb-e11} isolates the coefficient domain: for each seed, it compares Evol.\ Merge (simplex) with the candidate that had the highest ID AVG among the first 36 evaluations of the corresponding run on $[0,1]^8$ in Table~\ref{tab:main}, evaluated on the same GPU node.
The effect of the domain changes direction with the model: on OOD, the $[0,1]^8$ solutions are 4.27--10.84 points lower on Gemma but 1.02--2.90 points higher on Llama; on ID, they are 7.32--8.37 points higher on Llama and between $-2.81$ and $+0.53$ on Gemma.
Their coefficient sums are 2.84--4.10, whereas their largest normalized coefficients (0.20--0.26) are similar to those of the simplex solutions, so the differences reflect overall scale as well as proportions.

\begin{table}[htbp]
\centering
\scriptsize
\setlength{\tabcolsep}{4pt}
\begin{tabular}{@{}lcrrccc@{}}
\toprule
 & & \multicolumn{2}{c}{$[0,1]^8$ $-$ simplex (AVG)} & \multicolumn{2}{c}{$[0,1]^8$ solution} & Simplex solution \\
\cmidrule(lr){3-4}\cmidrule(lr){5-6}\cmidrule(l){7-7}
Arch. & Seed & ID & OOD & $\sum_k\lambda_k$ & Max share & $\max_k\lambda_k$ \\
\midrule
Gemma & 42 & $-2.81$$^{*}$\,{\scriptsize $[-4.23, -1.41]$} & $-7.60$$^{*}$\,{\scriptsize $[-9.68, -5.58]$} & 2.84 & 0.26 & 0.25 \\
Gemma & 43 & $+0.53$\phantom{$^{*}$}\,{\scriptsize $[-0.93, 2.00]$} & $-10.84$$^{*}$\,{\scriptsize $[-12.98, -8.61]$} & 3.67 & 0.24 & 0.28 \\
Gemma & 44 & $-0.40$\phantom{$^{*}$}\,{\scriptsize $[-1.79, 1.03]$} & $-4.27$$^{*}$\,{\scriptsize $[-6.18, -2.33]$} & 3.56 & 0.25 & 0.31 \\
\addlinespace[2pt]
Llama & 42 & $+7.75$$^{*}$\,{\scriptsize $[6.15, 9.37]$} & $+2.90$$^{*}$\,{\scriptsize $[1.08, 4.74]$} & 2.93 & 0.21 & 0.26 \\
Llama & 43 & $+8.37$$^{*}$\,{\scriptsize $[6.63, 10.09]$} & $+1.02$\phantom{$^{*}$}\,{\scriptsize $[-0.95, 3.07]$} & 4.10 & 0.20 & 0.21 \\
Llama & 44 & $+7.32$$^{*}$\,{\scriptsize $[5.73, 8.94]$} & $+2.30$$^{*}$\,{\scriptsize $[0.46, 4.22]$} & 3.31 & 0.21 & 0.25 \\
\bottomrule
\end{tabular}

\caption{CMA-ES on $[0,1]^8$ versus the simplex at 36 evaluations, per seed. The $[0,1]^8$ candidate is the best of the first 36 evaluations of the run in Table~\ref{tab:main}; both are evaluated on the same GPU node. Differences are in AVG points with paired 95\% intervals ($^{*}$: excluding zero). Max share is $\max_k\lambda_k/\sum_j\lambda_j$.}
\label{tab:reb-e11}
\end{table}

\FloatBarrier
\subsection{Stability of the Selected Coefficients}
\label{app:reb-stability}

Given the 36 measurements, SLIM's coefficients are deterministic, so their variability comes from the evaluations themselves.
Item-level outputs are available only for the 36 lattice points evaluated on the GPU node of Appendix~\ref{app:reb-protocol}, so we refit SLIM on these values alone (SLIM$_{\mathrm{R}}$) and use them in this analysis; the best lattice point is the same point as in Table~\ref{tab:selection_ablation} on both architectures.
SLIM$_{\mathrm{R}}$ selects coefficients close to SLIM's (Gemma: Instruction A, Instruction B, and Safety at 0.28, 0.45, and 0.27 instead of 0.29, 0.45, and 0.26; Llama: Instruction B and Safety at 0.68 and 0.32 instead of 0.70 and 0.30).
Relative to SLIM evaluated on the same GPU node, it changes OOD AVG by $+0.02$ $[-0.49, 0.55]$ (Gemma) and $-0.24$ $[-1.07, 0.57]$ (Llama) and ID AVG by $-0.88$ $[-1.40, -0.37]$ and $+0.61$ $[-0.19, 1.41]$, so ID differences of about one point between nearby models should be read with care.

Following Section~\ref{sec:surrogate_selection}, we resample the benchmark items of these 36 evaluations 1{,}000 times (the first 1{,}000 replicates of Appendix~\ref{app:reb-protocol}), recompute the 36 scores, and measure the $\ell_1$ distance of each replicate's selection from $\lambda^{(0)}$, here the solution of SLIM$_{\mathrm{R}}$ or the best lattice point selected with all items.
Table~\ref{tab:reb-stability} shows the results.
On Gemma, SLIM keeps the same three experts in every replicate (median $\ell_1$ distance 0.031), whereas the best lattice point switches between Instruction A+B (ID AVG 65.34) and Instruction B+Safety (65.19) in 40.9\% of replicates, each switch moving the distance by 1.0; SLIM$_{\mathrm{R}}$'s solution combines both pairs.
On Llama, SLIM adds a small weight on Instruction A (median 0.07, at most 0.31) in 25.8\% of replicates; its mean distance is comparable to that of the best lattice point (0.065 versus 0.078), and it exceeds 0.1 more often (17.5\% versus 7.8\%) but 0.3 less often (4.0\% versus 7.8\%).
The resampling reflects only the finite benchmark items, not run-to-run or machine-to-machine variation, and it concerns the coefficients rather than performance.

\begin{table}[htbp]
\centering
\footnotesize
\setlength{\tabcolsep}{3.5pt}
\begin{tabular}{@{}lrrrrrr@{}}
\toprule
 & \multicolumn{4}{c}{SLIM refit on each resample} & \multicolumn{2}{c}{Best lattice point} \\
\cmidrule(lr){2-5}\cmidrule(l){6-7}
(a) & Median $\ell_1$ & Max $\ell_1$ & Support changed (\%) & Mean $\ell_1$ & Switched (\%) & Mean $\ell_1$ \\
\midrule
Gemma-2-9B & 0.031\,{\scriptsize $[0.006, 0.077]$} & 0.127 & 0.0 & 0.034 & 40.9 & 0.409 \\
Llama-3.1-8B & 0.029\,{\scriptsize $[0.002, 0.352]$} & 0.629 & 25.8 & 0.065 & 7.8 & 0.078 \\
\bottomrule
\end{tabular}

\vspace{6pt}
\begin{tabular}{@{}lrrrr@{}}
\toprule
 & \multicolumn{2}{c}{Gemma-2-9B} & \multicolumn{2}{c}{Llama-3.1-8B} \\
\cmidrule(lr){2-3}\cmidrule(l){4-5}
(b) Resamples with (\%) & SLIM & Best lattice point & SLIM & Best lattice point \\
\midrule
$\ell_1>0.05$ & 18.9 & 40.9 & 29.8 & 7.8 \\
$\ell_1>0.1$ & 0.1 & 40.9 & 17.5 & 7.8 \\
$\ell_1>0.2$ & 0.0 & 40.9 & 9.5 & 7.8 \\
$\ell_1>0.3$ & 0.0 & 40.9 & 4.0 & 7.8 \\
$\ell_1>0.4$ & 0.0 & 40.9 & 1.6 & 7.8 \\
$\ell_1>0.5$ & 0.0 & 40.9 & 0.8 & 7.8 \\
\bottomrule
\end{tabular}

\caption{Stability of the selected coefficients under 1{,}000 item-level resamples of the 36 lattice points evaluated on the GPU node of Appendix~\ref{app:reb-protocol}. (a) $\ell_1$ distance from the full-data solution; the median comes with its 2.5--97.5\% range, and the support counts experts with $\lambda_k>10^{-6}$. (b) Share of resamples whose $\ell_1$ distance exceeds a threshold; every switch of the best lattice point moves the distance by 1.0.}
\label{tab:reb-stability}
\end{table}

\FloatBarrier
\subsection{Global Scale of SLIM's Coefficients}
\label{app:reb-scale}

We keep SLIM's proportions and scale them as $\lambda=s\widehat{\lambda}$ with $s\in\{1.0,1.5,\ldots,3.5\}$, choosing $s$ by ID AVG measured on the same GPU node, which costs six ID evaluations beyond SLIM's 36 (five if the evaluation of SLIM's own solution, $s=1$, is reused; Table~\ref{tab:reb-e30}).
On Gemma, $s=1.5$ is selected (ID AVG 66.86 versus 66.43) and raises OOD AVG from 50.56 to 53.65 on the same GPU node; the choice rests on an ID difference of 0.43, and these values are point estimates without intervals.
On Llama, $s=1$ is selected, and coding collapses to at most 0.4 for $s\ge2$.
Selecting proportions and scale jointly would require, for example, a lattice at each of $K$ scales ($36K$ evaluations) or a single surrogate over both the mixture and its amount; we leave this to future work.

\begin{table}[htbp]
\centering
\footnotesize
\setlength{\tabcolsep}{4pt}
\begin{tabular}{@{}lrrrrrr@{}}
\toprule
Scale & Instr. & Math & Coding & Multi. & Safety & AVG \\
\midrule
\multicolumn{7}{@{}l}{\emph{Gemma-2-9B}, ID} \\
\quad $s=1.0$ & 61.0 & 71.6 & 59.0 & 56.5 & 84.1 & 66.43 \\
\quad $s=1.5$ & 64.7 & 65.0 & 57.0 & 55.8 & 91.9 & \textbf{66.86} \\
\quad $s=2.0$ & 58.4 & 49.4 & 53.6 & 54.3 & 94.7 & 62.09 \\
\quad $s=2.5$ & 49.5 & 37.1 & 53.6 & 52.6 & 97.2 & 58.02 \\
\quad $s=3.0$ & 47.3 & 31.5 & 49.4 & 49.2 & 98.1 & 55.10 \\
\quad $s=3.5$ & 41.4 & 19.1 & 39.0 & 42.7 & 97.8 & 48.01 \\
\addlinespace[2pt]
\multicolumn{7}{@{}l}{\emph{Gemma-2-9B}, OOD} \\
\quad $s=1.0$ (SLIM) & 38.8 & 41.6 & 59.1 & 54.6 & 58.6 & 50.56 \\
\quad $s=1.5$ (selected) & 47.7 & 43.0 & 64.6 & 54.0 & 59.0 & 53.65 \\
\addlinespace[2pt]
\multicolumn{7}{@{}l}{\emph{Llama-3.1-8B}, ID} \\
\quad $s=1.0$ & 38.4 & 74.5 & 57.2 & 32.1 & 81.2 & \textbf{56.70} \\
\quad $s=1.5$ & 39.4 & 79.2 & 55.4 & 31.8 & 76.6 & 56.46 \\
\quad $s=2.0$ & 32.0 & 77.1 & 0.4 & 40.5 & 73.4 & 44.69 \\
\quad $s=2.5$ & 29.2 & 50.4 & 0.4 & 35.9 & 79.4 & 39.06 \\
\quad $s=3.0$ & 24.4 & 0.9 & 0.0 & 8.0 & 81.6 & 22.97 \\
\quad $s=3.5$ & 14.2 & 0.0 & 0.0 & 4.3 & 76.6 & 19.01 \\
\bottomrule
\end{tabular}

\caption{ID scores of $\lambda=s\widehat{\lambda}$ (Gemma on an H200 NVL GPU node, Llama on an H100 PCIe GPU node), with the selected scale in bold, and, for Gemma, OOD scores at the selected scale compared with SLIM on the same GPU node; Llama selects $s=1$.}
\label{tab:reb-e30}
\end{table}
\FloatBarrier

\clearpage
\section{Safety Metric and Over-Refusal}
\label{app:reb-safety}

Both safety benchmarks contain only harmful prompts (Appendix~\ref{app:benchmarks}); this appendix measures over-refusal and the sensitivity of the selection to the safety weight.

\subsection{Over-Refusal}
\label{app:reb-overrefusal}

To measure over-refusal, we evaluated all 54 final models of the 11 methods in Table~\ref{tab:reb-overrefusal} on the 940 benign WildGuardTest prompts (489 vanilla and 451 adversarial) with the same generation settings and refusal classifier, obtaining the refusal rate $R_b$; running the harmful prompts through the same pipeline reproduced SLIM's OOD safety scores (58.61 on Gemma and 75.57 on Llama).
Table~\ref{tab:reb-overrefusal} reports $R_b$, the balanced safety score $S_{\mathrm{bal}}=(R_h+100-R_b)/2$, where $R_h$ is the harmful-prompt refusal rate, and the OOD AVG with safety replaced by $S_{\mathrm{bal}}$ (AVG$_{\mathrm{bal}}$) or removed (AVG$_4$), each ranked among the 11 methods.
On Gemma, SLIM refuses 5.6\% of benign prompts and ranks fourth under both AVG and AVG$_{\mathrm{bal}}$.
On Llama, SLIM refuses 16.3\% of benign prompts (11.5\% vanilla, 21.5\% adversarial), the highest mean rate among the 11 methods, but its high $R_h$ (75.6) keeps $S_{\mathrm{bal}}$ high (79.6), and it moves from first to second, behind only evolutionary merging with 200 evaluations.
Among the methods with 36 evaluations, SLIM's rank is unchanged on both architectures, although other methods reorder.
Removing safety (AVG$_4$) moves SLIM from first to third on Llama, behind GP-BO and evolutionary merging (gen$=$20), and from fourth to third on Gemma: on Llama, SLIM trades multilingual quality (31.9) for safety (75.6).

\begin{table}[htbp]
\centering
\scriptsize
\setlength{\tabcolsep}{3.5pt}
\begin{tabular}{@{}lrrrrrrr@{}}
\toprule
Method & \#Evals & $R_h$ & $R_b$ & $S_{\mathrm{bal}}$ & AVG (rank) & AVG$_{\mathrm{bal}}$ (rank) & AVG$_{4}$ (rank) \\
\midrule
\multicolumn{8}{@{}l}{\emph{Gemma-2-9B}} \\
SLIM (quadratic) & 36 & 58.6 & 5.6 & 76.5 & 50.56 (4) & 54.13 (4) & 48.54 (3) \\
SLIM$_{\mathrm{R}}$ & 36 & 58.7 & 5.6 & 76.6 & 50.58 (3) & 54.14 (3) & 48.54 (4) \\
Best lattice point & 36 & 53.3 & 4.0 & 74.6 & 52.13 (2) & 56.40 (2) & 51.85 (1) \\
GP-BO & 36 & $58.0_{\pm5.6}$ & $4.5_{\pm0.7}$ & $76.8_{\pm2.5}$ & 52.69 (1) & 56.44 (1) & 51.36 (2) \\
Random design + ridge & 36 & $66.6_{\pm4.5}$ & $19.3_{\pm16.4}$ & $73.6_{\pm6.0}$ & 47.11 (6) & 48.52 (7) & 42.24 (7) \\
Evol.\ Merge (simplex) & 36 & $57.5_{\pm1.3}$ & $8.2_{\pm1.2}$ & $74.7_{\pm0.7}$ & 46.41 (7) & 49.84 (6) & 43.64 (6) \\
MAP (simplex) & 36 & $59.1_{\pm3.8}$ & $11.9_{\pm7.8}$ & $73.6_{\pm2.2}$ & 44.00 (9) & 46.89 (9) & 40.21 (9) \\
Random best-of-36 & 36 & $52.6_{\pm4.1}$ & $5.1_{\pm2.3}$ & $73.7_{\pm1.1}$ & 44.23 (8) & 48.46 (8) & 42.14 (8) \\
Evol.\ Merge ($[0,1]^8$, 36 evals) & 36 & $54.0_{\pm0.7}$ & $4.1_{\pm0.3}$ & $75.0_{\pm0.5}$ & 38.84 (11) & 43.02 (11) & 35.04 (11) \\
MAP ($n=45$, $[0,1]^8$) & 45 & $54.7_{\pm10.2}$ & $6.1_{\pm3.6}$ & $74.3_{\pm3.6}$ & 40.91 (10) & 44.83 (10) & 37.47 (10) \\
Evol.\ Merge (gen$=$20, $[0,1]^8$) & 200 & $56.7_{\pm0.5}$ & $4.4_{\pm0.3}$ & $76.2_{\pm0.2}$ & 48.85 (5) & 52.74 (5) & 46.89 (5) \\
\midrule
\multicolumn{8}{@{}l}{\emph{Llama-3.1-8B}} \\
SLIM (quadratic) & 36 & 75.6 & 16.3 & 79.6 & 47.28 (1) & 48.10 (2) & 40.21 (3) \\
SLIM$_{\mathrm{R}}$ & 36 & 75.0 & 15.0 & 80.0 & 47.04 (2) & 48.04 (3) & 40.04 (4) \\
Best lattice point & 36 & 67.3 & 13.2 & 77.0 & 43.23 (4) & 45.18 (6) & 37.22 (7) \\
GP-BO & 36 & $48.4_{\pm14.4}$ & $5.0_{\pm3.3}$ & $71.7_{\pm5.6}$ & 42.81 (5) & 47.46 (4) & 41.40 (2) \\
Random design + ridge & 36 & $62.6_{\pm3.3}$ & $12.3_{\pm1.4}$ & $75.2_{\pm1.6}$ & 41.48 (6) & 44.00 (7) & 36.21 (10) \\
Evol.\ Merge (simplex) & 36 & $43.4_{\pm5.2}$ & $6.4_{\pm1.9}$ & $68.5_{\pm1.8}$ & 38.82 (9) & 43.84 (8) & 37.67 (6) \\
MAP (simplex) & 36 & $45.0_{\pm21.5}$ & $12.2_{\pm10.3}$ & $66.4_{\pm8.7}$ & 36.61 (11) & 40.89 (11) & 34.51 (11) \\
Random best-of-36 & 36 & $46.3_{\pm8.2}$ & $8.7_{\pm3.8}$ & $68.8_{\pm2.3}$ & 38.88 (8) & 43.38 (9) & 37.01 (8) \\
Evol.\ Merge ($[0,1]^8$, 36 evals) & 36 & $47.2_{\pm1.0}$ & $2.8_{\pm1.0}$ & $72.2_{\pm0.2}$ & 40.90 (7) & 45.90 (5) & 39.33 (5) \\
MAP ($n=45$, $[0,1]^8$) & 45 & $38.9_{\pm1.5}$ & $2.4_{\pm1.4}$ & $68.2_{\pm0.5}$ & 37.01 (10) & 42.89 (10) & 36.55 (9) \\
Evol.\ Merge (gen$=$20, $[0,1]^8$) & 200 & $52.6_{\pm0.6}$ & $3.3_{\pm0.1}$ & $74.6_{\pm0.3}$ & 45.71 (3) & 50.12 (1) & 43.99 (1) \\
\bottomrule
\end{tabular}

\caption{Over-refusal and safety-adjusted OOD AVG. $R_h$: refusal rate on the 749 harmful WildGuardTest prompts (the OOD safety score); $R_b$: refusal rate on 940 benign prompts; $S_{\mathrm{bal}}=(R_h+100-R_b)/2$. AVG$_{\mathrm{bal}}$ replaces the safety score by $S_{\mathrm{bal}}$, and AVG$_4$ averages the other four domains. Ranks are among the 11 listed methods and use unrounded values; seeded methods show mean $\pm$ standard deviation over three seeds for $R_h$, $R_b$, and $S_{\mathrm{bal}}$ and means in the AVG columns.}
\label{tab:reb-overrefusal}
\end{table}

\FloatBarrier
\subsection{Sensitivity to the Safety Weight}
\label{app:reb-ws}

Table~\ref{tab:reb-ws} shows how the selected coefficients respond when the safety weight $w_s$ in the objective $(M_{\mathrm{inst}}+M_{\mathrm{math}}+M_{\mathrm{code}}+M_{\mathrm{multi}}+w_sM_{\mathrm{safety}})/(4+w_s)$ is reduced (predictions only; no new measurements).
Gemma drops the Safety expert only at $w_s=0$; Llama adds Instruction A at $w_s=0.75$ and drops Safety at $w_s\le0.5$, and its best lattice point switches from Instruction B+Safety to Instruction A+B at $w_s\approx0.88$ (0.82 with the lattice of SLIM$_{\mathrm{R}}$, Appendix~\ref{app:reb-stability}).

\begin{table}[htbp]
\centering
\footnotesize
\setlength{\tabcolsep}{4pt}
\begin{tabular}{@{}lrrrcrrrc@{}}
\toprule
 & \multicolumn{4}{c}{Gemma-2-9B} & \multicolumn{4}{c}{Llama-3.1-8B} \\
\cmidrule(lr){2-5}\cmidrule(l){6-9}
 & \multicolumn{3}{c}{SLIM coefficients} & Best & \multicolumn{3}{c}{SLIM coefficients} & Best \\
\cmidrule(lr){2-4}\cmidrule(lr){6-8}
$w_s$ & IA & IB & S & lattice point & IA & IB & S & lattice point \\
\midrule
1 & 0.29 & 0.45 & 0.26 & IA+IB & 0 & 0.70 & 0.30 & IB+S \\
0.75 & 0.31 & 0.47 & 0.22 & IA+IB & 0.22 & 0.66 & 0.13 & IA+IB \\
0.5 & 0.36 & 0.48 & 0.16 & IA+IB & 0.40 & 0.60 & 0 & IA+IB \\
0.25 & 0.49 & 0.48 & 0.03 & IA+IB & 0.46 & 0.54 & 0 & IA+IB \\
0 & 0.58 & 0.42 & 0 & IA+IB & 0.52 & 0.48 & 0 & IA+IB \\
\bottomrule
\end{tabular}

\caption{Coefficients of SLIM and the best lattice point when the safety weight $w_s$ in the objective is reduced (lattice measurements of Section~\ref{sec:experiments}; predictions only). Only Instruction A (IA), Instruction B (IB), and Safety (S) receive nonzero weight; 0 denotes weights below $10^{-6}$.}
\label{tab:reb-ws}
\end{table}
\FloatBarrier


\end{document}